\pdfoutput=1
\documentclass[10pt,twocolumn,letterpaper]{article}

\PassOptionsToPackage{table}{xcolor}

\usepackage[pagenumbers]{cvpr}

\usepackage{amsmath,amsfonts,bm}

\def\eqref#1{equation~\ref{#1}}
\def\1{\bm{1}}

\DeclareMathAlphabet{\mathsfit}{\encodingdefault}{\sfdefault}{m}{sl}
\SetMathAlphabet{\mathsfit}{bold}{\encodingdefault}{\sfdefault}{bx}{n}

\usepackage{array}
\usepackage{amsthm}
\usepackage{thmtools}
\usepackage{thm-restate}
\usepackage{float}
\usepackage{placeins}
\usepackage{multirow}
\usepackage{needspace}
\usepackage[most]{tcolorbox}
\usepackage{booktabs}

\newlength{\reconlab}
\newlength{\heatlab}

\newlength{\abllab}
\newcolumntype{Q}[1]{>{\centering\arraybackslash}p{#1}}

\definecolor{oodcolor}{RGB}{255,243,224}   % orange = OOD
\definecolor{ourscolor}{RGB}{222,235,250}  % blue = Ours
\definecolor{fuseregcolor}{RGB}{255,237,213}  % orange = FuseReg rows
\definecolor{figorange}{RGB}{246,133,12}   % Figure 3 orange
\definecolor{figblue}{RGB}{55,117,191}     % Figure 3 blue
\definecolor{heatorange}{HTML}{F87400}     % vivid heatmap orange
\definecolor{heatblue}{HTML}{1A74F2}       % vivid heatmap blue
\definecolor{theorytitle}{HTML}{B85C00}    % dark orange for readable titles
\definecolor{theoryrule}{RGB}{246,133,12}  % Figure 3 orange accent
\definecolor{theorywash}{HTML}{FFF3E0}     % light orange background

\tcbset{
  mainresult/.style={
    enhanced,
    frame hidden,
    colback=theorywash,
    borderline west={1.5pt}{0pt}{theoryrule},
    coltitle=theorytitle,
    fonttitle=\normalfont\bfseries\upshape,
    description color=black,
    description font=\normalfont\upshape,
    separator sign none,
    detach title,
    fontupper=\normalfont,
    boxrule=0pt,
    sharp corners,
    boxsep=0pt,
    left=7pt,
    right=6pt,
    top=5pt,
    bottom=5pt,
    before upper={%
      \setlength{\parskip}{3pt}%
      \setlength{\abovedisplayskip}{5pt}%
      \setlength{\belowdisplayskip}{5pt}%
      \setlength{\abovedisplayshortskip}{3pt}%
      \setlength{\belowdisplayshortskip}{4pt}%
      \tcbtitle\par\vspace{2pt}%
    },
    before skip=6pt plus 2pt minus 1pt,
    after skip=6pt plus 2pt minus 1pt,
  }
}
\newtcbtheorem[
  use counter*=lemma
]{mainlemma}{Lemma}{mainresult,label type=lemma}{lem}
\newtcbtheorem[
  use counter*=proposition
]{mainproposition}{Proposition}{mainresult,label type=proposition}{prop}
\newtcbtheorem[
  use counter*=corollary
]{maincorollary}{Corollary}{mainresult,label type=corollary}{cor}
\newtcbtheorem[
  use counter*=remark
]{mainremark}{Remark}{mainresult,label type=remark}{rem}
\newtcbtheorem[
  use counter*=hypothesis
]{mainhypothesis}{Hypothesis}{mainresult,label type=hypothesis}{hyp}

\newtcolorbox{insight}{
  enhanced,
  colback=figblue!5!white,
  colframe=figblue,
  boxrule=0.6pt,
  arc=2pt,
  boxsep=0pt,
  left=7pt,
  right=7pt,
  top=5pt,
  bottom=5pt,
  fontupper=\normalfont,
  before upper={\textcolor{figblue}{\bfseries Insight.}\ },
  before skip=4pt plus 1pt minus 1pt,
  after skip=5pt plus 1pt minus 1pt,
}

\newcommand{\ours}{FuseReg}

\definecolor{cvprblue}{rgb}{0.21,0.49,0.74}
\usepackage[pagebackref,breaklinks,colorlinks,allcolors=cvprblue]{hyperref}
\usepackage{etoc}

\AtEndPreamble{
  \crefname{table}{Table}{Tables}
  \crefname{lemma}{Lemma}{Lemmas}
  \Crefname{lemma}{Lemma}{Lemmas}
  \crefname{proposition}{Proposition}{Propositions}
  \Crefname{proposition}{Proposition}{Propositions}
  \crefname{corollary}{Corollary}{Corollaries}
  \Crefname{corollary}{Corollary}{Corollaries}
  \crefname{remark}{Remark}{Remarks}
  \Crefname{remark}{Remark}{Remarks}
  \crefname{hypothesis}{Hypothesis}{Hypotheses}
  \Crefname{hypothesis}{Hypothesis}{Hypotheses}
}

\def\confName{CVPR}
\def\confYear{2027}

\title{Rethinking the Reconstruction–Generation Gap in Representation Autoencoders}

\author{%
Hongyang Du$^{1,2,*}$\quad Yunfei Xie$^{3,*}$\quad Junjie Ye$^{1,*}$\quad Jiawei Yang$^{1}$\quad Xiaoyan Cong$^{2}$\quad Haodong Zhang$^{2}$\\
Yongchao Huang$^{4}$\quad Haiyu Wu$^{5}$\quad Zongxia Li$^{6}$\quad Shihang Gui$^{2}$\quad Dawei Liu$^{7}$\\
Runhao Li$^{1}$\quad Jingcheng Ni$^{2}$\quad Chen Wei$^{3,\dagger}$\quad Randall Balestriero$^{2,\dagger}$\quad Yue Wang$^{1,\dagger}$\\[4pt]
{\normalsize $^{1}$USC PSI Lab\quad $^{2}$Brown University\quad $^{3}$Rice University\quad $^{4}$University of Aberdeen}\\
{\normalsize $^{5}$University of Notre Dame\quad $^{6}$University of Maryland, College Park\quad $^{7}$University of Pennsylvania}\\[2pt]
{\small $^{*}$Equal contribution\quad $^{\dagger}$Equal advising \quad \small Correspondence: \texttt{hongyang\_du@brown.edu}}
}

\begin{document}
\etocdepthtag.toc{mtmain}

\maketitle

\begin{abstract}
Fusing shallower encoder layers into a Representation Autoencoder (RAE) sharpens pixel reconstruction but degrades diffusion generation. Prior studies and the research community commonly attribute this trade-off to the latent content itself, assuming detail-rich features are harder to model. We show that this gap stems from geometric mismatch at the encoder-decoder interface. Freezing a foundation encoder preserves rich visual information, but imposes an encoder-native Euclidean metric shaped without pixel decoding, whereas the decoder amplifies sampling errors sharply along sensitive directions. At fixed information, reducing pixel error admits two dual geometric interventions: regularizing the decoder to flatten its sensitivity to latent variations, and aligning diffusion coordinates with the frozen decoder's metric at identical clean reconstruction. On ImageNet-256, aligning diffusion coordinates largely closes the guided generation penalty of multi-layer fusion and reaches state-of-the-art 0.984 gFID and 2.14 $\mathrm{FD}_r^6$ in 80 epochs. RAEs should take information from the encoder and geometry from the decoder.
Project page: \url{https://hongyang-du.github.io/FuseLift/}.
\end{abstract}

\label{abstract}

\section{Introduction}
\label{sec:intro}

\begin{figure}[t]
  \centering
  \includegraphics[width=\linewidth]{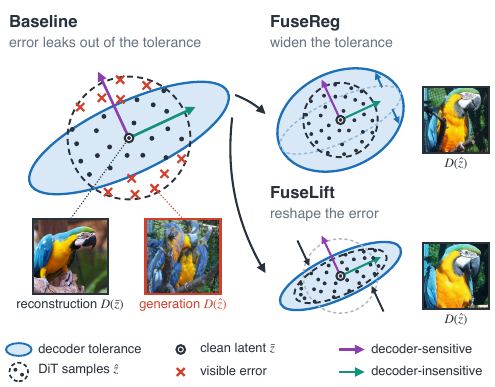}
  \caption{\textbf{Same information, different geometry.} Around a clean multi-layer latent $\bar z$, diffusion makes isotropic sampling errors (dashed circles), whereas a pixel decoder tolerates far smaller errors along sensitive directions than insensitive ones (blue ellipses). Even when clean reconstruction $D(\bar z)$ is sharp, off-axis errors spill outside decoder tolerance and corrupt generation $D(\hat z)$. At fixed encoded information, pixel error shrinks either by widening decoder tolerance or by reshaping generator errors.}
  \label{fig:teaser}
\end{figure}

Latent generative modeling has progressed from compressing pixels in dedicated autoencoders~\citep{rombach2022ldm,peebles2023dit} to building latent spaces directly on frozen vision foundation encoders~\citep{simeoni2025dinov3,tschannen2025siglip} in Representation Autoencoders (RAEs)~\citep{zheng2025rae}. Because foundation encoders capture high-level semantics in deep layers and fine spatial structure in shallow layers, they offer a natural basis for both understanding and synthesis. Yet tapping into shallower layers exposes a persistent reconstruction--generation dilemma~\citep{yao2025lightningdit,singh2026raev2}: multi-layer fusion markedly sharpens pixel reconstruction while degrading diffusion generation, with unguided and guided sampling peaking at different depths $k$. Prevailing wisdom treats this penalty as informational, assuming detail-rich latents are intrinsically harder to fit and tuning $k$ to trade visual fidelity for sample quality.

We argue that this trade-off is largely a failure of \emph{interface geometry}, not an overload of \emph{latent information}. A latent's information, what it encodes about the image, is fixed by the encoder and $k$, whereas its geometry governs how perturbations translate into pixel errors. Diffusion injects isotropic Gaussian noise and minimizes squared Euclidean error as if all latent directions mattered equally, yet a pixel decoder amplifies errors far more steeply along sensitive directions than insensitive ones.

Jointly trained VAEs couple latent scaling with decoder tolerance as sampling noise and gradients flow through both networks~\citep{rombach2022ldm}. Freezing a representation encoder preserves rich features yet severs that link: clean reconstruction supervises the decoder solely on $\bar z(x)$, while diffusion operates in pixel-agnostic encoder coordinates. Under multi-layer fusion, this decoupling creates two geometric mismatches (\Cref{fig:teaser}). \emph{Off the clean manifold}, standard training feeds only the exact $k$-layer average, so per-layer deviations cancel to zero and leave the decoder brittle along cross-layer disagreement directions, which structure-agnostic Gaussian noise~\citep{zheng2025rae,singh2026raev2} fails to regularize. \emph{Near the clean manifold}, once guidance pulls trajectories out of the low-rank disagreement span, diffusion still weights all $d$ channels equally even though the decoder's input projection stretches pixel-critical directions and compresses others.

Holding encoded information strictly fixed on full-fusion DINOv3-L ($k{=}23$) ImageNet-256 latents, two geometry-only interventions resolve these mismatches. First, holding a trained DiT-XL and its exact 50k sampled latents fixed, \textbf{\ours} swaps only the decoder: training on size-normalized averages of random layer subsets anchors the expected input at full fusion while flattening disagreement sensitivity. Decoding the \emph{exact same} latents lowers 80-epoch unguided gFID from $3.01$ to $2.21$, showing that baseline samples already contain sharper structure than the standard decoder renders. Second, holding the frozen decoder and every clean reconstruction identical ($27.04$\,dB PSNR), \textbf{FuseLift} moves diffusion one affine step into the decoder, right after its linear input projection, weighting directional signal-to-noise ratio and squared loss by first-layer sensitivity before nonlinear blocks erode linear semantics.

At scale, the two interventions govern complementary regimes. Widening off-manifold tolerance with \ours\ drives unguided gains, whereas aligning all-channel coordinates with FuseLift narrows the converged guided gap, lowering full-fusion gFID from $1.25$ to $1.07$ to match the seven-layer baseline at $4.5$\,dB higher PSNR. On the seven-layer decoder, FuseLift reaches state-of-the-art gFID of $0.984$ excluding FD-loss methods~\citep{yang2026rfl}, alongside $2.14$ $\mathrm{FD}_r^6$, outperforming strong pixel-space, latent, and autoregressive baselines~\citep{li2025jit,leng2025repae,zheng2025rae,yu2026bar,singh2026raev2} across both Inception and modern vision backbones. More broadly, tracing targets across decoder depth places encoder-space diffusion (linear semantics, pixel-agnostic metric) and pixel-space diffusion (pixel metric, nonlinear semantics) on one axis, where a single affine step captures both. RAEs need not sacrifice visual detail for diffusability: they should take \emph{information} from the encoder and \emph{geometry} from the decoder.

\paragraph{Contributions.}
\begin{itemize}[leftmargin=1em, itemsep=1pt, topsep=2pt]
    \item \textbf{A geometric account of the gap.}
    We separate latent information from interface geometry, tracing the multi-layer penalty to off-manifold disagreement brittleness and near-manifold metric mismatch.

    \item \textbf{Two information-preserving interventions.}
    \textbf{\ours} flattens decoder disagreement sensitivity via mean-preserving random layer subsets, and \textbf{FuseLift} aligns diffusion coordinates in the frozen decoder's affine prefix at identical clean reconstruction.

    \item \textbf{Regime and depth characterization.}
    Controlled sweeps across guidance regimes, encoder families, and decoder depths isolate when tolerance versus coordinate alignment governs generation, improving guided quality across both full-fusion and seven-layer RAEs.
\end{itemize}

\section{Related Work}
\label{sec:related}

% Queue the toy figure and wide cosine panel on page 2 so the panel tops page 3.

% Queue the wide cosine-similarity panel here for the top of page 3.

\paragraph{RAEs and latent alignment.}
RAEs pair frozen vision foundation encoders with trainable pixel decoders to supply semantic diffusion latents~\citep{zheng2025rae}, recently extended to video~\citep{guo2026vrae,xie2026videorae}. Subsequent work modifies tokenizer architectures or latent spaces~\citep{jia2025dinotok,bi2025vfmvae,chen2025aligntok}, or aligns DiT hidden states with pretrained features~\citep{yu2024repa,leng2025repae,singh2025whatmatters,shin2026pixelrepa}. These methods still diffuse under a Euclidean metric detached from pixel-decoder sensitivity. Keeping the encoder frozen, we separate latent information from interface geometry, widening decoder error tolerance (\ours) and reshaping diffusion coordinates (FuseLift).

\paragraph{Multi-layer feature fusion.}
RAEv2~\citep{singh2026raev2} averages the last $k$ encoder layers and shows that unguided sampling favors $k{=}1$ whereas guided sampling peaks near $k{=}7$, leaving full-fusion ($k{=}23$) fidelity out of reach without accounting for the regime split. DRoRAE~\citep{zhu2026beyondlast} and attentive multi-layer fusion~\citep{ciernik2026multilayer} learn modules that adaptively aggregate features across encoder layers for tokenization and probing, modifying the fused representation itself. Rather than truncating depth or adding a learned fusion module, we keep the uniform full-fusion latent fixed at inference and address the unguided and guided gaps through interface geometry.

\paragraph{Structured latent perturbation.}
Standard RAE decoders inject additive isotropic Gaussian noise during Stage-1 training~\citep{zheng2025rae,singh2026raev2}, ignoring inter-layer structure, whereas dropout, nested dropout~\citep{srivastava2014dropout,rippel2014learning}, LayerDrop, and Matryoshka training~\citep{fan2019layerdrop,kusupati2022matryoshka} vary active units inside a trainable backbone. \ours\ instead samples size-normalized subsets of frozen encoder layers, anchoring the expected input at the full-fusion latent while reducing decoder sensitivity along cross-layer disagreement directions.

\section{Preliminary}
\label{sec:Preliminary}
\label{sec:preliminaries}

\begin{figure}[t]
  \centering
  \includegraphics[width=\linewidth]{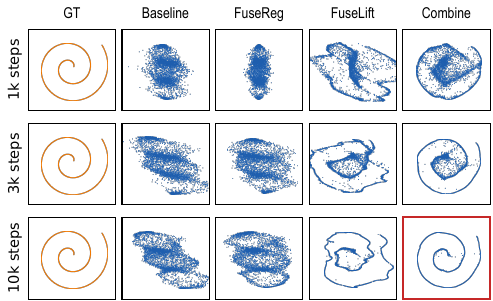}
  \caption{\textbf{Two orthogonal geometric mismatches on a synthetic spiral.} Clean latents share identical reconstruction (\textbf{GT}), yet isotropic diffusion errors interact with an anisotropic decoder to cause both off-manifold scatter and in-plane smearing (\textbf{Baseline}). Flattening off-manifold sensitivity $G_D$ (\textbf{\ours}) removes the scatter, reweighting diffusion coordinates $C_G$ (\textbf{FuseLift}) removes the smearing, and combining both recovers the spiral when the two sensitive axes are orthogonal. \textcolor[HTML]{C62828}{(Orthogonality is an idealized condition, not always true in real life.)}}
  \label{fig:toy-spiral}
\end{figure}

Given an input image $x$, a Representation Autoencoder (RAE) extracts layer-normalized patch features $f_1,\dots,f_K$ (shallow to deep, channel width $d$) from a frozen visual encoder $E$. Fixed fusion averages $k$ selected layers (the last $k$ in RAEv2~\citep{singh2026raev2}), and full fusion ($k{=}K$) averages all layers into $\bar z=K^{-1}\sum_{i=1}^K f_i$. Stage 1 trains a pixel decoder $D_\phi$ to reconstruct $x$ from $\bar z$. Stage 2 freezes both $E$ and $D_\phi$ and trains a Diffusion Transformer (DiT) on $\bar z$ under isotropic Gaussian noise and squared Euclidean loss (fixed channel standardization is absorbed into $\bar z$ and $D_\phi$; see \S\ref{app:fusereg-method}, \S\ref{app:interface}). At sampling, the DiT draws $\hat z=\bar z+\delta$ with sampling error $\delta$, rendered into pixels as $\hat x=D_\phi(\hat z)$.

\paragraph{Information and geometry.}
The \emph{information} in $\bar z$ is the image content fixed by the encoder and the $k$ fused layers, whereas its \emph{geometry} governs how sampling errors $\delta$ distort the rendered image. Clean reconstruction supervises $D_\phi$ only on $\bar z$, while freezing $E$ locks in Euclidean diffusion coordinates without feedback from pixel decoding. Near $\bar z$, small perturbations act through the decoder Jacobian $J_{D_\phi}(\bar z)$, weighting latent directions by the symmetric sensitivity matrix $G_D=J_{D_\phi}(\bar z)^\top J_{D_\phi}(\bar z)$. Letting $C_G=\mathbb E[\delta\delta^\top\mid\bar z]$ denote the generator's conditional error second moment, a first-order expansion gives expected squared pixel error (\S\ref{app:interface}):
\begin{equation}
  \mathbb E\!\left[\|D_\phi(\bar z+\delta)-D_\phi(\bar z)\|_2^2
    \,\middle|\,\bar z\right]
  \approx\operatorname{tr}(G_D C_G).
  \label{eq:interface-error-main}
\end{equation}
Pixel artifacts grow whenever error $C_G$ remains large along directions where decoder sensitivity $G_D$ is steep.

\paragraph{Reducing error at fixed information.}
Changing the encoder or $k$ alters encoded content, whereas an intervention is \emph{information-preserving} if it keeps $\bar z$ or maps it by an invertible affine transform on its range. At fixed information, \Cref{eq:interface-error-main} exposes two complementary levers: training the decoder so $G_D$ flattens along error-prone directions (\S\ref{sec:fusereg}), or freezing the decoder and reweighting diffusion coordinates so $C_G$ shrinks along sensitive directions at identical clean reconstruction (\S\ref{sec:fuselift}).

\paragraph{Synthetic spiral.}
In \Cref{fig:toy-spiral}, a 2D spiral embedded in 3D shares identical clean reconstruction across settings, while a linear decoder stretches one in-plane axis by $8\times$ and reacts sharply along the third axis where frozen layers vary around their mean. Flattening off-manifold sensitivity in $G_D$ removes the vertical scatter, reweighting diffusion coordinates $C_G$ along the stretched axis removes horizontal smearing, and combining both recovers the spiral when those two axes are orthogonal. When the two subspaces overlap on real encoders, flattening $G_D$ along a shared direction reduces the marginal gain from also shrinking $C_G$ there (\S\ref{sec:experiments}).

\paragraph{Evaluation setup.}
\label{sec:exp-setup}
Unless otherwise specified, our controlled experiments use the RAEv2 DINOv3-L setup with $k{=}23$, and \ours\ uses $p_{\mathrm{dec}}{=}0.95$; see \S\ref{app:decoder-rate} and \S\ref{app:decoder-swap} for the rate-selection results.
Main runs use class-conditional ImageNet-256~\citep{russakovsky2015imagenet}. Matched runs share architecture, parameters, optimizer, and step budgets (16-epoch decoders, 40-epoch sweeps, 80-epoch DiT-XL runs with EMA). Reconstruction uses PSNR, SSIM, LPIPS, and reconstruction FID (rFID) on 50k validation images. Generation uses generation FID (gFID), Inception Score (IS), and mean representation Fr\'echet distance across six extractors ($\mathrm{FD}_r^6$~\citep{yang2026rfl}) on 50k samples (50 Euler steps), with internal guidance~\citep{singh2026raev2,zhou2026internal} at scale $1.78$ over denoising times $t\in[0.1,1]$ from noise ($t{=}0$) to data ($t{=}1$).

\section{Reshaping Decoder Geometry}
\label{sec:fusereg}
\label{sec:method}

\subsection{FuseReg: Layer-wise View Regularization}
\label{sec:theory}

\begin{figure}[t]
  \centering
  \includegraphics[width=\linewidth]{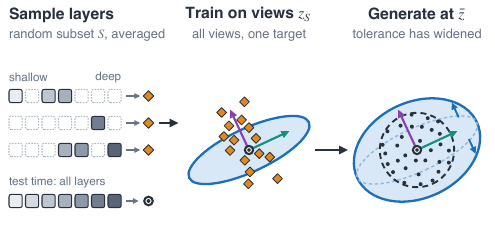}
  \caption{\textbf{Widening decoder tolerance with \ours.} Averaging random nonempty subsets of frozen encoder layers forms mean-preserving views $z_S$ reconstructed to the same image, widening tolerance around $\bar z$ without changing the all-layer inference latent.}
  \label{fig:fusereg-concept}
\end{figure}

% Queue the wide evidence panel early so it stays with the FuseReg discussion.
\begin{figure*}[t]
  \centering
  \includegraphics[width=\textwidth]{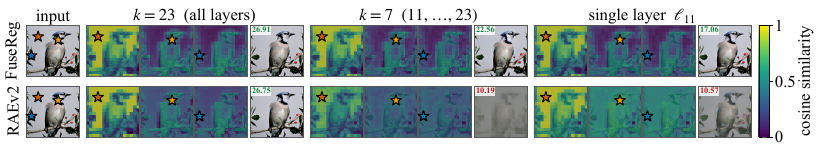}
  \caption{\textbf{Standard full-fusion decoding collapses off the all-layer consensus.} Internal patch similarity (star queries) and reconstructions (insets: PSNR in dB) on all layers, seven layers, or one layer: standard RAEv2 loses spatial structure once shallow layers are dropped (bottom), whereas one \ours\ decoder stays coherent across subsets (top).}
  \label{fig:cosine-robust}
\end{figure*}

A standard full-fusion DINOv3-L decoder ($k{=}23$) reconstructs clean all-layer latents accurately, yet dropping even a single shallow layer costs up to $3$\,dB PSNR (\Cref{fig:layer-usage-k23}a, gray), and feeding sparser subsets collapses spatial structure (\Cref{fig:cosine-robust}, bottom). This brittleness arises because clean reconstruction supervises the decoder only on the exact all-layer mean $\bar z$: individual layer deviations $f_i-\bar z$, which we call \emph{cross-layer disagreements}, cancel to zero in $\bar z$ and leave decoder sensitivity $G_D$ unconstrained and steep along disagreement directions (\Cref{prop:separation}, \S\ref{app:theory}).

To flatten sensitivity along these directions without altering the inference latent, \textbf{\ours} trains the decoder to reconstruct the target image from random subsets of frozen encoder layers (\Cref{fig:fusereg-concept}). During Stage-1 training, each encoder layer is dropped independently at rate $p_{\mathrm{dec}}$, conditioned on keeping a nonempty subset $S$, and the retained layers are averaged into a training view:
\begin{equation}
  z_S=\frac{1}{|S|}\sum_{i\in S}f_i.
  \label{eq:fusereg-main}
\end{equation}
We train the decoder on $z_S$ under the standard reconstruction loss ($p_{\mathrm{dec}}{=}0.95$ by default) without additive Gaussian noise. Test-time inference still feeds the unmodified all-layer average $\bar z$, leaving both the encoder and the diffusion target untouched (\S\ref{app:fusereg-method}).

Because layers are sampled symmetrically and divided by $|S|$, the expected training view stays anchored at full fusion, $\mathbb E_S[z_S]=\bar z$ (\Cref{lem:subset-mean}, \S\ref{app:theory}). View perturbations therefore remain centered on $\bar z$ and vary strictly within the cross-layer disagreement subspace. For a linear decoder under squared error, training across random views equals the clean full-fusion objective plus an explicit quadratic penalty on disagreement sensitivity (\S\ref{app:theory}).

\subsection{Decoder Tolerance and Layer Attribution}
\label{sec:decoder-analysis}

\begin{figure}[t]
  \centering
  \includegraphics[width=\columnwidth]{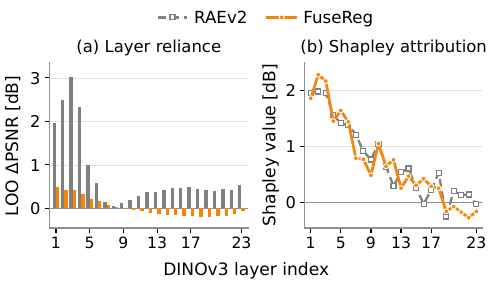}
  \caption{\textbf{\ours\ removes single-layer brittleness while preserving shallow-layer contribution.} \textbf{(a)} Leave-one-out PSNR drops show that \ours\ eliminates sharp sensitivity to removing individual shallow layers. \textbf{(b)} Shapley attributions confirm that shallow layers remain the primary contributors to reconstruction.}
  \label{fig:layer-usage-k23}
\end{figure}

\paragraph{Spatial coherence across subsets.}
Evaluating a trained decoder on reduced layer subsets probes cross-layer disagreement directions directly (\Cref{fig:cosine-robust}, \S\ref{app:qualitative}). Whereas the standard full-fusion decoder loses internal patch correspondence as soon as shallow layers are omitted, a single \ours\ decoder maintains coherent spatial structure and faithful reconstructions across all 23 layers, the last seven layers, and even a single intermediate layer.

\paragraph{Lower single-layer reliance, preserved shallow detail.}
In leave-one-out tests on the 23-layer input, \ours\ reduces the standard decoder's $3$\,dB shallow-layer PSNR drops to below $0.5$\,dB (\Cref{fig:layer-usage-k23}a). Computing each layer's Shapley value, its average marginal PSNR gain across random subset orderings (\Cref{fig:layer-usage-k23}b), confirms that the five shallowest layers remain the largest contributors: \ours\ preserves shallow detail while removing brittle reliance on cross-layer consensus. Matching the same perturbation energy with isotropic Gaussian noise or token and channel dropout fails to achieve this tolerance (\S\ref{app:subset-diagnostics}, \S\ref{app:extreme-shifts}).

\section{Giving DiTs Decoder Geometry}
\label{sec:fuselift}
\label{sec:bypass}
\label{sec:h1-method}

\subsection{FuseLift: Affine Prefix Diffusion}

\begin{figure}[t]
  \centering
  \includegraphics[width=\linewidth]{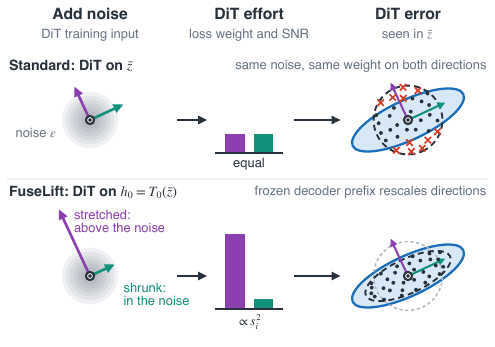}
  \caption{\textbf{Aligning generator geometry with FuseLift.} Diffusing on encoder latents $\bar z$ assigns equal noise and loss weight to all directions, letting errors spill along sensitive axes (top). Diffusing after the frozen decoder's affine input layer $h_0$ scales directional signal-to-noise ratio and squared loss by first-layer gains $s_i^2$, fitting errors inside tolerance at identical clean reconstruction (bottom).}
  \label{fig:fuselift-concept}
\end{figure}

Across $K$ encoder layers, centered layer deviations sum to zero and span at most $K{-}1$ directions per token, covering at most $22$ dimensions inside DINOv3-L's $1024$-channel space. Because \ours\ perturbs inputs only within that low-rank span, decoder sensitivity across the remaining channels stays unchanged, while standard diffusion still injects equal noise and equal loss weight across all $1024$ channels. To align generator errors with decoder sensitivity across the full latent space without retraining the decoder, \textbf{FuseLift} moves diffusion from encoder latents $\bar z$ into an internal activation of the frozen decoder (\Cref{fig:fuselift-concept}).

Split the frozen decoder at depth $j$ into a prefix $T_j$ and a rendering suffix $R_j$, producing internal activation $h_j=T_j(\bar z)$. Here $h_0$ lies immediately after the decoder's affine input projection (including fixed positional embeddings, ahead of block $1$), while $h_j$ ($j\ge 1$) is the output of decoder block $j$. FuseLift defaults to $j{=}0$: applying fixed invertible per-channel standardization $\mathcal N_0$ defines the lifted diffusion target $u_0$, and the frozen suffix renders generated samples $\hat u_0$ back to pixels:
\begin{equation}
  u_0=\mathcal N_0(h_0),\qquad
  \hat x=R_0\bigl(\mathcal N_0^{-1}(\hat u_0)\bigr).
  \label{eq:lift-target-main}
\end{equation}
Only the DiT is trained on $u_0$, keeping the baseline parameter count, noise schedule, Euclidean loss, and guidance recipe unchanged (\S\ref{app:fuselift-method}). Because standardization is inverted before the frozen suffix, every clean image reconstructs identically (\Cref{prop:lift-reconstruction}, \S\ref{app:fuselift}), and affine $T_0$ retains every component of $\bar z$ read by the decoder.

\subsection{Why an Affine Step Reshapes Diffusion}
\label{sec:why-fuselift}

Combine the frozen input projection and channel standardizations into a single affine map $u_0=A\bar z+a$ (\S\ref{sec:lift-affine}). Per-channel standardization equalizes marginal channel variances without removing cross-channel correlations, so $A$ preserves distinct directional gains $s_i$ along its right singular vectors $v_i$ (\S\ref{sec:lift-snr}).

Under flow matching at denoising time $t\in(0,1)$, adding unit Gaussian noise and minimizing squared error in $u_0$-space rather than $\bar z$-space rescales both the squared-error weight on an in-range latent error $\delta$ and the directional signal-to-noise ratio ($\operatorname{SNR}_i$, for clean directional variance $\sigma_i^2$) by $s_i^2$ (\Cref{fig:fuselift-concept}, bottom; \Cref{prop:lift-affine-geometry} in \S\ref{sec:lift-affine}):
\begin{equation}
  \|A\delta\|_2^2=\sum_i s_i^2\langle v_i,\delta\rangle^2,\ \
  \operatorname{SNR}_i(t)=\frac{t^2}{(1-t)^2}s_i^2\sigma_i^2.
  \label{eq:lift-gain-main}
\end{equation}
Directions stretched by the decoder's input layer stand further above the noise and incur steeper loss penalties during training, whereas compressed directions recede into the noise. Because every latent perturbation passes through $A$ before reaching the nonlinear decoder blocks, weighting diffusion by $s_i^2$ acts as a first-layer proxy for full decoder sensitivity (\S\ref{sec:lift-spectrum}) while keeping high-level semantics linearly accessible (\S\ref{sec:lift-depth}).

\begin{table*}[t]
  \centering
  \caption{\textbf{Separating decoder tolerance and diffusion coordinates at fixed full-fusion information ($k{=}23$).}
  Holding encoded content fixed on 80-epoch DiT-XL, \textbf{\ours} (orange) decodes the \emph{exact same} sampled latents as standard RAEv2 and sharply improves unguided generation despite higher clean rFID, whereas \textbf{FuseLift} (blue) diffuses in the frozen RAEv2 decoder's affine prefix at \emph{identical} clean reconstruction to improve guided generation (\textbf{bold}: best per column).}
  \label{tab:main-generation}
  \small
  \setlength{\tabcolsep}{2pt}
  % Fixed-width columns keep row shading continuous across the full table.
  \newlength{\mainmethodwidth}
  \settowidth{\mainmethodwidth}{\small RAEv2 ($k{=}23$)}
  \begin{tabular}{p{\mainmethodwidth}*{8}{Q{\dimexpr(\textwidth-\mainmethodwidth-18\tabcolsep)/8\relax}}}
    \toprule
    & \multicolumn{4}{c}{Reconstruction} & \multicolumn{2}{c}{Unguided} & \multicolumn{2}{c}{Guided} \\
    \cmidrule(lr){2-5}\cmidrule(lr){6-7}\cmidrule(lr){8-9}
    Method & PSNR$\uparrow$ & SSIM$\uparrow$ & LPIPS$\downarrow$ & rFID$\downarrow$ & gFID$\downarrow$ & IS$\uparrow$ & gFID$\downarrow$ & IS$\uparrow$ \\
    \midrule
    RAEv2 & 27.04 & 0.806 & 0.091 & \textbf{0.18} & 3.01 & 207.4 & 1.25 & 256.8 \\
    \rowcolor{fuseregcolor}
    + FuseReg & \textbf{27.52} & \textbf{0.826} & \textbf{0.088} & 0.42 & \textbf{2.21} & \textbf{212.8} & 1.25 & \textbf{260.1} \\
    \rowcolor{ourscolor}
    + FuseLift & 27.04 & 0.806 & 0.091 & \textbf{0.18} & 2.95 & 193.6 & \textbf{1.07} & 256.1 \\
    \bottomrule
  \end{tabular}
\end{table*}

\section{Main Results}
\label{sec:main-results}

\subsection{Fixed Latents, Swapped Decoder}
\label{sec:decoder-benefits}
\label{sec:decoder-swap}
\label{sec:fusereg-insight}

\begin{figure}[t]
  \centering
  \includegraphics[width=\columnwidth]{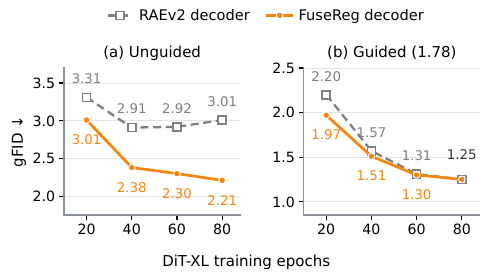}
  \caption{\textbf{Decoder swap improves unguided generation yet vanishes as guided sampling converges.} Rendering identical sampled latents through \ours\ instead of RAEv2 steadily lowers unguided gFID \textbf{(a)}, whereas early guided gains vanish once trajectories converge near the clean manifold \textbf{(b)}.}
  \label{fig:decoder-swap}
\end{figure}

To isolate decoder tolerance from generator errors, we hold a full-fusion ($k{=}23$) DiT-XL and its exact 50k sampled latents fixed at every checkpoint and swap only the decoder (\Cref{fig:decoder-swap}). Without guidance, rendering those identical samples through \ours\ instead of standard RAEv2 cuts 80-epoch gFID from $3.01$ to $2.21$ while raising Inception Score (\Cref{fig:decoder-swap}a, \Cref{tab:main-generation}). On clean validation images, \ours\ also improves pixelwise fidelity (PSNR, SSIM, and LPIPS) despite a higher clean rFID (\Cref{tab:main-generation}), showing that clean-manifold distribution metrics do not govern off-manifold sampling tolerance.

Under internal guidance, swapping in \ours\ improves early checkpoints yet converges to the same $1.25$ gFID as standard RAEv2 by 80 epochs (\Cref{fig:decoder-swap}b), even as the unguided gain continues to widen. Matching the low-rank disagreement bound in \S\ref{sec:fuselift}, guidance pulls trajectories close to the clean manifold, where residual errors lie mostly outside the $22$-dimensional cross-layer disagreement span.

% Generated by tables/src/build_training_efficiency.py.
\begin{table*}[t]
\centering
\caption{\textbf{Guided generation on ImageNet-256 across Inception and modern vision backbones.}
Comparing pixel-space diffusion (JiT, PixelDiT), continuous latent diffusion, discrete autoregressive modeling (BAR-L), and RAEs under gFID and six-extractor Fr\'echet distance ($\mathrm{FD}_r^6$~\citep{yang2026rfl}). Under the same 80-epoch setup and frozen decoders as RAEv2, \textbf{FuseLift} (blue) matches seven-layer RAEv2 at 23-layer fidelity ($k{=}23, h_0$) and achieves the best overall gFID and $\mathrm{FD}_r^6$ at $k{=}7, h_0$ (\textbf{bold}: best; \underline{underlined}: second). SFD: adaptive dopri5.}
\label{tab:training-efficiency}
\footnotesize
\setlength{\tabcolsep}{2.0pt}
\renewcommand{\arraystretch}{1.0}
\newlength{\benchmarkmethodwidth}
\newlength{\benchmarkcitewidth}
\settowidth{\benchmarkmethodwidth}{\textbf{FuseLift} ($k{=}23$, $h_0$)}
\settowidth{\benchmarkcitewidth}{SiT-XL/2-REPA~\citep{yu2024repa}}
\ifdim\benchmarkcitewidth>\benchmarkmethodwidth
  \setlength{\benchmarkmethodwidth}{\benchmarkcitewidth}
\fi
\begin{tabular}{@{}p{\benchmarkmethodwidth}*{11}{Q{\dimexpr(\textwidth-\benchmarkmethodwidth-22\tabcolsep)/11\relax}}@{}}
\toprule
\multirow{2}{*}{Method} & \multirow{2}{*}{Epochs}
 & \multirow{2}{*}{Sampler} & \multirow{2}{*}{Steps}
 & \multirow{2}{*}{gFID$\downarrow$}
 & \multicolumn{6}{c}{Representation Fr\'echet distance (FD$_r$)$\downarrow$}
 & \multirow{2}{*}{$\mathrm{FD}_r^6\!\downarrow$} \\
\cmidrule(lr){6-11}
 & & & &
 & Incep. & \makebox[\linewidth][c]{ConvNeXt} & DINOv2 & MAE & SigLIP & CLIP & \\
\midrule
SiT-XL/2~\citep{ma2024sit} & 1400 & SDE & 250 & 1.92 & 1.14 & 1.56 & 6.64 & 5.38 & 14.66 & 15.52 & 7.48 \\
SiT-XL/2-REPA~\citep{yu2024repa} & 1400 & SDE & 250 & 1.30 & 0.77 & 1.05 & 4.14 & 4.13 & 9.89 & 12.66 & 5.44 \\
LightningDiT~\citep{yao2025lightningdit} & 800 & ODE & 250 & 1.24 & 0.74 & 0.94 & 3.65 & 3.27 & 8.63 & 10.26 & 4.58 \\
DDT-XL~\citep{wang2025ddt} & 800 & ODE & 250 & 1.29 & 0.77 & 1.01 & 4.25 & 4.33 & 10.32 & 13.82 & 5.75 \\
REPA-E~\citep{leng2025repae} & 800 & SDE & 250 & 1.12 & 0.67 & 1.21 & 2.43 & 2.71 & 5.07 & 6.26 & 3.06 \\
REG~\citep{wu2025reg} & 800 & SDE & 250 & 1.45 & 0.71 & 0.89 & 3.14 & 3.23 & 8.26 & 10.75 & 4.50 \\
RAE-XL~\citep{zheng2025rae} & 800 & ODE & 50 & 1.14 & 0.68 & 1.80 & 2.09 & 3.48 & 3.80 & 7.74 & 3.27 \\
JiT-H/16~\citep{li2025jit} & 600 & Heun & 50 & 1.87 & 1.18 & 2.48 & 4.26 & 6.02 & 11.81 & 20.35 & 7.68 \\
PixelDiT-XL~\citep{yu2026pixeldit} & 320 & ODE & 100 & 1.64 & 0.98 & 1.04 & 3.39 & 3.82 & 8.59 & 8.59 & 4.40 \\
SFD-XL~\citep{Pan2025SFD} & 800 & ODE & adapt. & 1.05 & \underline{0.62} & 0.92 & 1.52 & \textbf{1.68} & 3.89 & 5.78 & 2.40 \\
BAR-L~\citep{yu2026bar} & 400 & \makebox[\linewidth][c]{AR/MBM} & $256{\times}4$ & \underline{0.998} & \textbf{0.60} & 0.76 & 3.31 & 4.47 & 6.74 & 5.97 & 3.64 \\
RAEv2~\citep{singh2026raev2} & 80 & ODE & 50 & 1.08 & 0.71 & \underline{0.75} & \textbf{1.07} & 2.76 & \underline{2.90} & \textbf{4.95} & \underline{2.19} \\
\midrule
\rowcolor{ourscolor}
\textbf{FuseLift} ($k{=}23$, $h_0$) & 80 & ODE & 50 & 1.07 & 0.68 & \textbf{0.71} & 1.40 & 2.43 & 3.26 & 5.33 & 2.30 \\
\rowcolor{ourscolor}
\textbf{FuseLift} ($k{=}7$, $h_0$) & 80 & ODE & 50 & \textbf{0.984} & 0.64 & 0.79 & \underline{1.16} & \underline{2.20} & \textbf{2.77} & \underline{5.24} & \textbf{2.14} \\
\bottomrule
\end{tabular}
\end{table*}

\subsection{Fixed Reconstruction, Relocated Coordinates}
\label{sec:lift-results}

\begin{figure}[t]
  \centering
  \includegraphics[width=\columnwidth]{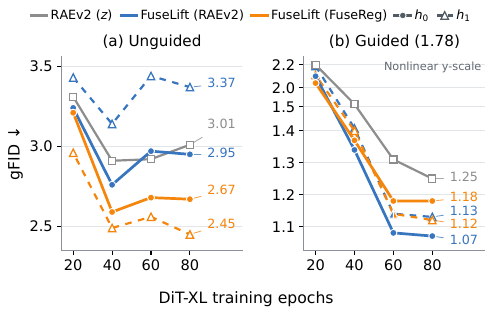}
  \caption{\textbf{Prefix coordinate alignment drives guided gains yet barely moves unguided generation.} Full-fusion trajectories comparing encoder space ($z$) against FuseLift at the affine input layer ($h_0$) and after block 1 ($h_1$). Without guidance, relocating RAEv2 to $h_0$ has little effect unless paired with \ours\ \textbf{(a)}; under guidance, $h_0$ on frozen RAEv2 achieves the steepest gain, outperforming nonlinear and combined variants \textbf{(b)}.}
  \label{fig:fuselift-xl}
\end{figure}

Freezing the full-fusion RAEv2 decoder keeps clean reconstruction identical ($27.04$\,dB PSNR) while relocating DiT-XL from encoder coordinates ($z$) to the affine input prefix ($h_0$). Under guidance, FuseLift lowers 80-epoch full-fusion gFID from $1.25$ to $1.07$ (\Cref{fig:fuselift-xl}b, \Cref{tab:main-generation}) and sharply reduces six-extractor representation distance $\mathrm{FD}_r^6$ (\Cref{tab:ditxl_all}), matching the guided seven-layer baseline at $4.5$\,dB higher PSNR. Applying the same $h_0$ relocation to the seven-layer decoder pushes guided gFID down to $0.984$, cutting the like-for-like guided gap between 23-layer and 7-layer models roughly in half across both gFID and $\mathrm{FD}_r^6$ (\Cref{tab:ditxl_all}).

Without guidance, $h_0$ relocation barely moves full-fusion gFID and slightly lowers Inception Score (\Cref{fig:fuselift-xl}a, \Cref{tab:main-generation}), because large unguided excursions still trigger off-manifold decoder brittleness. At full scale, \ours\ thus drives unguided gains via off-manifold tolerance, whereas FuseLift narrows the guided gap via coordinate alignment.

\subsection{Comparison with Prior Methods}
\label{sec:system-comparison}

\begin{figure}[t]
  \centering
  \includegraphics[width=0.90\columnwidth]{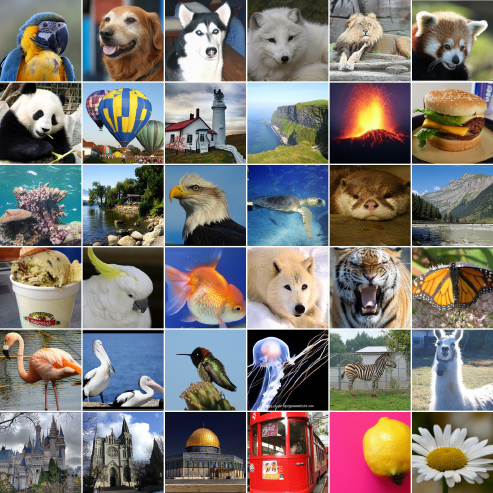}
  \caption{\textbf{Qualitative samples from FuseLift.} Class-conditional generations from diffusing at $h_0$ of the frozen seven-layer decoder ($0.984$ gFID; uncurated class grids in \S\ref{app:generation-samples}).}
  \label{fig:fuselift-mosaic}
\end{figure}

Under the same 80-epoch training and 50-step sampling setup as RAEv2, FuseLift ($h_0$) on the seven-layer decoder reaches $0.984$ guided gFID and $2.14$ $\mathrm{FD}_r^6$ (\Cref{tab:training-efficiency}, samples in \Cref{fig:fuselift-mosaic}). FuseLift improves both gFID and aggregate $\mathrm{FD}_r^6$ over matched RAEv2 at $k{=}23$ and $k{=}7$. The seven-layer configuration also leads the compared pixel-space (JiT-H/16~\citep{li2025jit}, PixelDiT-XL~\citep{yu2026pixeldit}), latent diffusion (SiT-XL/2-REPA~\citep{yu2024repa}, REPA-E~\citep{leng2025repae}, RAE-XL~\citep{zheng2025rae}, SFD-XL~\citep{Pan2025SFD}), and autoregressive (BAR-L~\citep{yu2026bar}) baselines on both metrics.

\section{Ablations and Analysis}
\label{sec:experiments}
\label{sec:h1-results}

\subsection{Guidance and Intervention Interaction}
\label{sec:decoder-target-ablations}

Combining \ours\ and FuseLift outperforms the encoder-space baseline in both unguided and guided generation (\Cref{tab:ditxl_all}), yet each single intervention remains strongest in its target regime. Unlike the orthogonal spiral (\Cref{fig:toy-spiral}), two effects couple them on real encoders. First, partial overlap between cross-layer disagreement directions and high-gain prefix directions in \Cref{eq:interface-error-main} reduces the marginal benefit of suppressing both terms along shared axes. Second, FuseLift inherits its input projection from the frozen decoder: measuring effective dimensionality via the token-covariance participation ratio (\S\ref{sec:lift-spectrum}), training the decoder with \ours\ concentrates $h_0$ variance into substantially fewer active dimensions (\Cref{tab:decoder_layer_sweep}), narrowing the target spectrum modeled by the DiT. Stepping deeper from $z$ to $h_0$ and $h_1$ also monotonically lowers unguided Inception Score at both $k{=}23$ and $k{=}7$ (\Cref{tab:ditxl_all}), reflecting an intrinsic target shift beyond guidance strength.

\subsection{Generator-Side FuseReg Across Model Scales}
\label{sec:exp-grid}

Random layer subsets can also regularize Stage-2 training by corrupting DiT inputs while still predicting the clean all-layer mean (\S\ref{app:fusereg-method}). On 40-epoch unguided DiT-Base, input randomization helps on its own, and applying \ours\ to both stages cuts gFID by $29$--$40\%$ consistently across DINOv3-L, SigLIP2-L, and EUPE-B (\Cref{tab:fusereg-summary}; full grids in \S\ref{app:generator-results}, \S\ref{app:siglip}). At DiT-XL scale, however, decoder-only \ours\ outperforms joint training under both unguided and guided sampling, showing that generator-side subset augmentation primarily aids smaller models under tight training budgets.

\subsection{Where to Place the Diffusion Target}
\label{sec:lift-depth}

\begin{figure}[t]
  \centering
  \includegraphics[width=\columnwidth]{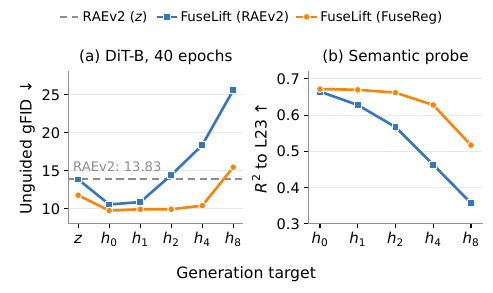}
  \caption{\textbf{One affine step into the decoder is optimal.} Shifting the diffusion target from encoder space ($z$) across frozen decoder depths ($h_0$: affine input; $h_j$: after block $j$), generation peaks at $h_0$ and degrades in nonlinear blocks \textbf{(a)}, tracking linear-probe predictability ($R^2$) of final-layer semantics \textbf{(b)}.}
  \label{fig:fuselift-depth}
\end{figure}

Varying the split depth across the frozen decoder traces a continuous path from encoder space ($z$) toward pixel synthesis at identical clean reconstruction (\Cref{fig:fuselift-depth}). On 40-epoch DiT-Base, stepping once to the affine input layer ($h_0$) yields the strongest unguided and guided generation under both decoders, whereas moving into nonlinear decoder blocks ($h_1$--$h_8$) steadily degrades sample quality (\Cref{fig:fuselift-depth}a, \Cref{tab:ditb_layer_sweep}). Held-out ridge linear probes (\Cref{fig:fuselift-depth}b, \Cref{tab:decoder_layer_sweep}) explain why: linear predictability of final-layer DINOv3 semantics remains nearly intact at $h_0$ yet falls progressively through deeper blocks as alignment shifts toward shallow features. Diffusing at $h_0$ adopts the decoder's first-layer sensitivity metric before nonlinear blocks trade linear semantics for pixel synthesis.

% Queue the ablation tables together for the final main-text page.

\begin{table*}[t]
\centering
\caption{\textbf{Orthogonal crossing of decoder tolerance and diffusion target depth on DiT-XL.}
Crossing standard RAEv2 and \textbf{\ours} decoders with diffusion in encoder space ($\mathbf{z}$), at the frozen decoder's affine input ($\mathbf{h}_0$), or after block 1 ($\mathbf{h}_1$), alongside seven-layer ($k{=}7$) transfer across 20--80 epochs. \ours\ on $\mathbf{z}$ leads unguided full fusion, FuseLift at $\mathbf{h}_0$ on frozen RAEv2 leads converged guided generation at both $k{=}23$ and $k{=}7$, combining both yields sub-additive gains, and nonlinear $\mathbf{h}_1$ consistently trails $\mathbf{h}_0$ (\textbf{bold}: best per block and column).}
\label{tab:ditxl_all}
\setlength{\tabcolsep}{3pt}
\footnotesize
\begin{tabular*}{\textwidth}{@{\extracolsep{\fill}}lll *{12}{c}@{}}
\toprule
 & & & \multicolumn{6}{c}{Unguided} & \multicolumn{6}{c}{Guided} \\
\cmidrule(lr){4-9}\cmidrule(lr){10-15}
 & & & \multicolumn{4}{c}{gFID$\downarrow$} & IS$\uparrow$ & $\mathrm{FD}_r^6\!\downarrow$
     & \multicolumn{4}{c}{gFID$\downarrow$} & IS$\uparrow$ & $\mathrm{FD}_r^6\!\downarrow$ \\
\cmidrule(lr){4-7}\cmidrule(lr){8-9}\cmidrule(lr){10-13}\cmidrule(lr){14-15}
Tokenizer & Decoder & Latent & 20 & 40 & 60 & 80 & 80 & 80 & 20 & 40 & 60 & 80 & 80 & 80 \\
\midrule
\multirow{6}{*}{RAEv2 $k{=}23$}
 & RAEv2   & $\mathbf{z}$ & 3.31   & 2.91 & 2.92   & 3.01 & 207.4     &  6.01    & 2.20 & 1.57 & 1.31 & 1.25 & 256.8 & 3.06 \\
  & FuseReg & $\mathbf{z}$ & 3.01   & \textbf{2.38} & \textbf{2.30} & \textbf{2.21} & \textbf{212.8} & 6.23 & \textbf{1.97} & 1.51 & 1.30 & 1.25 & \textbf{260.1} & 3.49 \\
 & RAEv2   & $\mathbf{h}_0$ & 3.24 & 2.76 & 2.97 & 2.95 & 193.6 & \textbf{5.77} & 2.10 & \textbf{1.34} & \textbf{1.08} & \textbf{1.07} & 256.1 & \textbf{2.30} \\
  & FuseReg & $\mathbf{h}_0$ & 3.21 & 2.59 & 2.68 & 2.67 & 195.1 & 6.40 & 2.04 & 1.37 & 1.18 & 1.18 & 253.8 & 3.21 \\
 & RAEv2   & $\mathbf{h}_1$ & 3.43 & 3.14 & 3.44 & 3.37 & 177.8 & 6.31 & 2.19 & 1.41 & 1.14 & 1.13 & 240.4 & 2.55 \\

 & FuseReg & $\mathbf{h}_1$ & \textbf{2.96} & 2.49 & 2.56 &  2.45 & 201.8 & 5.96 & 2.11 & 1.40 & 1.14 &  1.12 & 259.3 & 3.06 \\
\midrule
\multirow{3}{*}{RAEv2 $k{=}7$}
 & RAEv2   & $\mathbf{z}$   & 2.61 & 1.89 & \textbf{1.67} & \textbf{1.65} & \textbf{226.7} & 4.11 & 2.34 & 1.46 & 1.11 & 1.08 & \textbf{260.0} & 2.19 \\
 & RAEv2   & $\mathbf{h}_0$  & \textbf{2.44} & \textbf{1.88} & 1.78 & 1.76 & 209.2 & \textbf{3.97} & \textbf{2.11} & \textbf{1.32} & \textbf{1.02} & \textbf{0.98} & 253.4 & \textbf{2.14} \\
 & RAEv2   & $\mathbf{h}_1$ & 2.72 & 2.23 &  2.27   & 2.23    & 193.1 & 4.67 & 2.39 & 1.42 & 1.06 & 1.04    & 249.4 & 2.35 \\
\bottomrule
\end{tabular*}
\end{table*}

\begin{table}[t]
  \centering
  \caption{\textbf{Decoder versus generator subset regularization across encoders and model scales (40-epoch gFID$\downarrow$).}
  Comparing standard training (\textbf{Baseline}), subset-trained decoding (\textbf{Decoder-only}), subset-corrupted DiT inputs (\textbf{DiT-only}), and both (\textbf{Joint}; shared generators decode identical samples). Joint regularization helps DiT-B across all encoders, whereas DiT-XL reaches its best unguided and guided quality from decoder tolerance alone.}
  \label{tab:fusereg-summary}
  \footnotesize
  \setlength{\tabcolsep}{3.0pt}
  \renewcommand{\arraystretch}{1.08}
  \begin{tabular*}{\linewidth}{@{\extracolsep{\fill}}l rrr rr@{}}
    \toprule
    & \multicolumn{3}{c}{DiT-B (unguided)}
    & \multicolumn{2}{c}{DiT-XL (DINOv3)} \\
    \cmidrule(lr){2-4}\cmidrule(lr){5-6}
    Setting & DINOv3 & SigLIP2 & EUPE & Unguid. & Guided \\
    \midrule
    Baseline     & 13.96 & 13.77 & 12.17 & 2.91 & 1.57 \\
    Decoder-only & 12.96 &  9.73 &  8.52 & \textbf{2.38} & \textbf{1.51} \\
    DiT-only     & 12.09 & 11.26 & 11.10 & 2.96 & 1.67 \\
    Joint        & \textbf{9.93} & \textbf{8.26} & \textbf{8.00} & 2.42 & 1.58 \\
    \bottomrule
  \end{tabular*}
\end{table}

\section{Discussion}
\label{sec:main-discussion}

\paragraph{Three choices in $k$.}
Choosing fusion depth $k$ simultaneously sets what the latent encodes, how sensitively the decoder reads it, and in which metric the generator models it. Separating these roles lets guided full fusion match the seven-layer baseline selected by $k$-tuning at far sharper reconstruction, while placing latent and pixel diffusion along a single axis across decoder depth (\Cref{fig:fuselift-depth}). Encoder space offers linear semantics in a pixel-agnostic metric, pixel space offers the pixel metric without linear semantics, and one affine step into the decoder keeps the encoder's information while taking the decoder's first-layer metric.

\paragraph{Why guidance separates the two interventions.}
The local error expansion in \Cref{eq:interface-error-main} clarifies why decoder tolerance and coordinate alignment govern opposite sampling regimes. Without guidance, broader sampling errors carry generated latents farther off the clean manifold, where finite-displacement training across layer subsets (\ours) prevents decoder collapse. Under internal guidance, trajectories concentrate near the clean manifold, where small residual errors span all encoder channels beyond the low-rank disagreement subspace and make full-channel prefix reweighting (FuseLift) decisive. Together, these information-preserving interventions establish a constructive lower bound on how much of the reconstruction--generation trade-off stems from interface geometry.

\section{Limitations and Future Work}
\label{sec:limitations}

First, \ours\ regularizes the decoder only through perturbations formed by subsets of existing frozen encoder layers, which span at most $K{-}1$ disagreement directions per token rather than expanding tolerance across the full latent space. Consequently, subset training yields its largest gains when generator errors remain broad (under unguided sampling, early checkpoints, or smaller generators), whereas once guided trajectories converge near the clean manifold, residual errors lie largely outside that subspace. Developing broader, structure-aware regularizers or explicit Jacobian penalties that widen decoder tolerance across arbitrary off-manifold directions is a direct direction for future work. Second, while our geometry-only interventions let guided full fusion match the encoder-space seven-layer baseline at far higher reconstruction fidelity, they do not completely close the like-for-like gap across fusion depths: seven-layer fusion retains an advantage unguided and a narrow lead when both tokenizers are lifted to $h_0$. Eliminating that residual gap may require co-adapting encoder representations or capturing nonlinear suffix curvature beyond the first affine projection. Finally, our study focuses on class-conditional ImageNet-256 under a fixed guidance recipe. Extending decoder-aligned diffusion geometry to text-to-image synthesis, higher resolutions, and video tokenizers~\citep{guo2026vrae,xie2026videorae} is a natural next step.

\section{Conclusion}
\label{sec:conclusion}

Incorporating detail-rich shallow layers into Representation Autoencoders sharpens pixel reconstruction yet has long appeared to penalize diffusion generation. By holding encoded information strictly fixed, we showed that much of this penalty stems not from the modeling difficulty of visual detail, but from geometric mismatch across the frozen encoder--decoder interface. Flattening off-manifold decoder sensitivity along cross-layer disagreements (\ours) recovers visual structure already present in unguided samples, while relocating diffusion one affine step into the frozen decoder (FuseLift) aligns noise injection and loss weighting with pixel-decoder sensitivity at identical clean reconstruction, matching shallower-fusion generation at full-fusion fidelity and improving guided quality across fusion depths. Decoupling what a tokenizer encodes from the geometry in which a generator diffuses offers a general principle for visual generation: take \emph{information} from the encoder and \emph{geometry} from the decoder.

\FloatBarrier

{
    \small
    \bibliographystyle{ieeenat_fullname}
    \bibliography{references}
}

% ---------------------------------------------------------------------
% Supplementary material is included in the complete arXiv PDF.
% ---------------------------------------------------------------------
% Supplementary material: single-column layout (allowed by CVPR for the supplement).
\onecolumn
\hypersetup{pageanchor=false}
\setcounter{page}{1}
\appendix
\setcounter{topnumber}{3}
\setcounter{bottomnumber}{2}
\setcounter{totalnumber}{5}
\setcounter{dbltopnumber}{2}
\widowpenalty=150
\clubpenalty=150
\raggedbottom
\setcounter{secnumdepth}{2}
% Number appendix figures and tables by section, with distinct hyperlink targets.
\counterwithin{figure}{section}
\renewcommand{\theHfigure}{appendix.\thesection.\arabic{figure}}
\counterwithin{table}{section}
\renewcommand{\theHtable}{appendix.\thesection.\arabic{table}}
\makeatletter
\renewcommand*\l@section{\@dottedtocline{1}{0em}{1.8em}}
\renewcommand*\l@subsection{\@dottedtocline{2}{1.8em}{2.5em}}
\makeatother
\etocdepthtag.toc{mtappendix}
\etocsettagdepth{mtmain}{none}
\etocsettagdepth{mtappendix}{subsection}
\etocsettocstyle{\section*{Appendix Contents}\noindent\rule{\linewidth}{0.4pt}\par\vspace{2pt}}{\noindent\rule{\linewidth}{0.4pt}\par}
% Open the supplement with the index, followed by three uncurated sample pages.
\noindent\begin{minipage}{\linewidth}
\begin{center}
  {\Large\bfseries \thetitle\par}
  \vspace{0.5em}
  {\large Supplementary Material\par}
  \vspace{1em}
\end{center}
\small
\tableofcontents
\end{minipage}
\medskip

\section{Uncurated Generation Samples}
\label{app:generation-samples}

\begingroup
\setlength{\intextsep}{5pt}
\begin{figure}[H]
  \centering
  \includegraphics[width=\linewidth]{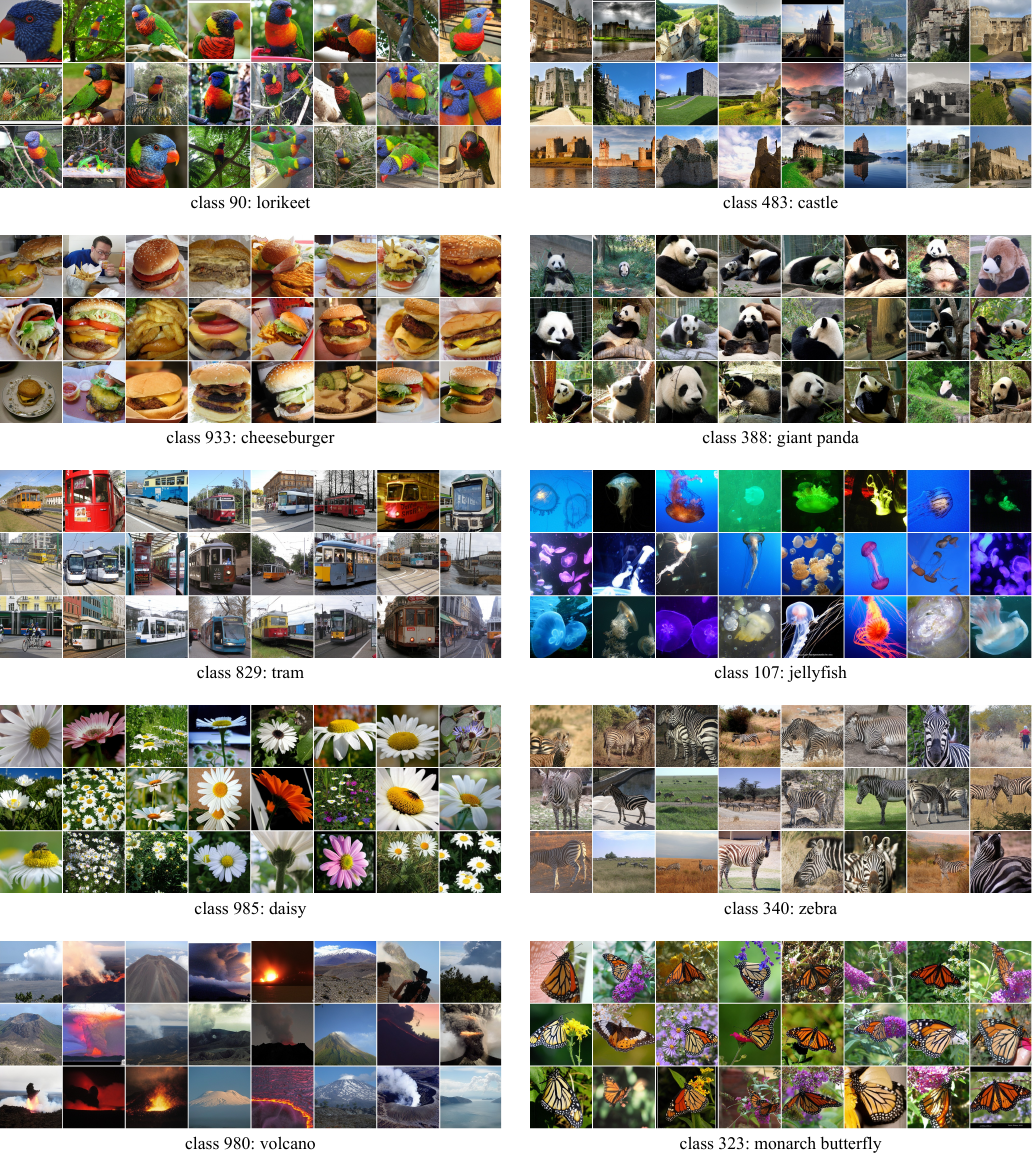}
  \caption{\textbf{Uncurated ImageNet $256\times256$ samples from FuseLift (1/3).}
  Classes are mixed across pages; each shows the first 24 images from the 50k evaluation set in original sampling order (80-epoch DiT-XL at $h_0$ of the frozen 7-layer RAEv2 decoder, EMA, 50 Euler steps, internal guidance $1.78$, seed 1).}
  \label{fig:uncurated-animals}
\end{figure}

\begin{figure}[H]
  \centering
  \includegraphics[width=\linewidth]{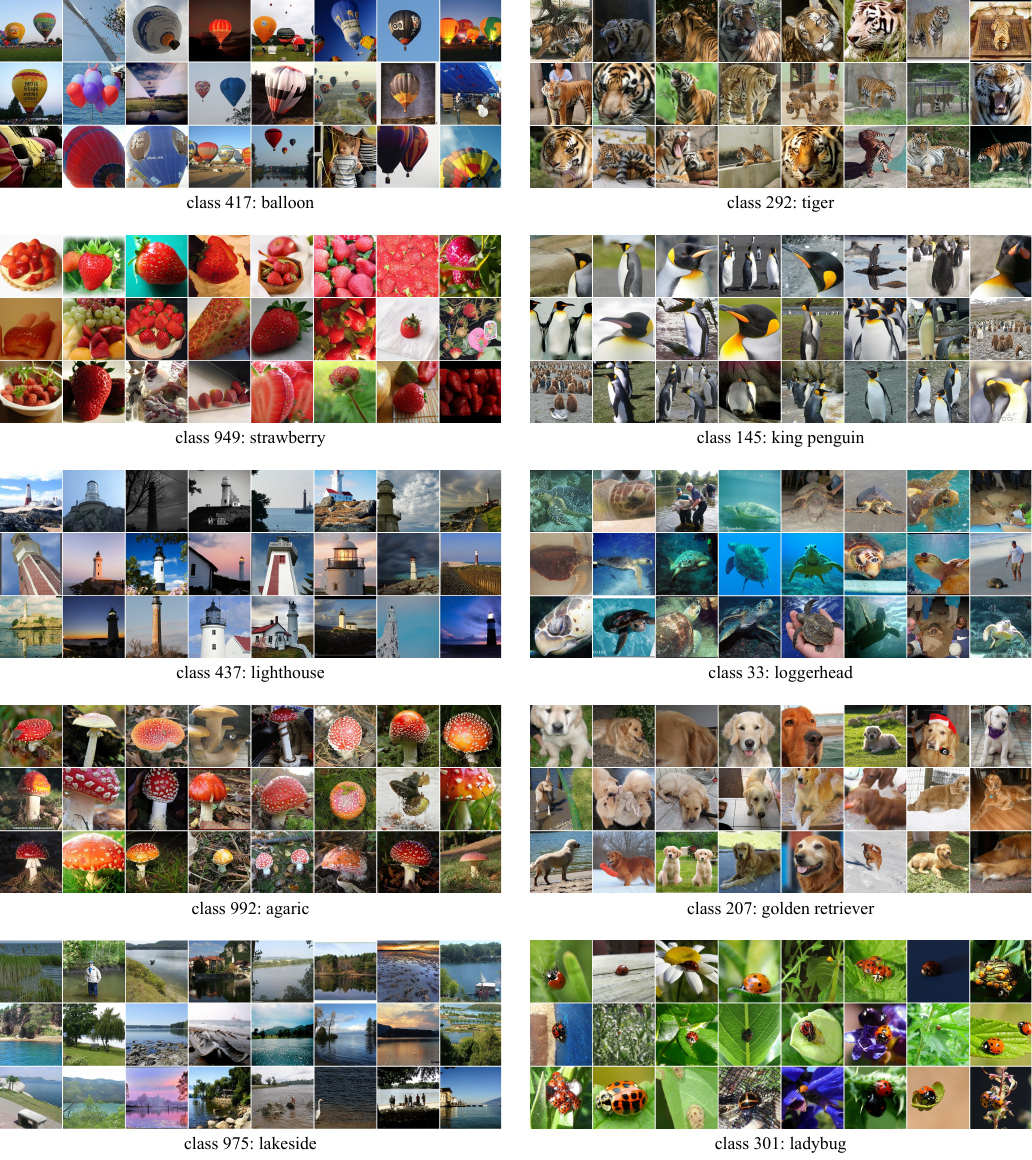}
  \caption{\textbf{Uncurated ImageNet $256\times256$ samples from FuseLift (2/3).}
  Each class shows 24 samples in an $8\times3$ grid with its class name and ImageNet ID below, following the sampling setup and within-class ordering of \Cref{fig:uncurated-animals}.}
  \label{fig:uncurated-mixed}
\end{figure}

\begin{figure}[H]
  \centering
  \includegraphics[width=\linewidth]{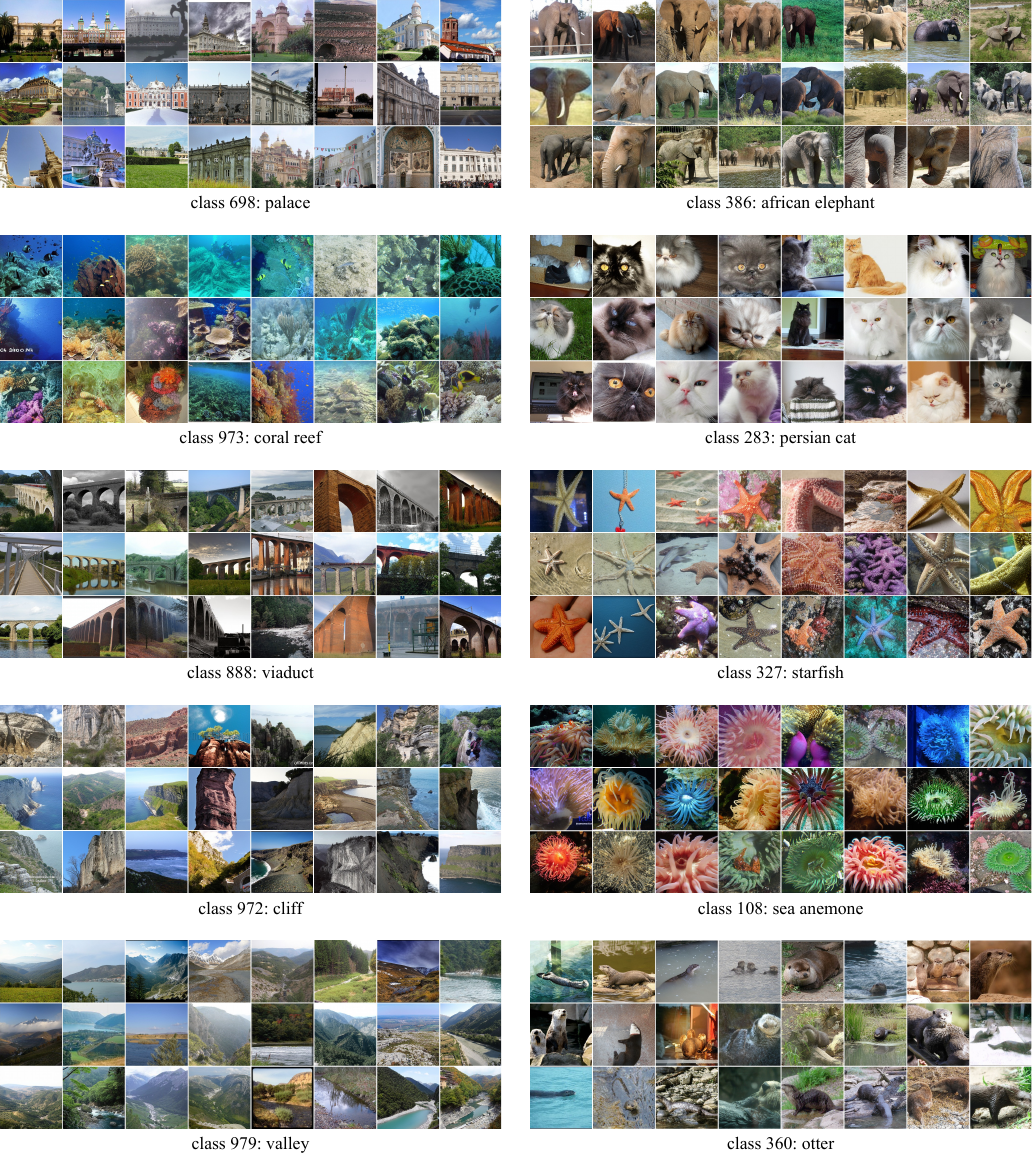}
  \caption{\textbf{Uncurated ImageNet $256\times256$ samples from FuseLift (3/3).}
  Each class shows 24 samples with its class name and ImageNet ID below, following the sampling setup and within-class ordering of \Cref{fig:uncurated-animals}.}
  \label{fig:uncurated-scenes}
\end{figure}

\endgroup

% Empirical results and samples precede the detailed methods and proofs.
\Needspace{8\baselineskip}
\section{Extended Related Work}
\label{app:related-extended}

We organize prior work by the component it changes. The first group develops
representation autoencoders and reconstruction-oriented tokenizers; the second changes
how representations are aligned, combined across depth, or sampled during training.

\subsection{Representation autoencoders and tokenizer design}
\label{app:rw-tokenizers}

\paragraph{Image representation autoencoders.}
Latent diffusion~\citep{rombach2022ldm} and diffusion
transformers~\citep{peebles2023dit} standardized training a diffusion model
inside the latent space of a separately trained autoencoder rather than on
raw pixels. In the standard pipeline, the tokenizer is a reconstruction-centric
VAE. RAE~\citep{zheng2025rae} instead pairs a frozen self-supervised vision encoder
with a learned decoder, producing a semantically structured latent that remains
reconstructable. RAEv2~\citep{singh2026raev2} fuses the last $k$ encoder layers and
shows that $k$ controls a reconstruction--generation Pareto
frontier. Our work takes this result as its starting point: a fixed-$k$ fusion asks one
latent to serve two consumers with different preferences.

\paragraph{Video extensions.}
Recent work carries representation autoencoding beyond images. VideoRAE~\citep{xie2026videorae} extracts
multi-scale hierarchical features from frozen video foundation models and compresses
them with a lightweight self-attention projector, producing continuous latents for
diffusion transformers and discrete tokens for autoregressive
models. V-RAE constructs compact video latents from frozen
vision-foundation features using temporal pooling, and evaluates them for
reconstruction, semantic probing, generation, and future
prediction~\citep{guo2026vrae}. These studies extend the scope of RAEs to temporally
structured data; their focus is video compression and temporal modeling, whereas \ours\
targets robustness to alternative layer fusions from a fixed image encoder.

\paragraph{Reconstruction-focused adaptation and efficiency.}
Several works reduce the reconstruction gap by modifying the encoder, latent geometry,
or training recipe. DINO-Tok~\citep{jia2025dinotok} adapts DINO for visual tokenization;
VFM-VAE~\citep{bi2025vfmvae} and
AlignTok~\citep{chen2025aligntok} reshape or align foundation-model features for
reconstruction and generation; and RPiAE~\citep{gong2026rpiae} pivots the representation
for both generation and editing. The Prism Hypothesis~\citep{fan2025prism} and
PS-VAE~\citep{zhang2025psvae} directly study the tension between semantic content and
pixel fidelity. LV-RAE~\citep{liu2026improvingrecon} analyzes off-manifold decoder
sensitivity and uses latent-noise augmentation to improve decoder robustness,
whereas FlatDINO~\citep{calvo2026laminating} compresses the
spatial representation into a much shorter token sequence for efficient diffusion.
These directions are complementary to \ours, which freezes the encoder and trains one
decoder across a broad distribution of sampled layer subsets.

\subsection{Representation alignment and guidance}
\label{app:rw-alignment}

\paragraph{Representation alignment and internal guidance.}
REPA~\citep{yu2024repa} distills a pretrained encoder's representation into
an intermediate DiT layer via a projection head; REPA-E~\citep{leng2025repae}
back-propagates this alignment loss into the tokenizer itself, and
\citet{singh2025whatmatters} isolate spatial structure, rather than global semantic
content, as an important ingredient. RAEv2 finds that REPA complements RAE by improving
intermediate spatial structure~\citep{singh2026raev2}, whereas
PixelREPA~\citep{shin2026pixelrepa} shows that direct REPA can fail in pixel space.
Together, these results indicate that the benefit of alignment depends on the
architecture and the space in which alignment is applied. AutoGuidance uses a separately trained
degraded model~\citep{karras2024autoguidance}; internal guidance instead reuses
intermediate features from the same diffusion model~\citep{singh2026raev2,zhou2026internal}.
Both motivate making the decoder robust to the latent fusion presented downstream.

\Needspace{10\baselineskip}
\subsection{Layer-fusion learning and stochastic configuration}
\label{app:rw-readouts}

\paragraph{Learned multi-layer fusion.}
Deeply supervised networks attach prediction heads at multiple
depths~\citep{lee2015deeplysupervised}. More directly, two contemporaneous papers
study which frozen encoder layers to use and how to combine them.
DRoRAE~\citep{zhu2026beyondlast} introduces a lightweight depth-routed fusion
module with layer-wise expert MLPs and energy-constrained adaptive routing across
encoder layers, trained in three phases: training a decoder, aligning the fusion
module with the frozen decoder, and fine-tuning the decoder; it reports gains in both rFID and
gFID over a last-layer RAE baseline. Attentive multi-layer
fusion~\citep{ciernik2026multilayer} instead learns a dynamic attention mechanism
across ViT layers for downstream representation probing rather than generation.
Both confirm our central premise that useful information is spread unevenly
across depth and a single final layer underuses it. However, both rely on learned
aggregation modules that modify the fused representation itself, whereas \ours\
and FuseLift keep the parameter-free full-layer mean fixed at inference and
address interface geometry.

\paragraph{Stochastic configuration training.}
Dropout~\citep{srivastava2014dropout} originally regularizes by randomly
zeroing units; nested (ordered)
dropout~\citep{rippel2014learning} instead retains a randomly sized prefix by
dropping trailing units to recover an ordered, PCA-like basis in a linear
autoencoder. Our analysis in \S\ref{app:theory} builds on this mechanism and
contrasts its ordering with permutation symmetry. Stochastic
depth~\citep{huang2016stochasticdepth} randomly skips residual blocks during
training and uses the full network at test time. LayerDrop~\citep{fan2019layerdrop}
drops Transformer layers during training to enable shallower inference
sub-networks without retraining. Slimmable
networks~\citep{yu2018slimmable,yu2019universally} support multiple channel
widths, while Matryoshka Representation
Learning~\citep{kusupati2022matryoshka} trains embeddings that stay accurate
under nested dimensionality truncations. LayerDrop, slimmable networks, and
Matryoshka representations offer deployment trade-offs through reduced depth,
width, or embedding dimension. \ours\ applies layer dropout to the choice of frozen encoder layers that
form the latent for the decoder or generator, placing every nonempty layer subset
in the training support. The probability-weighted guarantee in \Cref{cor:coverage} and the empirical
subset analysis in \S\ref{app:subset-diagnostics} characterize the fusion
robustness.

\section{Experimental Setup and Evaluation}
\label{app:experimental-protocol}

\subsection{Encoder, layer fusions, and evaluation metrics}
\label{app:eval-setting}

The primary experiments and the supplementary reconstruction and generation results
in \Cref{app:decoder-results,app:generator-results} use a frozen, pretrained
DINOv3-L encoder and ImageNet at $256{\times}256$ resolution. Experiments with
other pretrained encoder families are presented separately in \S\ref{app:siglip}.
The candidate pool contains $K{=}23$ transformer layers. The main
reconstruction comparisons use the full $k{=}23$ fusion, the $k{=}7$ fusion over
layers $\{\ell_{11}, \ell_{13}, \ldots, \ell_{23}\}$, and the single layer
$\ell_{11}$. \ours\ normalizes each sampled fusion by its number of retained
layers (\Cref{eq:drop}); fixed-fusion baselines retain their stated scale
conventions.

We evaluate reconstruction with PSNR, SSIM, VGG LPIPS (on $[-1,1]$ inputs following RAEv2 offline evaluation), and rFID on 50k images, and generation
with gFID and IS over 50k class-conditional samples using 50 Euler steps in our controlled RAE experiments. External methods use the sampling configurations in \Cref{tab:sampling-protocols}. Fixed-fusion RAEv2 baseline decoders follow RAEv2 in applying per-sample additive Gaussian latent noise ($\sigma\sim\mathcal{U}(0,0.8)$), whereas \ours\ decoders add no latent noise. The
per-layer diagnostic in \Cref{fig:layer-usage-k23} uses the same 10,000 held-out
images for both decoders and retains RAEv2's fixed $\ell_{23}$ token-mean surrogate
in every probe.

\subsection{Training budgets and regularization grids}
\label{app:training-grids}

Decoder sweeps train a ViT decoder from scratch for 16 epochs with the pixel,
perceptual, adversarial, and classifier-surrogate terms. Architecture, parameter
count, optimization settings, and training budget are matched across reproduced
decoder variants. Controlled perturbation and multi-fusion ablations (\Cref{tab:perturbation-ablation}) train 5-epoch DINOv3-L decoders (batch size 256, MSE + LPIPS, GAN from epoch 3) and evaluate on 10k validation images without stochastic augmentation. The DiT rate grids use 40 training epochs with the same controls within
each generator scale (DiT-XL hidden size 1440), while the guided DiT-XL convergence comparison (\Cref{tab:ditxl_all}) trains for 80 epochs and evaluates checkpoints at 20, 40, 60, and 80 epochs. \Cref{tab:appendix-protocol}
summarizes the main rate sweeps. DiT-Base is evaluated without guidance; DiT-XL is evaluated
both without guidance and with internal guidance at scale $1.78$.

\begin{table}[H]
  \centering
  \caption{\textbf{Training budgets and \ours\ rate grids.}
  Decoder and generator rates are varied independently. In the generator studies,
  each trained DiT is evaluated with the corresponding decoder-rate grid.}
  \label{tab:appendix-protocol}
  \small
  \setlength{\tabcolsep}{5pt}
  \begin{tabular}{lcl}
    \toprule
    Component & Epochs & Configuration / Drop rates \\
    \midrule
    Decoder (DINOv3-L, $K{=}23$) & 16 & $p_{\mathrm{dec}}\in\{0,0.05,0.1,0.3,0.5,0.7,0.9,0.95\}$ \\
    Perturbation \& fusion controls & 5 & Energy-matched controls, multi-fusion, \ours\ ($p_{\mathrm{dec}}{=}0.95$) \\
    DiT-Base (DINOv3-L, $K{=}23$) & 40 & $p_{\mathrm{dit}},p_{\mathrm{dec}}\in\{0,0.05,0.1,0.3,0.5,0.7,0.9\}$ \\
    DiT-Base (SigLIP2-L, EUPE-B) & 40 & $p_{\mathrm{dit}},p_{\mathrm{dec}}\in\{0,0.3,0.6,0.9\}$ \\
    DiT-XL sweep (DINOv3-L, $K{=}23$) & 40 & $p_{\mathrm{dit}}\in\{0,0.5,0.7,0.9\}$; decoder rates as in row 1 \\
    DiT-XL trajectory (DINOv3-L, $K{=}23$) & 80 & Checkpoints at $20,40,60,80$ ep; $p_{\mathrm{dit}}{=}0$, $p_{\mathrm{dec}}\in\{0,0.95\}$ \\
    \bottomrule
  \end{tabular}
\end{table}
\FloatBarrier

\subsection{Sampling configurations for compared methods}
\label{app:sampling-protocols}

\Cref{tab:sampling-protocols} records the checkpoints, sampling algorithms, step budgets, and guidance settings used for the comparisons in \Cref{tab:main-generation,tab:training-efficiency}. The controlled RAEv2, \ours, and FuseLift runs use the same 50-step Euler sampler and internal guidance schedule.

% Generated from tables/src/sampling_protocols.json.
\begin{table}[H]
\centering
\caption{\textbf{Checkpoints and sampling configurations.}
ODE uses Euler unless specified; SDE uses Euler--Maruyama. BAR-L's $256{\times}4$ denotes 256 autoregressive tokens with 4 MBM decoding steps per token.
Guidance intervals retain each sampler's time convention; SFD-XL and PixelDiT-XL guidance settings follow their released configurations.}
\label{tab:sampling-protocols}
\small
\setlength{\tabcolsep}{4pt}
\renewcommand{\arraystretch}{1.08}
\newlength{\protocolwidth}
\setlength{\protocolwidth}{\dimexpr\linewidth-6\tabcolsep\relax}
\begin{tabular}{@{}>{\raggedright\arraybackslash}p{0.27\protocolwidth}>{\raggedright\arraybackslash}p{0.15\protocolwidth}>{\raggedright\arraybackslash}p{0.27\protocolwidth}>{\raggedright\arraybackslash}p{0.31\protocolwidth}@{}}
\toprule
Model / checkpoint & Training / sampling budget & Sampler configuration & Guidance \\
\midrule
\textbf{BAR-L}\newline BAR-L + BAR-FSQ 16-bit (ft) & $\sim\!400$ epochs\newline $256{\times}4$ sampling steps & AR/MBM: Raster order over 256 tokens; 4 MBM decoding steps per token; bit allocation $[2,2,5,7]$. & CFG increases linearly from $1$ to $5.3$ along the sequence; temperature $3.0$. \\
\addlinespace[4pt]
\textbf{REPA-E}\newline SiT-XL/1 + E2E-VAVAE & 800 epochs\newline 250 sampling steps & SDE: Euler--Maruyama. & CFG $2.4$ on $t\in[0,0.73]$. \\
\addlinespace[4pt]
\textbf{RAE-XL}\newline DiT$^{\mathrm{DH}}$-XL + DINOv2-B; ViT-XL decoder & 800 epochs\newline 50 sampling steps & ODE: Euler, fp32. & AutoGuidance $1.42$; DiT$^{\mathrm{DH}}$-S guidance model trained for 14 epochs. \\
\addlinespace[4pt]
\textbf{LightningDiT}\newline LightningDiT-XL/1 + VA-VAE f16d32 & 800 epochs\newline 250 sampling steps & ODE: Euler; timestep shift $0.3$. & CFG $6.7$; guidance interval starts at $0.125$. \\
\addlinespace[4pt]
\textbf{DDT-XL}\newline DDT-XL/2; 22 encoder + 6 decoder layers & 800 epochs (4M train steps)\newline 250 sampling steps & ODE: Euler; time shift $1.0$; last step $0.04$. & CFG $2.0$ on $t\in[0.3,1.0]$. \\
\addlinespace[4pt]
\textbf{SiT-XL/2-REPA}\newline SiT-XL/2 + REPA & 1400 epochs\newline 250 sampling steps & SDE: Euler--Maruyama. & CFG $1.8$ on $t\in[0,0.7]$. \\
\addlinespace[4pt]
\textbf{REG}\newline SiT-XL/2 + REG & 800 epochs (4M train steps)\newline 250 sampling steps & SDE: Euler--Maruyama. & CFG $2.3$; class-token CFG $2.3$. \\
\addlinespace[4pt]
\textbf{SiT-XL/2}\newline SiT-XL/2 & 1400 epochs (7M train steps)\newline 250 sampling steps & SDE: Euler--Maruyama; last step $0.04$. & CFG $1.5$, applied only to the first 3 latent channels. \\
\addlinespace[4pt]
\textbf{JiT-H/16}\newline JiT-H/16; pixel-space diffusion & 600 epochs\newline 50 sampling steps & ODE: Heun. & CFG $2.2$ on $t\in[0.1,1.0]$. \\
\addlinespace[4pt]
\textbf{PixelDiT-XL}\newline PixelDiT-XL; ImageNet-256 & 320 epochs (1.6M train steps)\newline 100 sampling steps & ODE: Second-order multistep FlowDPM solver. & CFG $2.75$ on $t\in[0.1,0.9]$; time shift $1.0$. \\
\addlinespace[4pt]
\textbf{SFD-XL}\newline SFD-XL; ImageNet-256 & 800 epochs (4M train steps)\newline adaptive sampling steps & ODE: dopri5; adaptive step size, atol $10^{-6}$, rtol $10^{-3}$. & AutoGuidance $1.5$; LightningDiT-B guidance model at 70K steps. \\
\addlinespace[4pt]
\midrule
\textbf{RAEv2}\newline DiT-XL + DINOv3-L ($k{=}7$); encoder-space target $z$; EMA & 80 epochs\newline 50 sampling steps & ODE: Euler. & Internal guidance $1.78$ on $t\in[0.1,1]$. \\
\addlinespace[4pt]
\rowcolor{fuseregcolor}
\textbf{FuseReg}\newline DiT-XL + DINOv3-L ($k{=}23$); encoder-space target $z$; EMA & 80 epochs\newline 50 sampling steps & ODE: Euler. & Internal guidance $1.78$ on $t\in[0.1,1]$. \\
\addlinespace[4pt]
\rowcolor{ourscolor}
\textbf{FuseLift}\newline DiT-XL + DINOv3-L ($k\in\{7,23\}$); RAEv2 affine target $h_0$; EMA & 80 epochs\newline 50 sampling steps & ODE: Euler. & Internal guidance $1.78$ on $t\in[0.1,1]$. \\
\bottomrule
\end{tabular}
\end{table}

\FloatBarrier

\section{Supplementary Reconstruction Results}
\label{app:supplementary-results}
\label{app:decoder-results}

All reconstruction results in this section use a frozen, pretrained DINOv3-L encoder.

% Full-width (table*) so each fusion can report PSNR / SSIM / LPIPS / rFID.
% LPIPS: VGG LPIPS from the RAEv2 offline eval, inputs in [-1,1], mean over 50k images.
\begin{table}[H]
  \centering
  \caption{\textbf{One \ours\ decoder supports full, sparse, and single-layer fusions without retraining.}
  Evaluated on 50k ImageNet-256 validation images with frozen DINOv3-L ($K{=}23$).
  Light-gray cells mark each RAEv2 decoder's deterministic training aggregate
  ($k{=}7$ or $k{=}23$); the orange \ours\ row ($p_{\mathrm{dec}}{=}0.95$)
  trains across nonempty subsets of all $K{=}23$ layers. Fixed-fusion decoders obtain lower rFID on
  their exact training fusion, while \ours\ achieves higher paired fidelity (PSNR, SSIM, LPIPS) across
  all three inputs and prevents off-support degradation.}
  \label{tab:recon-main}
  \small
  \setlength{\tabcolsep}{4pt}
  \settowidth{\reconlab}{\small \ours$_{p{=}.95}$}
  \begin{tabular}{@{}p{\reconlab}*{12}{Q{\dimexpr(\textwidth-\reconlab-24\tabcolsep)/12\relax}}@{}}
    \toprule
    & \multicolumn{4}{c}{7-layer subset ($k{=}7$)} & \multicolumn{4}{c}{Full fusion ($K{=}23$)} & \multicolumn{4}{c}{Single layer ($\ell_{11}$)} \\
    \cmidrule(lr){2-5}\cmidrule(lr){6-9}\cmidrule(lr){10-13}
    Decoder & PSNR$\uparrow$ & SSIM$\uparrow$ & LPIPS$\downarrow$ & rFID$\downarrow$ & PSNR$\uparrow$ & SSIM$\uparrow$ & LPIPS$\downarrow$ & rFID$\downarrow$ & PSNR$\uparrow$ & SSIM$\uparrow$ & LPIPS$\downarrow$ & rFID$\downarrow$ \\
    \midrule
    RAEv2$_{K{=}7}$  & \cellcolor{gray!12}22.58 & \cellcolor{gray!12}0.626 & \cellcolor{gray!12}0.153 & \cellcolor{gray!12}\textbf{0.30} & 18.37 & 0.531 & 0.211 & 1.62 & 17.50 & 0.491 & 0.281 & 6.35 \\
    RAEv2$_{K{=}23}$ & 12.51 & 0.369 & 0.517 & 16.10 & \cellcolor{gray!12}27.04 & \cellcolor{gray!12}0.806 & \cellcolor{gray!12}0.091 & \cellcolor{gray!12}\textbf{0.18} & 14.31 & 0.447 & 0.384 & 11.21 \\
    \rowcolor{fuseregcolor}
    \ours$_{p{=}.95}$ & \textbf{23.77} & \textbf{0.678} & \textbf{0.144} & 0.60 & \textbf{27.52} & \textbf{0.826} & \textbf{0.088} & 0.42 & \textbf{25.13} & \textbf{0.735} & \textbf{0.115} & \textbf{0.45} \\
    \bottomrule
  \end{tabular}
\end{table}

\subsection{Decoder-rate selection across layer fusions}
\label{app:decoder-rate}

\begin{table}[H]
  \centering
  \caption{\textbf{\ours\ controls decoder robustness across layer fusions.}
  Each row is one decoder trained for 16 epochs with rate $p_{\mathrm{dec}}$ and
  evaluated on 50k ImageNet-256 images using three input fusions. All runs retain
  the GAN and classifier-surrogate terms. The $p_{\mathrm{dec}}{=}0$ row is our
  reproduced full-fusion baseline. \textbf{Bold}/\underline{underline}: best/second-best
  within each metric and fusion.}
  \label{tab:drop-sweep}
  \small
  \setlength{\tabcolsep}{2.5pt}
  \newlength{\dropsweeplabelwidth}
  \settowidth{\dropsweeplabelwidth}{$0$ (RAEv2)}
  \begin{tabular}{@{}>{\raggedright\arraybackslash}p{\dropsweeplabelwidth}*{9}{Q{\dimexpr(\textwidth-\dropsweeplabelwidth-18\tabcolsep)/9\relax}}@{}}
  \toprule
  & \multicolumn{3}{c}{Fusion $k{=}7$} & \multicolumn{3}{c}{Fusion $k{=}23$} & \multicolumn{3}{c}{Single layer $\ell_{11}$} \\
  \cmidrule(lr){2-4} \cmidrule(lr){5-7} \cmidrule(lr){8-10}
  $p_{\mathrm{dec}}$ & \textbf{PSNR}~$\uparrow$ & \textbf{SSIM}~$\uparrow$ & \textbf{rFID}~$\downarrow$
                   & \textbf{PSNR}~$\uparrow$ & \textbf{SSIM}~$\uparrow$ & \textbf{rFID}~$\downarrow$
                   & \textbf{PSNR}~$\uparrow$ & \textbf{SSIM}~$\uparrow$ & \textbf{rFID}~$\downarrow$ \\
  \midrule
  \rowcolor{gray!12}
  $0$ (RAEv2) & 12.54 & 0.372 & 16.108 & 27.10 & 0.808 & \textbf{0.189} & 15.81 & 0.503 & 3.392 \\
  $0.05$ & 19.78 & 0.518 & 2.826 & \textbf{28.59} & \textbf{0.853} & \underline{0.273} & 20.76 & 0.557 & 1.881 \\
  $0.1$  & 20.52 & 0.550 & 1.718 & \underline{28.58} & \underline{0.852} & 0.285 & 21.33 & 0.586 & 1.209 \\
  $0.3$  & 22.05 & 0.612 & 0.762 & 28.48 & 0.849 & 0.294 & 22.90 & 0.657 & 0.576 \\
  $0.5$  & 22.93 & 0.646 & 0.655 & 28.43 & 0.848 & 0.294 & 23.98 & 0.695 & 0.522 \\
  $0.7$  & 23.47 & 0.665 & 0.634 & 28.20 & 0.842 & 0.322 & 24.58 & 0.714 & 0.493 \\
  $0.9$  & \underline{23.62} & \underline{0.671} & \underline{0.610} & 27.60 & 0.827 & 0.415 & \underline{24.82} & \underline{0.722} & \underline{0.458} \\
  $0.95$ & \textbf{23.77} & \textbf{0.678} & \textbf{0.604} & 27.52 & 0.826 & 0.421 & \textbf{25.13} & \textbf{0.735} & \textbf{0.455} \\
  \bottomrule
  \end{tabular}
  \end{table}

\Cref{tab:drop-sweep} separates full-fusion fidelity from robustness to reduced
layer inputs. Full-fusion PSNR peaks at $p_{\mathrm{dec}}{=}0.05$, while the
$k{=}7$ and $\ell_{11}$ inputs improve throughout the sweep. At
$p_{\mathrm{dec}}{=}0.95$, one \ours\ decoder reaches $23.77$, $27.52$, and $25.13$\,dB
across the three fusions. This is the checkpoint used in \Cref{tab:recon-main}.
The nearby $p_{\mathrm{dec}}{=}0.9$ setting retains slightly higher full-fusion
PSNR, illustrating how the rate can be selected for the intended mix of inputs.
\FloatBarrier

\subsection{Qualitative reconstruction under changed fusions}
\label{app:qualitative}

\begin{figure}[H]
  \centering
  \includegraphics[width=\textwidth]{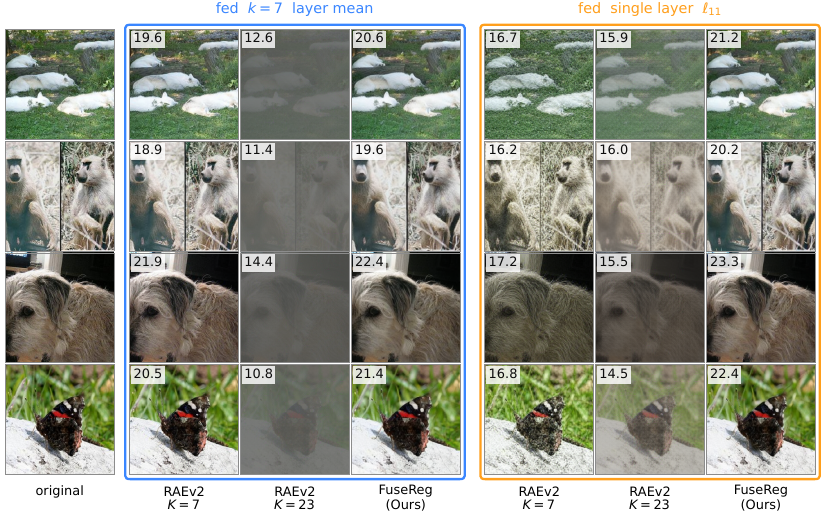}
  \caption{\textbf{One \ours\ decoder reconstructs both 7-layer and single-layer inputs.}
  Each row tests one image under the last 7 encoder layers ($k{=}7$, blue) and single layer $\ell_{11}$ (orange), with per-image PSNR (dB) insets. Standard RAEv2 decoders trained on a fixed 7-layer or 23-layer average break down whenever the test input differs from their training fusion, whereas one \ours\ decoder preserves image structure across both.}
  \label{fig:feed-readouts}
\end{figure}

\Cref{fig:feed-readouts} extends the feature-map comparison in \Cref{fig:cosine-robust} across four held-out ImageNet-256 validation
images under two fusions that differ from the full $k{=}23$ input. Under the $k{=}7$ fusion, the RAEv2$_{K=23}$ decoder collapses to low-contrast,
nearly uniform outputs (10.8--14.4\,dB), whereas \ours\ recovers object
structure and color and slightly exceeds the RAEv2$_{K=7}$ decoder trained on
this exact fusion (19.6--22.4 vs.\ 18.9--21.9\,dB). Under the single layer
$\ell_{11}$, both fixed-fusion baselines are evaluated off their training input
and degrade to 14.5--17.2\,dB, while the same \ours\ decoder remains at
20.2--23.3\,dB. These examples follow the aggregate PSNR trend in
\Cref{tab:recon-main}: one \ours\ decoder handles both a sparse
and a single-layer fusion without retraining.

\FloatBarrier

\subsection{Layer Reliance and Structured Perturbations}
\label{app:subset-diagnostics}
\label{sec:exp-recon}

\begin{figure}[H]
  \centering
  \includegraphics[width=\textwidth]{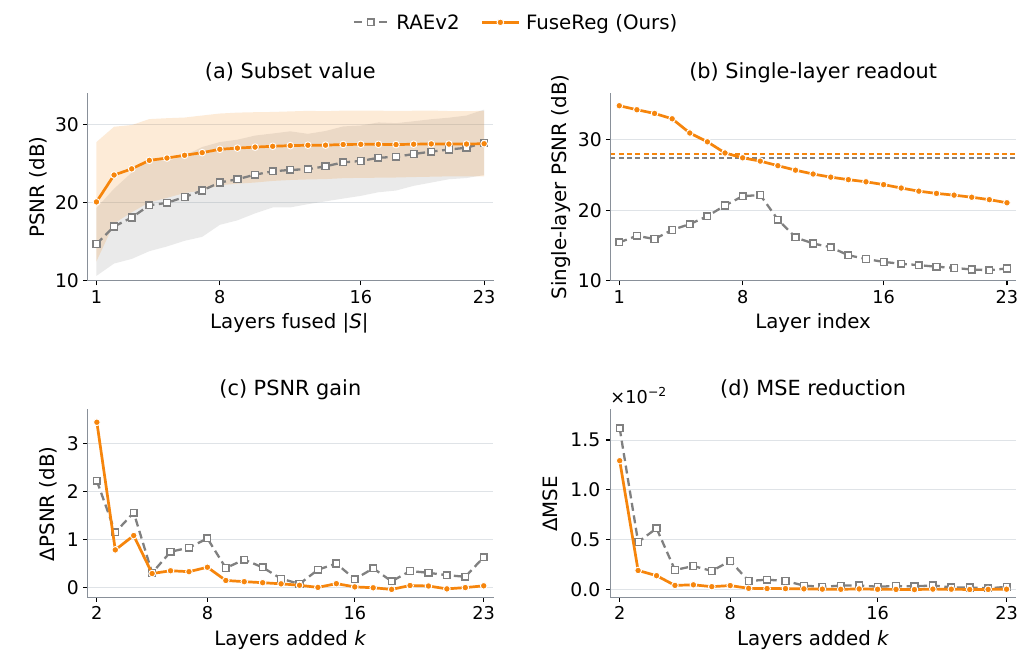}

  \begin{subfigure}[t]{0.24\textwidth}
    \centering
    \captionsetup{font=footnotesize}
    \phantomcaption
    \label{fig:sweep-value}
  \end{subfigure}\hfill
  \begin{subfigure}[t]{0.24\textwidth}
    \centering
    \captionsetup{font=footnotesize}
    \phantomcaption
    \label{fig:single-layer-readout}
  \end{subfigure}\hfill
  \begin{subfigure}[t]{0.24\textwidth}
    \centering
    \captionsetup{font=footnotesize}
    \phantomcaption
    \label{fig:sweep-marginal}
  \end{subfigure}\hfill
  \begin{subfigure}[t]{0.24\textwidth}
    \centering
    \captionsetup{font=footnotesize}
    \phantomcaption
    \label{fig:sweep-marginal-mse}
  \end{subfigure}

  \caption{\textbf{Reconstruction from layer subsets and individual layers.}
  Frozen 23-layer DINOv3-L comparing standard RAEv2 (gray) and \ours\ (orange).
  \textbf{(a)} Mean reconstruction PSNR (with 10th--90th percentile bands) versus the number of fused layers.
  \textbf{(b)} PSNR from each encoder layer alone on 10k validation images; dashed lines mark full-fusion reconstruction for this evaluation.
  \textbf{(c,d)} Step-by-step PSNR gain and MSE reduction when adding one more layer.
  \ours\ reconstructs accurately from individual layers and small subsets.}
  \label{fig:sweep-all}
\end{figure}

Building on the leave-one-out and Shapley probes in \Cref{fig:layer-usage-k23} (\S\ref{sec:decoder-analysis}), \Cref{fig:sweep-all} evaluates reconstruction from individual layers and random layer subsets of varying size. Since the encoder remains frozen, higher subset accuracy reflects improved decoding of information already present across depth. We sample random permutations of the 23 layers and
evaluate their prefixes, using the same permutations and
evaluation images for both decoders. Each prefix of length $k$
defines a subset containing $k$ layers.

\paragraph{Reconstruction quality from layer subsets.}
\Cref{fig:sweep-value} measures how reconstruction quality
changes with subset size. For each size, we aggregate per-image
PSNR across the sampled subsets; lines show means and bands
show the 10th--90th percentiles over image--subset pairs.
The curve reveals how quickly each decoder approaches its
full-fusion reconstruction quality as more layers become
available.

\paragraph{Single-layer reconstruction.}
\Cref{fig:single-layer-readout} feeds each encoder layer separately to the same
full-fusion decoder and measures PSNR on 10k ImageNet-256 validation images.
\ours\ reconstructs more accurately from every individual layer; dashed lines
show each decoder's full-fusion reconstruction.

The Shapley estimates in \Cref{fig:layer-usage-k23}b use the sampled permutations
described above: each layer's contribution is the average PSNR change when it
is added. The empty-set value is reconstruction from a zero latent input.
This measures contribution across subset contexts, complementing the
leave-one-out probe's evaluation in the presence of all other layers.

\paragraph{Gains from increasing subset size.}
\Cref{fig:sweep-marginal,fig:sweep-marginal-mse} measure the
marginal benefit of increasing subset size from $k{-}1$
to $k$. Let $\bar v(k)$ and $\bar e(k)$ denote mean PSNR
and mean per-image MSE, respectively, aggregated over images
and sampled subsets of size $k$. The two panels report
$\bar v(k)-\bar v(k{-}1)$ and
$\bar e(k{-}1)-\bar e(k)$.
Here, $k$ denotes subset size rather than an encoder layer
index. Computing MSE reductions directly from reconstruction
errors also checks whether diminishing gains persist beyond
the logarithmic PSNR scale.

\ours\ achieves higher reconstruction quality from small
subsets and approaches its full-fusion performance with fewer
layers. Its smaller marginal gains at larger subset sizes
therefore reflect earlier saturation after a stronger initial
readout. The corresponding decline in MSE reductions shows
that this pattern also holds on the underlying error scale.
Meanwhile, the broadly preserved Shapley ordering in \Cref{fig:layer-usage-k23}b indicates
that improved subset robustness coexists with unequal
contributions across layers. Together, these diagnostics support the interpretation that
\ours\ improves access to reconstruction information across
depths. They characterize the nonlinear decoder empirically
and complement the probability-weighted coverage analysis
in \Cref{cor:coverage}.

\paragraph{Energy-matched controls.}
\label{sec:exp-ablation}
\begin{table}[H]
  \centering
  \caption{\textbf{Energy-matched latent perturbation controls.}
  Decoders train on frozen DINOv3-L ($K{=}23$) for 5 epochs (batch size 256, MSE + LPIPS, GAN from epoch 3) and are evaluated on 10k validation images without stochastic augmentation. Isotropic Gaussian noise ($\sigma{=}0.667$) and inverted token/channel dropout are calibrated on $4{,}096$ training images to match the expected subset-perturbation energy $\mathbb{E}\lVert \delta_S \rVert_2^2$ ($\delta_S = z_S - \bar z$) of \ours\ at $p_{\mathrm{dec}}{=}0.95$.}
  \label{tab:perturbation-ablation}
  \small
  \setlength{\tabcolsep}{4.5pt}
  \settowidth{\abllab}{\small Deterministic multi-fusion}
  \begin{tabular}{@{}p{\abllab}*{9}{Q{\dimexpr(\textwidth-\abllab-18\tabcolsep)/9\relax}}@{}}
    \toprule
    & \multicolumn{3}{c}{Full fusion ($K{=}23$)} & \multicolumn{3}{c}{7-layer subset ($k{=}7$)} & \multicolumn{3}{c}{Single layer ($\ell_{11}$)} \\
    \cmidrule(lr){2-4}\cmidrule(lr){5-7}\cmidrule(lr){8-10}
    Decoder & PSNR$\uparrow$ & SSIM$\uparrow$ & rFID$\downarrow$ & PSNR$\uparrow$ & SSIM$\uparrow$ & rFID$\downarrow$ & PSNR$\uparrow$ & SSIM$\uparrow$ & rFID$\downarrow$ \\
    \midrule
    Gaussian noise ($\sigma{=}0.667$) & 24.12 & 0.688 & 3.47 & 13.18 & 0.397 & 15.17 & 18.24 & 0.534 & 7.47 \\
    Token dropout & 16.79 & 0.603 & 8.73 & 13.02 & 0.363 & 38.33 & 15.54 & 0.464 & 12.92 \\
    Channel dropout & 14.02 & 0.499 & 9.97 & 13.78 & 0.402 & 14.75 & 14.15 & 0.433 & 8.07 \\
    \rowcolor{fuseregcolor}
    \ours\ ($p_{\mathrm{dec}}{=}0.95$) & \textbf{26.12} & \textbf{0.776} & \textbf{2.45} & \textbf{23.14} & \textbf{0.624} & \textbf{3.68} & \textbf{24.13} & \textbf{0.672} & \textbf{2.98} \\
    \bottomrule
  \end{tabular}
\end{table}

We isolate the structure of the perturbation using 5-epoch decoder ablations
evaluated on 10k images (\Cref{tab:perturbation-ablation}). Gaussian noise and
inverted token/channel dropout are calibrated on 4,096 training images to
match the expected perturbation energy of \ours\ at $p_{\mathrm{dec}}{=}0.95$.
\ours\ achieves the strongest PSNR, SSIM, and rFID across all three tested
fusions. On single-layer inputs, it reaches $24.13$\,dB PSNR, compared with
$18.24$\,dB for energy-matched Gaussian noise. Matching the expected energy
$\mathbb E\|z_S-\bar z\|_2^2$ therefore does not reproduce the cross-fusion
tolerance of \ours: useful variation follows the directions in which encoder
layers disagree. Training all subset views toward the same image encourages
consistent decoding along these directions.
Stage-dependent generation behavior is discussed in \S\ref{sec:discussion}.
\FloatBarrier

\section{Supplementary Generation Results}
\label{app:generator-results}

All generation results in this section use a frozen, pretrained DINOv3-L encoder.

\subsection{Decoder replacement with fixed generated latents}
\label{app:decoder-swap}

\begin{figure}[H]
  \centering
  \includegraphics[width=\textwidth]{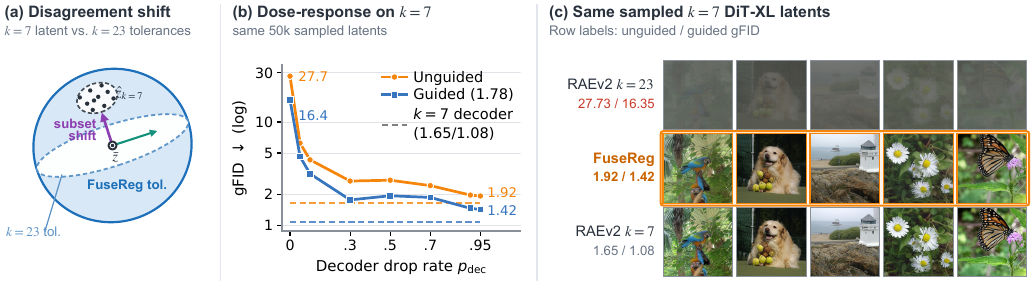}
  \caption{\textbf{Decoding 7-layer DiT samples with a 23-layer decoder.}
  \textbf{(a)} Seven-layer DiT samples center on the 7-layer mean, shifted from the 23-layer mean along layer-disagreement directions: they fall outside the narrow tolerance of a standard 23-layer RAEv2 decoder yet inside \ours.
  \textbf{(b)} Rendering the exact same 50k 7-layer DiT-XL latents with \ours\ decoders trained at larger subset-drop rates $p_{\mathrm{dec}}$ lowers gFID nearly monotonically toward a dedicated 7-layer decoder (\Cref{tab:dropdecode}).
  \textbf{(c)} Identical 7-layer DiT-XL latents rendered by three frozen decoders.}
  \label{fig:k7-stress}
\end{figure}

\Cref{tab:dropdecode,fig:k7-stress} give the full decoder-swap results, including the
$K{=}23$ settings in \Cref{fig:decoder-swap} and off-fusion $k{=}7$ transfer.
Each column pair keeps the DiT, fusion space, guidance
setting, and sampled latents fixed. This isolates whether a \ours\ decoder improves the final images
produced from an unchanged generator.

\begin{table}[H]
  \centering
  \caption{\textbf{\ours\ improves generation with a fixed generator.}
  Within each fusion and guidance setting, every decoder renders the same 50k
  latents from a fixed RAEv2 DiT ($p_{\mathrm{dit}}{=}0$). The gray row uses the
  reproduced fixed-$k{=}23$ decoder; other rows vary $p_{\mathrm{dec}}$ over the
  same $K{=}23$ layer pool. Internal guidance uses scale
  $1.78$~\citep{singh2026raev2}. \textbf{Bold}: best in each column.}
  \label{tab:dropdecode}
  \small
  \setlength{\tabcolsep}{4pt}
  \begin{tabular}{l cccc cccc}
    \toprule
    & \multicolumn{4}{c}{No guidance}
    & \multicolumn{4}{c}{Internal guidance (scale $1.78$)} \\
    \cmidrule(lr){2-5}\cmidrule(lr){6-9}
    & \multicolumn{2}{c}{DiT $k{=}23$} & \multicolumn{2}{c}{DiT $k{=}7$}
    & \multicolumn{2}{c}{DiT $k{=}23$} & \multicolumn{2}{c}{DiT $k{=}7$} \\
    \cmidrule(lr){2-3}\cmidrule(lr){4-5}\cmidrule(lr){6-7}\cmidrule(lr){8-9}
    Decoder $p_{\mathrm{dec}}$
      & gFID$\downarrow$ & IS$\uparrow$ & gFID$\downarrow$ & IS$\uparrow$
      & gFID$\downarrow$ & IS$\uparrow$ & gFID$\downarrow$ & IS$\uparrow$ \\
    \midrule
    \rowcolor{gray!12}
    $0.0$ (RAEv2) & 3.01 & 206.8 & 27.73 & 97.7 & \textbf{1.25} & \textbf{263.8} & 16.35 & 148.7 \\
    $0.05$ & 3.15 & 209.5 & 6.24 & 185.2 & 1.48 & 255.7 & 4.62 & 213.5 \\
    $0.1$  & 3.16 & 209.4 & 4.30 & 196.2 & 1.49 & 256.7 & 3.15 & 226.1 \\
    $0.3$  & 2.92 & 211.6 & 2.68 & 209.4 & 1.41 & 257.4 & 1.76 & 242.3 \\
    $0.5$  & 2.82 & 210.4 & 2.73 & 207.7 & 1.44 & 253.3 & 1.93 & 236.1 \\
    $0.7$  & 2.68 & 211.4 & 2.42 & 213.1 & 1.43 & 253.6 & 1.86 & 238.7 \\
    $0.9$  & 2.35 & 216.9 & 1.96 & 220.8 & 1.27 & 259.4 & 1.46 & 248.7 \\
    $0.95$ & \textbf{2.21} & \textbf{218.5} & \textbf{1.92} & \textbf{221.8} & \textbf{1.25} & 260.7 & \textbf{1.42} & \textbf{250.5} \\
    \bottomrule
  \end{tabular}
\end{table}

On the native $k{=}23$ fusion, increasing $p_{\mathrm{dec}}$ to $0.95$ lowers
unguided gFID from $3.01$ to $2.21$ and preserves guided gFID at $1.25$.
Intermediate rates have different effects
under guidance, motivating the joint-rate study below.

\paragraph{Extreme disagreement shifts.}
\label{app:extreme-shifts}
A seven-layer mean shifts from $\bar z$ purely along layer-disagreement
directions. The $k{=}7$ columns of \Cref{tab:dropdecode} and
\Cref{fig:k7-stress} use 50k latents from a fixed seven-layer DiT-XL to probe
decoder tolerance to this shift. Replacing the standard $k{=}23$ decoder with
\ours\ lowers gFID from $27.73$ to $1.92$ without guidance and from $16.35$ to
$1.42$ with guidance. Generation improves nearly monotonically as
$p_{\mathrm{dec}}$ increases, reaching within $0.27/0.34$ gFID
(unguided/guided) of a dedicated seven-layer decoder. Because the seven-layer
fusion lies inside \ours's training support, this test measures tolerance width;
it does not characterize the error distribution of a native $k{=}23$ generator.
\FloatBarrier

\subsection{Joint Regularization Across Model Scales}
\label{app:rate-intuition}

\Cref{tab:scale-b-grid} reports the full $7{\times}7$ unguided gFID and IS grids for DINOv3-L DiT-Base. \Cref{tab:scale} reports both metrics with and without internal guidance for DINOv3-L DiT-XL. All runs use the same frozen encoder and 40-epoch generator budget as the summary in \Cref{tab:fusereg-summary}; other encoder families are evaluated in \S\ref{app:siglip}.

% Paired DiT-Base metric grids on DINOv3-L.
\begin{table}[H]
\centering
\setlength{\fboxsep}{1.2pt}
\renewcommand{\arraystretch}{1.12}
\setlength{\tabcolsep}{2.5pt}
\caption{\textbf{DiT-Base generation metrics across regularization rates.}
Unguided class-conditional ImageNet-256 generation with DINOv3-L ($K{=}23$),
40 training epochs, and 50k samples. Rows vary the generator rate
$p_{\mathrm{dit}}$; columns vary the decoder rate $p_{\mathrm{dec}}$.
Every decoder in a row renders the same generated latents.
The \fcolorbox{green!60!black}{white}{green box} marks the reproduced RAEv2
fixed-fusion baseline $(0,0)$. Shading is relative to that baseline within
each metric (\textcolor{figblue}{Blue} $=$ better,
\textcolor{figorange}{Orange} $=$ worse). \textbf{Bold}: best per panel.}
\label{tab:scale-b-grid}
% Panel fragment for the combined DiT-Base metric table in dit_base_grid.tex.
\begin{subtable}[t]{0.485\textwidth}
\centering
\caption{Unguided gFID $\downarrow$.}
\label{tab:scale-b}
\footnotesize
\settowidth{\heatlab}{\footnotesize $p_{\text{dit}}\backslash p_{\text{dec}}$}
\begin{tabular}{@{}p{\heatlab}*{7}{Q{\dimexpr(\linewidth-\heatlab-14\tabcolsep)/7\relax}}@{}}
\toprule
$p_{\text{dit}}\backslash p_{\text{dec}}$ & 0 & 0.05 & 0.1 & 0.3 & 0.5 & 0.7 & 0.9 \\
\midrule
0    & \cellcolor{white}\smash{\fcolorbox{green!60!black}{white}{13.96}} & \cellcolor{heatorange!49}17.28 & \cellcolor{heatorange!49}17.26 & \cellcolor{heatorange!31}16.05 & \cellcolor{heatorange!20}15.28 & \cellcolor{heatorange!5}14.32 & \cellcolor{heatblue!15}12.96 \\
0.05 & \cellcolor{heatblue!7}13.52 & \cellcolor{heatorange!31}16.03 & \cellcolor{heatorange!32}16.08 & \cellcolor{heatorange!15}15.00 & \cellcolor{heatorange!5}14.28 & \cellcolor{heatblue!7}13.48 & \cellcolor{heatblue!26}12.21 \\
0.1  & \cellcolor{heatblue!24}12.37 & \cellcolor{heatorange!13}14.85 & \cellcolor{heatorange!14}14.93 & \cellcolor{white}13.97 & \cellcolor{heatblue!11}13.25 & \cellcolor{heatblue!21}12.52 & \cellcolor{heatblue!39}11.37 \\
0.3  & \cellcolor{heatblue!22}12.45 & \cellcolor{heatorange!2}14.12 & \cellcolor{heatorange!4}14.20 & \cellcolor{heatblue!9}13.33 & \cellcolor{heatblue!18}12.72 & \cellcolor{heatblue!28}12.09 & \cellcolor{heatblue!44}10.99 \\
0.5  & \cellcolor{heatblue!25}12.27 & \cellcolor{heatblue!7}13.47 & \cellcolor{heatblue!6}13.57 & \cellcolor{heatblue!18}12.72 & \cellcolor{heatblue!26}12.18 & \cellcolor{heatblue!36}11.52 & \cellcolor{heatblue!52}10.50 \\
0.7  & \cellcolor{heatblue!28}12.09 & \cellcolor{heatblue!19}12.71 & \cellcolor{heatblue!17}12.83 & \cellcolor{heatblue!29}12.04 & \cellcolor{heatblue!36}11.53 & \cellcolor{heatblue!45}10.96 & \cellcolor{heatblue!60}\textbf{9.93} \\
0.9  & \cellcolor{heatblue!2}13.84 & \cellcolor{heatblue!5}13.63 & \cellcolor{heatblue!4}13.71 & \cellcolor{heatblue!16}12.90 & \cellcolor{heatblue!23}12.44 & \cellcolor{heatblue!32}11.80 & \cellcolor{heatblue!49}10.68 \\
\bottomrule
\end{tabular}
\end{subtable}

\hfill
% Panel fragment for the combined DiT-Base metric table in dit_base_grid.tex.
\begin{subtable}[t]{0.485\textwidth}
\centering
\caption{Unguided IS $\uparrow$.}
\label{tab:scale-b-is}
\footnotesize
\settowidth{\heatlab}{\footnotesize $p_{\text{dit}}\backslash p_{\text{dec}}$}
\begin{tabular}{@{}p{\heatlab}*{7}{Q{\dimexpr(\linewidth-\heatlab-14\tabcolsep)/7\relax}}@{}}
\toprule
$p_{\text{dit}}\backslash p_{\text{dec}}$ & 0 & 0.05 & 0.1 & 0.3 & 0.5 & 0.7 & 0.9 \\
\midrule
0    & \cellcolor{white}\smash{\fcolorbox{green!60!black}{white}{107.5}} & \cellcolor{heatorange!24}100.3 & \cellcolor{heatorange!24}100.3 & \cellcolor{heatorange!15}103.0 & \cellcolor{heatorange!5}105.9 & \cellcolor{heatblue!1}107.7 & \cellcolor{heatblue!13}111.5 \\
0.05 & \cellcolor{heatblue!3}108.5 & \cellcolor{heatorange!15}102.8 & \cellcolor{heatorange!16}102.5 & \cellcolor{heatorange!7}105.4 & \cellcolor{heatblue!2}108.0 & \cellcolor{heatblue!6}109.4 & \cellcolor{heatblue!20}113.6 \\
0.1  & \cellcolor{heatblue!22}114.1 & \cellcolor{heatblue!2}108.0 & \cellcolor{heatblue!2}108.0 & \cellcolor{heatblue!9}110.2 & \cellcolor{heatblue!18}113.0 & \cellcolor{heatblue!24}114.7 & \cellcolor{heatblue!36}118.5 \\
0.3  & \cellcolor{heatblue!19}113.4 & \cellcolor{heatblue!7}109.5 & \cellcolor{heatblue!7}109.5 & \cellcolor{heatblue!14}111.7 & \cellcolor{heatblue!22}114.3 & \cellcolor{heatblue!27}115.7 & \cellcolor{heatblue!38}119.1 \\
0.5  & \cellcolor{heatblue!22}114.1 & \cellcolor{heatblue!15}112.1 & \cellcolor{heatblue!15}112.2 & \cellcolor{heatblue!23}114.4 & \cellcolor{heatblue!31}116.9 & \cellcolor{heatblue!35}118.2 & \cellcolor{heatblue!47}121.7 \\
0.7  & \cellcolor{heatblue!31}116.8 & \cellcolor{heatblue!30}116.6 & \cellcolor{heatblue!30}116.7 & \cellcolor{heatblue!37}118.8 & \cellcolor{heatblue!45}121.2 & \cellcolor{heatblue!49}122.3 & \cellcolor{heatblue!60}\textbf{125.7} \\
0.9  & \cellcolor{heatblue!13}111.4 & \cellcolor{heatblue!21}113.8 & \cellcolor{heatblue!20}113.7 & \cellcolor{heatblue!28}115.9 & \cellcolor{heatblue!34}117.8 & \cellcolor{heatblue!42}120.1 & \cellcolor{heatblue!53}123.5 \\
\bottomrule
\end{tabular}
\end{subtable}

\end{table}

% Paired gFID and IS rate grids for DiT-XL on DINOv3-L.
\begin{table}[H]
\centering
\setlength{\fboxsep}{1.2pt}
\caption{\textbf{DiT-XL generation metrics across regularization rates and guidance settings.}
Class-conditional ImageNet-256, DINOv3-L ($K{=}23$), 16-epoch decoders,
40-epoch DiT-XL, and 50k samples. Columns of panels compare unguided generation
with internal guidance at scale $1.78$; the top and bottom rows report gFID
and IS, respectively. Within each panel, rows vary the generator rate
$p_{\mathrm{dit}}$ and columns vary the decoder rate $p_{\mathrm{dec}}$,
with every decoder in a row rendering the same generated latents.
The \fcolorbox{green!60!black}{white}{green box} marks the fixed-fusion baseline
$(0,0)$. Shading is relative to that baseline within each panel
(\textcolor{figblue}{Blue} $=$ better, \textcolor{figorange}{Orange} $=$ worse).
\textbf{Bold}: best per panel.}
\label{tab:scale}
\label{tab:scale-xl-main}
\label{tab:ditxl-is}
\label{tab:k23_dit_dec_heatmap}
% gFID panel fragments: first row of the combined DiT-XL table in dit_scale.tex.
\renewcommand{\arraystretch}{1.12}
\setlength{\tabcolsep}{1.8pt}

\begin{subtable}[t]{0.485\textwidth}
\centering
\caption{Unguided gFID $\downarrow$.}
\label{tab:scale-xl}
\footnotesize
\settowidth{\heatlab}{\footnotesize $p_{\text{dit}}\backslash p_{\text{dec}}$}
\begin{tabular}{@{}p{\heatlab}*{8}{Q{\dimexpr(\linewidth-\heatlab-16\tabcolsep)/8\relax}}@{}}
\toprule
$p_{\text{dit}}\backslash p_{\text{dec}}$ & 0 & 0.05 & 0.1 & 0.3 & 0.5 & 0.7 & 0.9 & 0.95 \\
\midrule
0   & \cellcolor{white}\smash{\fcolorbox{green!60!black}{white}{2.91}} & \cellcolor{heatorange!34}3.21 & \cellcolor{heatorange!35}3.22 & \cellcolor{heatorange!10}3.00 & \cellcolor{white}2.91 & \cellcolor{heatblue!17}2.76 & \cellcolor{heatblue!46}2.50 & \cellcolor{heatblue!60}\textbf{2.38} \\
0.5 & \cellcolor{white}2.91 & \cellcolor{heatorange!32}3.19 & \cellcolor{heatorange!34}3.21 & \cellcolor{heatorange!12}3.02 & \cellcolor{heatorange!2}2.93 & \cellcolor{heatblue!12}2.80 & \cellcolor{heatblue!40}2.56 & \cellcolor{heatblue!52}2.45 \\
0.7 & \cellcolor{heatorange!6}2.96 & \cellcolor{heatorange!25}3.13 & \cellcolor{heatorange!29}3.17 & \cellcolor{heatorange!7}2.97 & \cellcolor{heatblue!1}2.90 & \cellcolor{heatblue!16}2.77 & \cellcolor{heatblue!43}2.53 & \cellcolor{heatblue!55}2.42 \\
0.9 & \cellcolor{heatorange!20}3.09 & \cellcolor{heatorange!32}3.19 & \cellcolor{heatorange!37}3.24 & \cellcolor{heatorange!11}3.01 & \cellcolor{heatorange!3}2.94 & \cellcolor{heatblue!17}2.76 & \cellcolor{heatblue!46}2.50 & \cellcolor{heatblue!59}2.39 \\
\bottomrule
\end{tabular}
\end{subtable}
\hfill
\begin{subtable}[t]{0.485\textwidth}
\centering
\caption{Guided gFID $\downarrow$.}
\label{tab:scale-xl-guided}
\label{tab:k23_gfid_ig178}
\footnotesize
\settowidth{\heatlab}{\footnotesize $p_{\text{dit}}\backslash p_{\text{dec}}$}
\begin{tabular}{@{}p{\heatlab}*{8}{Q{\dimexpr(\linewidth-\heatlab-16\tabcolsep)/8\relax}}@{}}
\toprule
$p_{\text{dit}}\backslash p_{\text{dec}}$ & 0 & 0.05 & 0.1 & 0.3 & 0.5 & 0.7 & 0.9 & 0.95 \\
\midrule
0 & \cellcolor{white}\smash{\fcolorbox{green!60!black}{white}{1.57}} & \cellcolor{heatorange!15}1.66 & \cellcolor{heatorange!15}1.66 & \cellcolor{heatorange!3}1.59 & \cellcolor{white}1.57 & \cellcolor{heatblue!26}1.54 & \cellcolor{heatblue!60}\textbf{1.50} & \cellcolor{heatblue!51}1.51 \\
0.5 & \cellcolor{heatorange!16}1.67 & \cellcolor{heatorange!50}1.88 & \cellcolor{heatorange!50}1.88 & \cellcolor{heatorange!32}1.77 & \cellcolor{heatorange!23}1.71 & \cellcolor{heatorange!15}1.66 & \cellcolor{heatorange!5}1.60 & \cellcolor{heatorange!3}1.59 \\
0.7 & \cellcolor{heatorange!16}1.67 & \cellcolor{heatorange!60}1.94 & \cellcolor{heatorange!60}1.94 & \cellcolor{heatorange!39}1.81 & \cellcolor{heatorange!28}1.74 & \cellcolor{heatorange!18}1.68 & \cellcolor{heatorange!5}1.60 & \cellcolor{heatorange!2}1.58 \\
0.9 & \cellcolor{heatorange!15}1.66 & \cellcolor{heatorange!58}1.93 & \cellcolor{heatorange!58}1.93 & \cellcolor{heatorange!34}1.78 & \cellcolor{heatorange!24}1.72 & \cellcolor{heatorange!13}1.65 & \cellcolor{heatblue!17}1.55 & \cellcolor{heatblue!34}1.53 \\
\bottomrule
\end{tabular}
\end{subtable}

\par\medskip
% IS panel fragments: second row of the combined DiT-XL table in dit_scale.tex.
\renewcommand{\arraystretch}{1.12}
\setlength{\tabcolsep}{1.8pt}

\begin{subtable}[t]{0.485\textwidth}
\centering
\caption{Unguided IS $\uparrow$.}
\label{tab:scale-xl-is}
\footnotesize
\settowidth{\heatlab}{\footnotesize $p_{\text{dit}}\backslash p_{\text{dec}}$}
\begin{tabular}{@{}p{\heatlab}*{8}{Q{\dimexpr(\linewidth-\heatlab-16\tabcolsep)/8\relax}}@{}}
\toprule
$p_{\text{dit}}\backslash p_{\text{dec}}$ & 0 & 0.05 & 0.1 & 0.3 & 0.5 & 0.7 & 0.9 & 0.95 \\
\midrule
0   & \cellcolor{white}\smash{\fcolorbox{green!60!black}{white}{207.4}} & \cellcolor{heatorange!13}204.4 & \cellcolor{heatorange!13}204.5 & \cellcolor{heatorange!3}206.8 & \cellcolor{heatblue!1}207.7 & \cellcolor{heatblue!7}208.9 & \cellcolor{heatblue!16}211.1 & \cellcolor{heatblue!24}212.8 \\
0.5 & \cellcolor{heatblue!11}209.8 & \cellcolor{heatorange!5}206.2 & \cellcolor{heatorange!6}206.1 & \cellcolor{heatblue!2}207.8 & \cellcolor{heatblue!8}209.3 & \cellcolor{heatblue!11}210.0 & \cellcolor{heatblue!21}212.1 & \cellcolor{heatblue!28}213.8 \\
0.7 & \cellcolor{heatblue!14}210.7 & \cellcolor{white}207.5 & \cellcolor{heatblue!1}207.7 & \cellcolor{heatblue!5}208.5 & \cellcolor{heatblue!12}210.1 & \cellcolor{heatblue!18}211.4 & \cellcolor{heatblue!28}213.8 & \cellcolor{heatblue!36}215.6 \\
0.9 & \cellcolor{heatblue!27}213.5 & \cellcolor{heatblue!21}212.1 & \cellcolor{heatblue!21}212.1 & \cellcolor{heatblue!29}214.0 & \cellcolor{heatblue!35}215.3 & \cellcolor{heatblue!40}216.5 & \cellcolor{heatblue!52}219.2 & \cellcolor{heatblue!60}\textbf{221.1} \\
\bottomrule
\end{tabular}
\end{subtable}
\hfill
\begin{subtable}[t]{0.485\textwidth}
\centering
\caption{Guided IS $\uparrow$.}
\label{tab:k23_is_ig178}
\footnotesize
\settowidth{\heatlab}{\footnotesize $p_{\text{dit}}\backslash p_{\text{dec}}$}
\begin{tabular}{@{}p{\heatlab}*{8}{Q{\dimexpr(\linewidth-\heatlab-16\tabcolsep)/8\relax}}@{}}
\toprule
$p_{\text{dit}}\backslash p_{\text{dec}}$ & 0 & 0.05 & 0.1 & 0.3 & 0.5 & 0.7 & 0.9 & 0.95 \\
\midrule
0 & \cellcolor{white}\smash{\fcolorbox{green!60!black}{white}{\textbf{262.6}}} & \cellcolor{heatorange!27}255.8 & \cellcolor{heatorange!24}256.7 & \cellcolor{heatorange!21}257.3 & \cellcolor{heatorange!18}258.2 & \cellcolor{heatorange!18}258.2 & \cellcolor{heatorange!17}258.5 & \cellcolor{heatorange!13}259.4 \\
0.5 & \cellcolor{heatorange!15}258.8 & \cellcolor{heatorange!52}249.8 & \cellcolor{heatorange!48}250.7 & \cellcolor{heatorange!41}252.3 & \cellcolor{heatorange!39}253.0 & \cellcolor{heatorange!38}253.2 & \cellcolor{heatorange!33}254.5 & \cellcolor{heatorange!27}255.8 \\
0.7 & \cellcolor{heatorange!22}257.1 & \cellcolor{heatorange!60}247.7 & \cellcolor{heatorange!58}248.2 & \cellcolor{heatorange!53}249.4 & \cellcolor{heatorange!46}251.1 & \cellcolor{heatorange!46}251.2 & \cellcolor{heatorange!43}251.8 & \cellcolor{heatorange!37}253.5 \\
0.9 & \cellcolor{heatorange!3}261.9 & \cellcolor{heatorange!40}252.6 & \cellcolor{heatorange!37}253.3 & \cellcolor{heatorange!30}255.1 & \cellcolor{heatorange!24}256.6 & \cellcolor{heatorange!23}257.0 & \cellcolor{heatorange!18}258.1 & \cellcolor{heatorange!10}260.1 \\
\bottomrule
\end{tabular}
\end{subtable}

\end{table}

On DiT-Base, the $(0,0)$ baseline gFID is $13.96$. Decoder-only regularization at
$p_{\mathrm{dec}}{=}0.9$ improves it by $1.00$, and generator-only regularization
at $p_{\mathrm{dit}}{=}0.7$ improves it by $1.87$. Their combination reaches
$9.93$, an improvement of $4.03$, which is $1.16$ beyond the additive extrapolation
of the two individual gains, and all six nonzero generator rates at $p_{\mathrm{dec}}{=}0.9$ improve over $(0,0.9)$. The analogous IS contrast at $(0.7,0.9)$ is $4.9$. This four-setting
comparison describes complementarity at the selected configuration rather than a
statistical interaction test.

The DiT-XL grids separate this effect from a universal rate prescription. Their lowest
unguided and guided gFID occurs with decoder regularization and
$p_{\mathrm{dit}}{=}0$ (\Cref{tab:scale-xl,tab:scale-xl-guided}).
The generator-side gFID benefit established on DiT-Base is therefore specific to
the tested model scale and training budget, while decoder robustness remains
useful across the two scales.

\paragraph{Guidance and metric preferences.}
\label{app:generation}
The gFID and IS panels of \Cref{tab:scale} evaluate the same DiT-XL settings;
internal guidance uses scale $1.78$. Without guidance, IS increases from $207.4$ at $(0,0)$ to $221.1$ at
$(p_{\mathrm{dit}},p_{\mathrm{dec}})=(0.9,0.95)$.
With internal guidance, the highest IS is $262.6$ at $(0,0)$, whereas
the lowest guided gFID is $1.50$ at $(0,0.9)$.
Thus the metric and guidance setting select different rate pairs. The distinction
motivates calibrating guidance and \ours\ together, as discussed in
\S\ref{sec:discussion}.
\FloatBarrier

\section{Generation Grids on Other Encoders}
\label{app:siglip}
\label{app:other-encoders-generation}

\Cref{tab:scale-other-encoders} evaluates joint regularization on frozen SigLIP2-L~\citep{tschannen2025siglip} ($K{=}23$) and EUPE-B~\citep{zhu2026efficientuniversalperceptionencoder} ($K{=}11$, encoder blocks $1$--$11$). We train decoders for 16 epochs and DiT-Base generators for 40 epochs on ImageNet-256, using the DINOv3-L recipes and rates $p_{\mathrm{dec}},p_{\mathrm{dit}}\in\{0,0.3,0.6,0.9\}$. Each encoder's paired panels report unguided gFID and IS. For each generator rate, all decoders render the same 50k latents sampled with 50 Euler steps; gFID uses the 50k ImageNet-256 validation images as reference.

\begin{table}[H]
\centering
\caption{\textbf{\ours\ at both stages across encoder families.}
Unguided generation on class-conditional ImageNet-256 with frozen SigLIP2-L
($K{=}23$, top) and EUPE-B ($K{=}11$, bottom) encoders, 16-epoch decoders,
and 40-epoch DiT-Base generators; 50k samples and 50 Euler steps.
Rows vary $p_{\mathrm{dit}}$; columns vary $p_{\mathrm{dec}}$, with sampled latents
fixed within each row. Each \fcolorbox{green!60!black}{white}{green box} marks
that encoder's fixed-fusion baseline. Shading is relative to this baseline
(\textcolor{figblue}{Blue} $=$ better, \textcolor{figorange}{Orange} $=$ worse),
with intensity scaled within each panel. \textbf{Bold}: best per panel.}
\label{tab:scale-other-encoders}
\setlength{\fboxsep}{1.2pt}
\setlength{\tabcolsep}{4pt}
\renewcommand{\arraystretch}{1.15}
% Encoder-specific panels; included by dit_scale_other_encoders.tex.
\begin{subtable}[t]{0.49\textwidth}
\centering
\caption{SigLIP2-L, gFID $\downarrow$.}
\label{tab:scale-siglip-b}
\small
\begin{tabular}{lcccc}
\toprule
$p_{\text{dit}}\backslash p_{\text{dec}}$ & 0 & 0.3 & 0.6 & 0.9 \\
\midrule
0    & \cellcolor{white}\smash{\fcolorbox{green!60!black}{white}{13.77}} & \cellcolor{heatblue!21}11.80 & \cellcolor{heatblue!31}10.91 & \cellcolor{heatblue!44}9.73 \\
0.3  & \cellcolor{heatblue!20}11.93 & \cellcolor{heatblue!38}10.33 & \cellcolor{heatblue!45}9.61 & \cellcolor{heatblue!55}8.69 \\
0.6  & \cellcolor{heatblue!27}11.26 & \cellcolor{heatblue!44}9.77 & \cellcolor{heatblue!51}9.08 & \cellcolor{heatblue!60}\textbf{8.26} \\
0.9  & \cellcolor{heatblue!20}11.97 & \cellcolor{heatblue!41}10.00 & \cellcolor{heatblue!49}9.30 & \cellcolor{heatblue!58}8.49 \\
\bottomrule
\end{tabular}
\end{subtable}
\hfill
\begin{subtable}[t]{0.49\textwidth}
\centering
\caption{SigLIP2-L, IS $\uparrow$.}
\label{tab:scale-siglip-b-is}
\small
\begin{tabular}{lcccc}
\toprule
$p_{\text{dit}}\backslash p_{\text{dec}}$ & 0 & 0.3 & 0.6 & 0.9 \\
\midrule
0    & \cellcolor{white}\smash{\fcolorbox{green!60!black}{white}{100.8}} & \cellcolor{heatblue!16}107.9 & \cellcolor{heatblue!24}111.8 & \cellcolor{heatblue!35}116.7 \\
0.3  & \cellcolor{heatblue!16}108.1 & \cellcolor{heatblue!32}115.3 & \cellcolor{heatblue!38}117.9 & \cellcolor{heatblue!49}122.9 \\
0.6  & \cellcolor{heatblue!26}112.7 & \cellcolor{heatblue!41}119.5 & \cellcolor{heatblue!49}123.0 & \cellcolor{heatblue!57}126.6 \\
0.9  & \cellcolor{heatblue!23}111.3 & \cellcolor{heatblue!43}120.4 & \cellcolor{heatblue!51}123.7 & \cellcolor{heatblue!60}\textbf{127.9} \\
\bottomrule
\end{tabular}
\end{subtable}

\par\vspace{0.6em}
% Encoder-specific panels; included by dit_scale_other_encoders.tex.
\begin{subtable}[t]{0.49\textwidth}
\centering
\caption{EUPE-B, gFID $\downarrow$.}
\label{tab:scale-eupe-b}
\small
\begin{tabular}{lcccc}
\toprule
$p_{\text{dit}}\backslash p_{\text{dec}}$ & 0 & 0.3 & 0.6 & 0.9 \\
\midrule
0    & \cellcolor{white}\smash{\fcolorbox{green!60!black}{white}{12.17}} & \cellcolor{heatblue!25}10.44 & \cellcolor{heatblue!36}9.68 & \cellcolor{heatblue!52}8.52 \\
0.3  & \cellcolor{heatblue!15}11.10 & \cellcolor{heatblue!38}9.56 & \cellcolor{heatblue!47}8.88 & \cellcolor{heatblue!60}\textbf{8.00} \\
0.6  & \cellcolor{heatblue!14}11.22 & \cellcolor{heatblue!37}9.62 & \cellcolor{heatblue!46}8.95 & \cellcolor{heatblue!58}8.13 \\
0.9  & \cellcolor{heatorange!26}13.97 & \cellcolor{heatblue!15}11.13 & \cellcolor{heatblue!27}10.31 & \cellcolor{heatblue!39}9.43 \\
\bottomrule
\end{tabular}
\end{subtable}
\hfill
\begin{subtable}[t]{0.49\textwidth}
\centering
\caption{EUPE-B, IS $\uparrow$.}
\label{tab:scale-eupe-b-is}
\small
\begin{tabular}{lcccc}
\toprule
$p_{\text{dit}}\backslash p_{\text{dec}}$ & 0 & 0.3 & 0.6 & 0.9 \\
\midrule
0    & \cellcolor{white}\smash{\fcolorbox{green!60!black}{white}{101.8}} & \cellcolor{heatblue!17}108.0 & \cellcolor{heatblue!25}111.0 & \cellcolor{heatblue!37}115.4 \\
0.3  & \cellcolor{heatblue!19}108.7 & \cellcolor{heatblue!36}114.9 & \cellcolor{heatblue!44}117.8 & \cellcolor{heatblue!56}122.3 \\
0.6  & \cellcolor{heatblue!22}109.8 & \cellcolor{heatblue!41}116.7 & \cellcolor{heatblue!50}120.2 & \cellcolor{heatblue!60}\textbf{123.7} \\
0.9  & \cellcolor{heatblue!3}102.9 & \cellcolor{heatblue!33}113.8 & \cellcolor{heatblue!42}117.0 & \cellcolor{heatblue!53}121.2 \\
\bottomrule
\end{tabular}
\end{subtable}

\end{table}

The complementary gains observed with DINOv3-L (\Cref{tab:scale-b-grid}) extend to both encoder families. Moving right along each row of \Cref{tab:scale-other-encoders}, stronger decoder regularization lowers gFID and raises IS. Relative to fixed fusion, applying \ours\ jointly reduces gFID from $13.77$ to $8.26$ for SigLIP2-L and from $12.17$ to $8.00$ for EUPE-B. In both grids, the best gFID and IS occur with a strongly regularized decoder, while the preferred generator rate depends on the encoder and metric.

\FloatBarrier

\section{FuseReg: Detailed Method}
\label{app:fusereg-method}

\ours\ targets dependence on a fixed fusion rather than searching for a different fixed
fusion. We first place existing fixed fusions in a common
formulation, then construct a mean-preserving distribution over layer subsets, and
finally apply it at the two prediction stages of an RAE.

\subsection{From a fixed fusion to a distribution over layer subsets}
\label{sec:selection-rules}

A fixed fusion exposes each downstream model to only one layer composition during
training. We express fixed subsets and learned deterministic weightings using common fusion
weights to compare how downstream models respond to changes in layer composition.

Given an image $x$ and a frozen encoder $E$, let
$f_i(x) = \operatorname{LN}(E_{\ell_i}(x)) \in \mathbb{R}^{N \times d}$
be the layer-normalized tokens for each of $K$ candidate layers ($\ell_i \in \{\ell_1, \dots, \ell_K\}$), with $N$ patch tokens and channel width $d$. A layer-fusion rule maps these features to a single
latent consumed by the pixel decoder or diffusion transformer:
\begin{equation}
  z_w(x)=\sum_{i=1}^K w_i f_i(x),
  \qquad \mathbf1^\top w=1,
  \label{eq:fusion-rule}
\end{equation}
where the weight vector $w\in\mathbb R^K$ is chosen independently of the input.
We use normalized weights throughout the analysis; empirical fixed-fusion baselines retain their stated scale convention.

Fixed single-layer and $k$-layer fusions use a constant weight vector $w$, whereas a learned
deterministic weighting optimizes $w$ while sharing it across inputs.
\ours\ instead samples random layer subsets during training, exposing downstream
consumers to diverse layer compositions.

\subsection{Training on normalized random-subset means}
\label{sec:regularizer}

\ours\ constructs a distribution over layer subsets that varies layer composition during training while preserving the deployment latent in expectation. For each training example and drop rate $p\in[0,1)$, draw independent layer-retention indicators $\widetilde s_i$ and condition on retaining a nonempty subset $S=\{i:\widetilde s_i=1\}$:
\begin{equation}
  \begin{gathered}
    \widetilde s_i \stackrel{\mathrm{iid}}{\sim}\mathrm{Bernoulli}(1-p),
    \qquad i=1,\dots,K,\\[-2pt]
    S\sim\pi_p:=
    \operatorname{Law}\!\left(
      \{i:\widetilde s_i=1\}\,\middle|\,\sum_{i=1}^K \widetilde s_i>0
    \right),\\[-1pt]
    z_S(x) = \frac{1}{|S|}\sum_{i\in S} f_i(x).
  \end{gathered}
  \label{eq:drop}
\end{equation}
The corresponding fusion weight vector is $w_S=\mathbf1_S/|S|$.
For $0<p<1$, $\pi_p$ has full support on all $2^K-1$ nonempty subsets. At inference, all $K$ layers are retained by
default so $z_S(x)$ becomes the full-fusion latent $\bar z(x)=K^{-1}\sum_{i=1}^K f_i(x)$.

Size normalization in \Cref{eq:drop} is essential: exchangeability and normalization give
$\mathbb{E}_S[z_S(x)\mid x]=\bar z(x)$, so varying $p$ controls cross-layer disagreement perturbations around $\bar z(x)$ without shifting the expected latent (\S\ref{app:theory}).

\subsection{Two prediction stages, two regularization rates}
\label{sec:two-stages}

Because pixel reconstruction and noisy latent denoising solve different prediction tasks, we assign independent drop rates $p_{\mathrm{dec}}$ and $p_{\mathrm{dit}}$ to Stage 1 and Stage 2.

\paragraph{Decoder: subset fusions to pixels ($p_{\mathrm{dec}}$).}
A ViT pixel decoder $D_\phi$ reconstructs $\hat{x}=D_\phi(z_S(x))$ from $S\sim\pi_{p_{\mathrm{dec}}}$ under the standard pixel, perceptual, and adversarial objective~\citep{zhang2018lpips,goodfellow2014gan}:
$\mathcal{L}_{\mathrm{dec}}=\lVert\hat{x}-x\rVert_2^2
+\lambda_{\mathrm{p}}\mathcal{L}_{\mathrm{LPIPS}}(\hat{x},x)
+\lambda_{\mathrm{g}}\mathcal{L}_{\mathrm{GAN}}(\hat{x})$.

\paragraph{DiT: noisy subset latents to the full-layer fusion ($p_{\mathrm{dit}}$).}
In Stage 2, the prediction target remains the deterministic full-fusion latent $\bar z(x)=K^{-1}\sum_{i=1}^K f_i(x)$. Forming the flow-matching interpolant~\citep{lipman2023flowmatching}
$z_{t,S}=t z_S(x)+(1-t)\epsilon$ from a sampled subset view ($S\sim\pi_{p_{\mathrm{dit}}}$, $t\sim\nu$, $\epsilon\sim\mathcal{N}(0,I)$, class label $c$), the DiT $g_\theta$ minimizes the time-weighted clean-target objective:
\begin{equation}
  \mathcal{L}_{\mathrm{DiT}}
  =\mathbb{E}_{(x,c),S,t,\epsilon}
  \Bigl[\rho(t)\bigl\lVert g_\theta(z_{t,S},t,c)-\bar z(x)\bigr\rVert_2^2\Bigr].
  \label{eq:dit-objective}
\end{equation}
Here $\rho(t)=\max(1-t,0.05)^{-2}$ and $\nu$ is the shifted logit-normal law in the noise-to-data time direction (\S\ref{app:linear-surrogates})~\citep{singh2026raev2}. Setting $p_{\mathrm{dit}}=0$ recovers the standard full-fusion objective, while $p_{\mathrm{dit}}>0$ trains the generator to recover the same full-layer target across varied subset inputs. The pair $(p_{\mathrm{dec}},p_{\mathrm{dit}})\in[0,1)^2$ forms the exchangeable two-dimensional slice of the per-layer rate relaxation in \S\ref{app:relaxation}.

\section{FuseLift: Detailed Method}
\label{app:fuselift-method}

FuseLift relocates the diffusion target inside a frozen decoder. This section
specifies target extraction, normalization, training, and evaluation protocols;
\S\ref{app:fuselift} gives the reconstruction identity and geometric analysis.

\subsection{Selecting a target inside a frozen decoder}
\label{app:lift-construction}

Follow \S\ref{sec:preliminaries} in writing $\bar z(x)$ for the full-fusion encoder representation, reserving $f_i$ for frozen encoder-layer features and $h_j$ for internal decoder activations. Split a trained pixel decoder $D_\phi$ at location $j\ge 0$:
\begin{equation}
  D_\phi=R_j\circ T_j,
  \qquad h_j(x)=T_j(\bar z(x)).
  \label{eq:lift-split}
\end{equation}
The prefix $T_j$ contains every operation needed to form the activation at that location, and $R_j$ contains all subsequent operations and the pixel output head. Shared preprocessing, fixed classifier-surrogate conditioning, positional terms, and any additional decoder state required by the suffix are included in this split and held identical for the same image. At $j=0$, the split is immediately after the input projection and before the first transformer block. At $j=1$, it is after the first transformer block.

\paragraph{Target normalization is part of the coordinate system.}
Let $\mathcal N_j(h)=S_j(h-\mu_j)$ be a fixed invertible affine target normalization, with nonzero channel scales. Define the actual diffusion target and its rendering map by
\begin{equation}
\begin{aligned}
  u_j(x)&=\mathcal N_j(T_j(\bar z(x))),\\
  \mathcal R_j&=R_j\circ\mathcal N_j^{-1}.
\end{aligned}
\label{eq:lift-normalized-target}
\end{equation}
Both $T_j$ and $R_j$, together with all normalization statistics, remain frozen during diffusion training. Normalization is inverted before a generated sample enters $R_j$. Identity normalization recovers $u_j=h_j$.

\subsection{Diffusion training and sampling}
\label{app:lift-training}
For $u_j=u_j(x)$, sample $t\sim\nu$, $\epsilon\sim\mathcal N(0,I)$, and class condition $c$, and train
\begin{equation}
\begin{aligned}
  u_{j,t}&=t u_j+(1-t)\epsilon,\\
  \mathcal L_{\mathrm{Lift}}^{(j)}(\theta)
  &=\mathbb E_{(x,c),t,\epsilon}\rho(t)\\
  &\quad\cdot\left\|g_\theta^{(j)}(u_{j,t},t,c)-u_j\right\|_2^2.
\end{aligned}
\label{eq:lift-objective}
\end{equation}
This uses the time law and weighting of \Cref{eq:dit-objective}, with input and output projections sized for the selected target. In the target-relocation runs, the encoder and decoder are frozen, only the DiT is optimized, and the REPA alignment target and internal-guidance protocol follow the encoder-space RAEv2 recipe~\citep{singh2026raev2}. Sampling produces $\hat u_j$, followed by deterministic rendering $\hat x=\mathcal R_j(\hat u_j)$.

\subsection{Depth-sweep protocols and extended generation metrics}
\label{app:lift-depth-protocol}

\paragraph{Linear-accessibility diagnostic.}
\Cref{tab:decoder_layer_sweep} uses 2048 ImageNet images, with 1536 images for
ridge fitting and 512 held out for evaluation. The encoder input is the same
DINOv3-L full-fusion latent at every depth. Hidden states are measured after
$j\in\{0,1,2,4,8,14,20,27\}$ of the 28 decoder blocks; $h_0$ includes the
decoder and positional embeddings. Features undergo per-position mean removal
and per-channel standardization. Ridge regularization is
$\lambda=10^{-2}$ times the mean diagonal of $X^\top X$.
The probes predict the encoder target and DINOv3-L layers $\ell_2, \ell_8, \ell_{14}, \ell_{17}$,
and $\ell_{23}$; participation ratio is computed from the token covariance.
These are diagnostics of linear accessibility and covariance geometry.

\paragraph{Suffix-decoding check.}
A separate 128-image diagnostic substitutes a dataset-mean CLS token and renders
each hidden-state target through the matching suffix. RAEv2 PSNR is
$28.54$--$28.55$\,dB across depths ($28.55$\,dB with the original CLS state);
\ours\ gives $28.32$--$29.07$\,dB ($29.06$\,dB with the original CLS state).
This conditioning-based diagnostic measures retained reconstruction ability.
The exact identity in \Cref{prop:lift-reconstruction} applies to the complete
frozen split with matching state, as specified in \S\ref{app:lift-construction}.

\paragraph{DiT-Base depth sweep.}
\Cref{tab:ditb_layer_sweep} reports separate 40-epoch DiT-Base runs on the
encoder target and each reported decoder target. Evaluation uses 50k samples,
50 steps, EMA, and ImageNet-training reference statistics. Guided sampling uses
internal guidance at scale $1.78$ on $t\in[0.1,1]$.
For the encoder-coordinate row, the same DiT samples are decoded by RAEv2 and
\ours; the decoder-target rows train separate DiTs on their respective frozen
decoder activations. This depth-sweep experiment is distinct from the
$p_{\mathrm{dit}}$ sweeps in \Cref{tab:scale}.

% Decoder layer sweep (CPU analysis, .fr_test/depth_sweep.py; results .fr_test/depth_{raev2k23,fusereg-p095}.json).
% 2048 ImageNet images (1536 fit / 512 held out), same encoder latent z (DINOv3-L, mean of L1-23 + L23 token mean).
% Features: per-position mean removed, per-channel standardized; ridge (lambda = 1e-2 * mean diag of X^T X), held-out R^2.
% Mean-CLS recon PSNR (128 extra images): RAEv2 28.54-28.55 at every depth (true-CLS 28.55);
%                                         FuseReg 28.32-29.07 (true-CLS 29.06) -> every h_k is (near-)lossless.
% CVPR two-column: table* spans both columns. requires: \usepackage{booktabs,graphicx}

\begin{table}[H]
\centering
\caption{Decoder layer sweep on RAEv2 $k{=}23$ latents. $\mathbf{h}_k$ is the hidden state after the first $k$ of the
28 decoder blocks ($\mathbf{h}_0$: decoder embedding + positional embedding). We report the held-out ridge $R^2$ of a
linear probe from each representation to the encoder latent $\mathbf{z}$ and to DINOv3-L layers L2 (low-level)
through L23 (semantic), and the participation ratio (PR) of the token covariance.
Probes use 2048 ImageNet images (1536 fit / 512 held out). A separate 128-image
suffix-decoding check with dataset-mean CLS conditioning retains approximately
28.5\,dB (RAEv2) and 28.3--29.1\,dB (FuseReg) PSNR across depths, measuring
reconstruction ability alongside the probes of linear accessibility.}
\label{tab:decoder_layer_sweep}
\setlength{\tabcolsep}{3.4pt}
\resizebox{\textwidth}{!}{%
\begin{tabular}{l ccccccc ccccccc}
\toprule
 & \multicolumn{7}{c}{RAEv2 decoder} & \multicolumn{7}{c}{FuseReg decoder ($p{=}0.95$)} \\
\cmidrule(lr){2-8}\cmidrule(lr){9-15}
 & & \multicolumn{5}{c}{$R^2$ to DINOv3 layer$\uparrow$} & &
   & \multicolumn{5}{c}{$R^2$ to DINOv3 layer$\uparrow$} & \\
\cmidrule(lr){3-7}\cmidrule(lr){10-14}
Latent & $R^2$ to $\mathbf{z}$ & L2 & L8 & L14 & L17 & L23 & PR
       & $R^2$ to $\mathbf{z}$ & L2 & L8 & L14 & L17 & L23 & PR \\
\midrule
$\mathbf{z}$      & --    & 0.301 & 0.415 & 0.627 & 0.665 & 0.684 & 59.8
                  & --    & 0.301 & 0.415 & 0.627 & 0.665 & 0.684 & 59.8 \\
\midrule
$\mathbf{h}_0$    & 0.957 & 0.300 & 0.410 & 0.604 & 0.637 & 0.665 & 49.9
                  & 0.969 & 0.301 & 0.411 & 0.611 & 0.646 & 0.672 & 29.1 \\
$\mathbf{h}_1$    & 0.836 & 0.349 & 0.449 & 0.586 & 0.599 & 0.628 & 100.6
                  & 0.951 & 0.319 & 0.427 & 0.619 & 0.651 & 0.670 & 64.1 \\
$\mathbf{h}_2$    & 0.726 & 0.392 & 0.473 & 0.534 & 0.531 & 0.567 & 106.8
                  & 0.925 & 0.329 & 0.435 & 0.620 & 0.648 & 0.662 & 74.9 \\
$\mathbf{h}_4$    & 0.580 & 0.418 & 0.488 & 0.429 & 0.412 & 0.463 & 146.0
                  & 0.842 & 0.354 & 0.464 & 0.602 & 0.618 & 0.628 & 105.3 \\
$\mathbf{h}_8$    & 0.444 & 0.462 & 0.425 & 0.315 & 0.295 & 0.357 & 135.4
                  & 0.664 & 0.389 & 0.495 & 0.508 & 0.504 & 0.517 & 156.4 \\
$\mathbf{h}_{14}$ & 0.353 & 0.494 & 0.347 & 0.227 & 0.211 & 0.282 & 103.7
                  & 0.475 & 0.447 & 0.454 & 0.365 & 0.352 & 0.375 & 198.9 \\
$\mathbf{h}_{20}$ & 0.309 & 0.486 & 0.296 & 0.182 & 0.169 & 0.244 & 95.1
                  & 0.382 & 0.502 & 0.375 & 0.270 & 0.252 & 0.301 & 167.5 \\
$\mathbf{h}_{27}$ & 0.301 & 0.481 & 0.288 & 0.175 & 0.163 & 0.237 & 121.4
                  & 0.328 & 0.495 & 0.317 & 0.217 & 0.207 & 0.260 & 52.5 \\
\bottomrule
\end{tabular}%
}
\end{table}

% DiT-B (40 ep) generation on decoder-layer latents h_k, RAEv2 k=23 encoder.
% 50K samples, 50 steps, EMA, ImageNet-train reference; guided = interval guidance 1.78 on t in [0.1,1].
% Sources (s3://tri-ml-sandbox-16011-us-west-2-datasets/raev3/sagemaker/):
%   RAEv2 z  : omnirobot-gfid-ditb-enc-raev2k23-ep40            (DiT-B on z, official RAEv2 decoder)
%   FuseReg z: omnirobot-gfid-ditb-enc-raev2k23-frp095-ep40     (SAME z DiT-B, decoded by FuseReg p0.95; decoder swap)
%   RAEv2 h_k: omnirobot-gfid-ditb-raev2k23-d{0,2,4,8}-ep40, h1 = ditb-raev2k23
%   FuseReg h_k: omnirobot-gfid-ditb-fusereg-p095{-d0,}-ep40;
%   FuseReg h2/h4/h8: omnirobot-gfid-ditb-fusereg-p095-d{2,4,8}-ep40
% requires: \usepackage{booktabs,graphicx}

\begin{table}[H]
\centering
\caption{DiT-B (40 epochs) trained on different decoder-layer latents of RAEv2 $k{=}23$.
$\mathbf{h}_k$: hidden state after the first $k$ decoder blocks of the corresponding decoder.
50K samples, 50 steps, EMA; guided results use internal guidance 1.78 on $t\in[0.1,1]$.
$^\dagger$The same $\mathbf{z}$-space DiT-B, decoded by the FuseReg decoder ($\mathbf{z}$ does not depend on the decoder).
Best per column in \textbf{bold}.}
\label{tab:ditb_layer_sweep}
\small
\setlength{\tabcolsep}{3.2pt}
\begin{tabular*}{\textwidth}{@{\extracolsep{\fill}}l rrrr rrrr@{}}
\toprule
 & \multicolumn{4}{c}{RAEv2 decoder} & \multicolumn{4}{c}{FuseReg decoder ($p{=}0.95$)} \\
\cmidrule(lr){2-5}\cmidrule(lr){6-9}
 & \multicolumn{2}{c}{w/o guid.} & \multicolumn{2}{c}{w/ guid.} & \multicolumn{2}{c}{w/o guid.} & \multicolumn{2}{c}{w/ guid.} \\
\cmidrule(lr){2-3}\cmidrule(lr){4-5}\cmidrule(lr){6-7}\cmidrule(lr){8-9}
Latent & gFID$\downarrow$ & IS$\uparrow$ & gFID$\downarrow$ & IS$\uparrow$ & gFID$\downarrow$ & IS$\uparrow$ & gFID$\downarrow$ & IS$\uparrow$ \\
\midrule
$\mathbf{z}$   & 13.83 & 109.9 & 4.66 & 169.9 & 11.75$^\dagger$ & \textbf{118.8}$^\dagger$ & 5.96$^\dagger$ & 163.9$^\dagger$ \\
$\mathbf{h}_0$ & \textbf{10.54} & \textbf{111.2} & \textbf{2.99} & \textbf{185.1} & \textbf{9.73} & 112.1 & \textbf{3.23} & \textbf{177.8} \\
$\mathbf{h}_1$ & 10.85 & 101.0 & 3.79 & 158.8 & 9.91 & 108.4 & 3.60 & 165.9 \\
$\mathbf{h}_2$ & 14.40 & 80.1  & 6.44 & 121.4 & 9.91 & 107.9 & 3.62 & 165.1 \\
$\mathbf{h}_4$ & 18.35 & 67.3  & 8.78 & 102.9 & 10.38 & 101.5 & 3.90 & 156.2 \\
$\mathbf{h}_8$ & 25.60 & 52.9  & 13.40 & 79.8 & 15.44 & 75.6 & 7.02 & 115.2 \\
\bottomrule
\end{tabular*}
\end{table}

\FloatBarrier

\FloatBarrier
\section{Prediction Error and Decoder Sensitivity}
\label{app:interface}

This section derives the common local framework in
\Cref{eq:interface-error-main} and relates the two interventions to it.
The method-specific identities are developed in
\S\ref{app:theory} and \S\ref{app:fuselift}.

\subsection{Local error propagation}
\label{app:interface-error}
\label{app:lift-local}

In \S\ref{sec:preliminaries}, fixed affine latent standardization was absorbed
into $\bar z$ and $D_\phi$ to keep main-text notation compact. Here we make the
standardized encoder target explicit as $\zeta=\mathcal N_z(\bar z)$ and write
$\mathcal D=D_\phi\circ\mathcal N_z^{-1}$ for its pixel rendering map. All state
required by the decoder is included in the coordinates or held fixed during
the perturbation. For a specified local prediction residual $\delta$, define
\begin{equation}
  G_D(\zeta)=J_{\mathcal D}(\zeta)^\top J_{\mathcal D}(\zeta),
  \qquad
  C_G(\zeta)=\mathbb E[\delta\delta^\top\mid\zeta].
  \label{eq:interface-matrices}
\end{equation}
For example, $\delta=g_\theta(\zeta_t,t,c)-\zeta$ can be a clean-target
prediction error at a selected denoising time; the conditional expectation
includes the noise and any other prediction randomness.
The second moment includes both variance and bias:
\begin{equation}
  C_G=\Sigma_G+b_Gb_G^\top,
  \qquad
  b_G=\mathbb E[\delta\mid\zeta],
  \quad \Sigma_G=\operatorname{Cov}(\delta\mid\zeta).
  \label{eq:interface-error-bias}
\end{equation}

Differentiability gives the deterministic local relation
\begin{equation}
  \|\mathcal D(\zeta+\delta)-\mathcal D(\zeta)\|_2^2
  =\delta^\top G_D(\zeta)\delta+o(\|\delta\|_2^2).
  \label{eq:lift-output-metric}
\end{equation}
The following bound justifies averaging this relation over local errors.

\par\noindent\begin{minipage}{\linewidth}
\begin{mainproposition}{Local image error couples decoder sensitivity and prediction error}{interface-error}
Fix $\zeta$ and suppose $\mathcal D$ is twice continuously differentiable
on a neighborhood of the closed ball $B_\rho(\zeta)$, with second derivative
bounded by $H$ in operator norm as a bilinear map. If
$\|\delta\|_2\le\rho$ almost surely conditional on $\zeta$, then
\begin{equation}
  \mathbb E\!\left[\|\mathcal D(\zeta+\delta)-\mathcal D(\zeta)\|_2^2
    \mid\zeta\right]
  =\operatorname{tr}\!\left(G_D(\zeta)C_G(\zeta)\right)+R_\zeta,
  \label{eq:interface-error-exact}
\end{equation}
where
\begin{equation}
  |R_\zeta|
  \le H\|J_{\mathcal D}(\zeta)\|_{\mathrm{op}}
      \mathbb E[\|\delta\|_2^3\mid\zeta]
    +\frac{H^2}{4}\mathbb E[\|\delta\|_2^4\mid\zeta].
  \label{eq:interface-error-remainder}
\end{equation}
No zero-mean assumption on the prediction error is needed.
\end{mainproposition}

\begin{insight}
For small latent errors, pixel changes depend on both the error's direction
and the decoder's sensitivity in that direction. This explains why similar
latent accuracy can produce different pixel errors, and motivates changing
either decoder sensitivity or generator errors.
\end{insight}
\end{minipage}\par

\begin{proof}
Write $J=J_{\mathcal D}(\zeta)$. Taylor's theorem gives
$\mathcal D(\zeta+\delta)-\mathcal D(\zeta)=J\delta+r$, with
$\|r\|_2\le H\|\delta\|_2^2/2$. The quadratic term obeys
\begin{equation}
  \mathbb E[\delta^\top J^\top J\delta\mid\zeta]
  =\operatorname{tr}\!\left(J^\top J
    \mathbb E[\delta\delta^\top\mid\zeta]\right).
\end{equation}
The remaining terms are $2\langle J\delta,r\rangle+\|r\|_2^2$.
Cauchy--Schwarz and the Taylor bound give
\Cref{eq:interface-error-remainder}.
\end{proof}

In particular,
\begin{equation}
  |R_\zeta|\le
  \left(H\|J_{\mathcal D}(\zeta)\|_{\mathrm{op}}\rho
    +\frac{H^2\rho^2}{4}\right)\operatorname{tr}(C_G(\zeta)).
\end{equation}
Thus the remainder is of smaller order than the latent error second moment
as the error neighborhood shrinks. When these terms are integrable,
averaging over clean targets gives
$\mathbb E_\zeta\operatorname{tr}(G_D(\zeta)C_G(\zeta))$ as the leading
term in image perturbation energy. This relation concerns a specified local
residual law; the experiments evaluate image quality on sampled distributions.

\subsection{Two interventions on the same interaction}
\label{app:interface-interventions}

\paragraph{FuseReg stabilizes responses to layer variation.}
Write $\mathcal N_z(v)=S_z(v-\mu_z)$ and
$\eta_S=\mathcal N_z(z_S)-\mathcal N_z(\bar z)$.
The subset identities in \Cref{lem:subset-mean} give
\begin{equation}
  \mathbb E_S[\eta_S\mid x]=0,
  \qquad
  C_S(x):=\mathbb E_S[\eta_S\eta_S^\top\mid x]
    =c_K(p)S_zV(x)S_z^\top.
  \label{eq:interface-subset-covariance}
\end{equation}
For a linear consumer $B$ acting on these coordinates, the additional
squared-loss term is exactly
\begin{equation}
  \mathbb E_{x,S}\|B\eta_S\|_2^2
  =\mathbb E_x\operatorname{tr}(B^\top B C_S(x)).
  \label{eq:interface-fusereg-penalty}
\end{equation}
For a linear image renderer, $B^\top B$ is its decoder metric. Here $C_S$
describes the structured training perturbations, while $C_G$ describes
prediction errors left by a trained generator. FuseReg uses the former to
encourage consistent decoding and prediction across layer subsets; its linear DiT penalty additionally
includes the time factor $\rho(t)t^2$ (\Cref{eq:dit-penalty-main}).

\paragraph{FuseLift changes the coordinates of prediction.}
For the normalized target map
$F_j=\mathcal N_j\circ T_j\circ\mathcal N_z^{-1}$, write
$\mathcal D=\mathcal R_j\circ F_j$ and
$G_{\mathcal R_j}=J_{\mathcal R_j}^\top J_{\mathcal R_j}$.
The chain rule gives
\begin{equation}
  G_D=J_{F_j}^\top G_{\mathcal R_j}J_{F_j}.
  \label{eq:interface-decoder-chain}
\end{equation}
The same local framework in the lifted coordinates pairs
$G_{\mathcal R_j}$ with the lifted generator's own error second moment.
For an affine prefix $F_j(\zeta)=A\zeta+a$, the training metric is
$M=A^\top A$, and the effective noise covariance on data-bearing directions
is $M^{-1}$ under the full-column-rank assumption
(\Cref{prop:lift-affine-geometry}). The complete image metric is
$A^\top G_{\mathcal R_j}A$.
Exactly corresponding errors have the same image effect under the chain
rule. FuseLift trains a new DiT with isotropic corruption and squared loss
in the selected coordinates, changing the prediction problem and hence
the residual errors attainable at a finite budget. This supplies the
coordinate intervention tested by the reconstruction-preserving comparisons.

\section{FuseReg: Mathematical Analysis and Proofs}
\label{app:theory}

This section analyzes the size-normalized, nonempty random-subset aggregation used
by \ours\ in \Cref{eq:drop}. We first derive properties of the sampling rule itself, then specialize
to linear squared-loss consumers, compare randomized and deterministic fusion, and
finally state how far these results extend to the nonlinear models used in practice.
The shared error-propagation framework and its relation to these penalties
are in \S\ref{app:interface}.

\paragraph{Scope at a glance.}
The conditioned subset probabilities, latent mean and covariance, support bound, and
per-layer relaxation for fixed subset losses are exact without assuming a linear
decoder or generator; the objective decompositions and closed-form optimizers require
homogeneous linear predictors and squared loss. They provide mechanism-level intuition
for the nonlinear ViT decoder and DiT rather than quantitative predictions for them.

\subsection{Mean preservation and disagreement covariance}
\label{app:normalized-moments}
\label{proof:subset-mean}

We write the sampling identities for vectorized layer features. Applying
the same fixed affine normalization to every layer preserves their form,
with the corresponding transformed moments
(\Cref{eq:interface-subset-covariance}). For linear DiT objectives, the
features and moments are understood in the actual normalized prediction
coordinates.

Fix $K\ge2$ flattened encoder-layer features $f_1(x),\ldots,f_K(x)\in\mathbb{R}^{Nd}$ with finite
second moments, and define their all-layer mean and deviations
\begin{equation}
  \bar z(x)=\frac1K\sum_{k=1}^K f_k(x),
  \qquad
  e_k(x)=f_k(x)-\bar z(x),
  \qquad
  \sum_{k=1}^K e_k(x)=0.
  \label{eq:consensus-split}
\end{equation}
For $p\in[0,1)$, let every layer be kept independently with probability $q=1-p$,
condition on the kept set $S$ being nonempty, and write
\begin{equation}
  z_S(x)=\frac1{|S|}\sum_{k\in S}f_k(x),
  \qquad
  V(x)=\frac1K\sum_{k=1}^K e_k(x)e_k(x)^\top .
  \label{eq:subset-mean}
\end{equation}
The population disagreement covariance is
\begin{equation}
  \Gamma=\mathbb E_x[V(x)].
  \label{eq:disagreement-covariance}
\end{equation}
The conditional probability of a particular nonempty subset configuration is
\begin{equation}
  \pi_p(S)
  =\frac{q^{|S|}p^{K-|S|}}{1-p^K}.
  \label{eq:subset-probability}
\end{equation}
Let $R=|S|$ under this conditioned law and define
\begin{equation}
  c_K(p)=
  \mathbb E\!\left[\frac{K-R}{R(K-1)}\right],
  \label{eq:ck-main}
\end{equation}

\par\noindent\begin{minipage}{\linewidth}
\begin{mainlemma}{Normalized subset fusion preserves the mean and isolates layer disagreement}{subset-mean}
For every fixed input $x$ and $K\ge2$,
\begin{equation}
  \begin{aligned}
    \mathbb E_S[z_S\mid x] &= \bar z,\\
    \mathbb E_S\!\left[(z_S-\bar z)(z_S-\bar z)^\top\mid x\right]
    &= c_K(p)V(x).
  \end{aligned}
  \label{eq:subset-moments-main}
\end{equation}
Consequently, the sampling-induced covariance averaged over inputs is
$c_K(p)\Gamma$.
\end{mainlemma}

\begin{insight}
Random subsets give different views centered on the full-layer average.
They vary along directions where layers disagree, so the decoder can learn
tolerance to layer changes without shifting the average representation.
\end{insight}
\end{minipage}\par

\begin{proof}[Proof of \Cref{lem:subset-mean}]
Conditioned on $R=r$, the kept set is uniform among all size-$r$ subsets. Therefore
\begin{equation}
  \Pr(k\in S\mid r)=\frac rK,
  \qquad
  \Pr(j,k\in S\mid r)=\frac{r(r-1)}{K(K-1)}\quad(j\ne k).
  \label{eq:subset-inclusion-probabilities}
\end{equation}
The conditional mean is
\begin{equation}
  \mathbb{E}_S[z_S\mid r]
  =\frac1r\frac rK\sum_k f_k
  =\bar z.
  \label{eq:subset-conditional-mean}
\end{equation}
For the conditional covariance, put $A_e=\sum_k e_ke_k^\top$ and use
$\sum_{j\ne k}e_je_k^\top=-A_e$:
\begin{equation}
  \begin{aligned}
    \mathbb{E}_S[(z_S-\bar z)(z_S-\bar z)^\top\mid r]
    &=\frac1{r^2}
      \left(
        \frac rK A_e
        +\frac{r(r-1)}{K(K-1)}(-A_e)
      \right)\\
    &=\frac{K-r}{r(K-1)}\,V(x).
  \end{aligned}
  \label{eq:subset-conditional-covariance}
\end{equation}
Let $\pi_r=\Pr(R=r)$ under the conditioned law above. By the law of total expectation,
\begin{equation}
  \begin{aligned}
    \mathbb E_S[z_S]
    &=\sum_{r=1}^K\pi_r\,\mathbb E_S[z_S\mid r]\\
    &=\sum_{r=1}^K\pi_r\,\bar z
      =\bar z.
  \end{aligned}
  \label{eq:subset-marginal-mean}
\end{equation}
Similarly, for the centered second moment,
\begin{equation}
  \begin{aligned}
    \mathbb E_S[(z_S-\bar z)(z_S-\bar z)^\top]
    &=\sum_{r=1}^K\pi_r\,\frac{K-r}{r(K-1)}V(x)\\
    &=\underbrace{\mathbb E_R\!\left[\frac{K-R}{R(K-1)}\right]}_{c_K(p)}V(x).
  \end{aligned}
  \label{eq:subset-marginal-covariance}
\end{equation}
These identities give \Cref{eq:subset-moments-main}.
\end{proof}

\par\noindent\begin{minipage}{\linewidth}
\begin{mainremark}{Drop rate controls disagreement variance}{cp}
Writing $q=1-p$ explicitly,
\begin{equation}
  c_K(p)=
  \frac{
    \sum_{r=1}^K
    \binom Kr q^r p^{K-r}
    \frac{K-r}{r(K-1)}
  }{1-p^K}.
  \label{eq:cp-explicit}
\end{equation}
Thus $c_K(0)=0$, $c_K(p)$ is analytic for $0<p<1$, and
$c_K(p)\to1$ as $p\to1$. Moreover, the conditioned subset size $R$ decreases in the usual stochastic
order as $p$ increases, while $(K-R)/(R(K-1))$ decreases with $R$; hence $c_K(p)$
is nondecreasing. Thus the bounded coefficient $c_K(p)\in[0,1]$ increases covariance
along layer-disagreement directions while leaving the expected latent equal to $\bar z$.
\end{mainremark}

\begin{insight}
A higher drop rate exposes the model to stronger differences between layers
while keeping the average representation fixed. It therefore controls the
amount of structured training variation without introducing a systematic shift.
\end{insight}
\end{minipage}\par

\subsection{Stage-specific objectives and optimal linear predictors}
\label{app:linear-surrogates}
\label{proof:stage-disagreement}

Let $y(x)$ have finite second moment and define
\begin{equation}
  \Sigma_{\bar z}=\mathbb{E}_x[\bar z\bar z^\top],
  \qquad
  \Gamma=\mathbb{E}_x[V(x)],
  \qquad
  A=\mathbb{E}_x[y\bar z^\top].
  \label{eq:moments}
\end{equation}

\paragraph{Stage-specific disagreement penalties.}
Using the disagreement covariance $\Gamma$ and scalar strength $c_K(p)$
defined above, the following proposition gives the exact objective
decompositions for linear predictors.

\par\noindent\begin{minipage}{\linewidth}
\begin{mainproposition}{Subset fusion induces stage-specific disagreement penalties}{stage-disagreement}
Let $D,W$ be homogeneous linear maps, $y$ a target with finite second moment,
and $\epsilon\sim\mathcal N(0,I)$ independent of $(x,S)$. Write
$\Omega(B)=\operatorname{tr}(B\Gamma B^\top)$.
For squared reconstruction and prediction of $\bar z$ from
$z_{t,S}=t z_S+(1-t)\epsilon$, the objectives satisfy
\begin{align}
  \mathcal L_p^{\mathrm{dec}}(D)
  &=\mathcal L_0^{\mathrm{dec}}(D)+c_K(p)\Omega(D),
  \label{eq:explicit-penalty}\\
  \mathcal L_{p,t}^{\mathrm{dit}}(W)
  &=\mathcal L_{0,t}^{\mathrm{dit}}(W)
    +\rho(t)t^2c_K(p)\Omega(W).
  \label{eq:dit-penalty-main}
\end{align}
Here $t\in(0,1]$, $\rho(t)$ is the weight in \Cref{eq:dit-objective}, and
$p=0$ denotes the corresponding full-fusion objective.
\end{mainproposition}

\begin{insight}
In the linear model, subset training discourages both networks from reacting
strongly to differences between encoder layers. The decoder learns to render
those variations consistently, while the DiT learns to recover the full-layer
target from varied inputs.
\end{insight}
\end{minipage}\par

\paragraph{Expanded objectives for \Cref{prop:stage-disagreement}.}
For the linear squared-loss surrogate, the decoder objective and its full-fusion
baseline are
\begin{equation}
\begin{aligned}
  \mathcal L_p^{\mathrm{dec}}(D)&=\mathbb E_{x,S}\|Dz_S-y\|^2,\\
  \mathcal L_0^{\mathrm{dec}}(D)&=\mathbb E_x\|D\bar z-y\|^2.
\end{aligned}
\end{equation}
For $z_{t,S}=t z_S+(1-t)\epsilon$ and
$\bar z_t=t\bar z+(1-t)\epsilon$, the corresponding DiT objectives are
\begin{equation}
\begin{aligned}
  \mathcal L_{p,t}^{\mathrm{dit}}(W)
  &=\rho(t)\mathbb E_{x,S,\epsilon}\|Wz_{t,S}-\bar z\|^2,\\
  \mathcal L_{0,t}^{\mathrm{dit}}(W)
  &=\rho(t)\mathbb E_{x,\epsilon}\|W\bar z_t-\bar z\|^2.
\end{aligned}
\end{equation}
The shared disagreement functional also satisfies
$\Omega(B)=K^{-1}\sum_k\mathbb E_x\|B(f_k-\bar z)\|^2$.

\begin{proof}[Proof of the decoder identity]
Write $z_S=\bar z+\delta_S$. By \Cref{lem:subset-mean},
\begin{equation}
  \mathbb E_S[\delta_S\mid x]=0,
  \qquad
  \mathbb E_S[\delta_S\delta_S^\top\mid x]=c_K(p)V(x).
  \label{eq:delta-moments}
\end{equation}
Conditioning on $x$ and expanding the square gives
\begin{align}
  \mathbb E_S\lVert Dz_S-y\rVert^2
  &=\mathbb E_S\lVert D\bar z-y+D\delta_S\rVert^2 \notag\\
  &=\lVert D\bar z-y\rVert^2
    +2(D\bar z-y)^\top D\,\mathbb E_S[\delta_S\mid x] \notag\\
  &\quad+\mathbb E_S[\delta_S^\top D^\top D\delta_S\mid x] \notag\\
  &=\lVert D\bar z-y\rVert^2
    +c_K(p)\operatorname{tr}\!\left(DV(x)D^\top\right).
  \label{eq:decoder-proof-expansion}
\end{align}
The cross term vanishes by the first identity in \Cref{eq:delta-moments}. For the
quadratic term, use $u^\top Bu=\operatorname{tr}(Buu^\top)$ and then the second
identity. Averaging \Cref{eq:decoder-proof-expansion} over $x$ replaces $V(x)$ by
$\Gamma$ and proves the decoder part of \Cref{prop:stage-disagreement}. Finally,
\begin{equation}
  \operatorname{tr}(D\Gamma D^\top)
  =\frac1K\sum_{k=1}^K\mathbb E_x
    \operatorname{tr}(De_ke_k^\top D^\top)
  =\frac1K\sum_{k=1}^K\mathbb E_x\lVert D(f_k-\bar z)\rVert^2,
\end{equation}
which gives the stated form of $\Omega(D)$.
\end{proof}

The same expansion gives the compact identity underlying the explanation in
\S\ref{sec:fusereg}:
\begin{equation}
  \mathcal L_p(D)=\mathcal L_0(D)
  +\mathbb E_{x,S}\bigl\|D\bigl(z_S(x)-\bar z(x)\bigr)\bigr\|_2^2.
  \label{eq:fusereg-sensitivity-main}
\end{equation}
The zero conditional mean removes the cross term, and
$c_K(p)\Omega(D)=\mathbb E_{x,S}\|D(z_S-\bar z)\|_2^2$ is the resulting
prediction-variation penalty.

\paragraph{Closed-form decoder.}
Expanding the resulting quadratic objective in $D$ gives
\begin{equation}
  \mathcal L_p^{\mathrm{dec}}(D)
  =\operatorname{tr}\!\left[D(\Sigma_{\bar z}+c_K(p)\Gamma)D^\top\right]
   -2\operatorname{tr}(AD^\top)+\mathbb E\lVert y\rVert^2.
\end{equation}
If $\Sigma_{\bar z}+c_K(p)\Gamma$ is positive definite, differentiating with respect
to $D$ gives the unique minimizer
\begin{equation}
  D_p^\star=A\bigl(\Sigma_{\bar z}+c_K(p)\Gamma\bigr)^{-1}.
  \label{eq:decoder-solution}
\end{equation}
The disagreement functional is zero exactly when
$D(f_k-\bar z)=0$ almost surely for every $k$; the closed form incorporates
this covariance-weighted sensitivity penalty.

\paragraph{Time weighting and clean-target prediction.}
RAEv2 parameterizes the DiT output as a clean-latent prediction and computes its
squared error after conversion to velocity~\citep{singh2026raev2}. Let $\tau$ denote
the implementation's data-to-noise time and set $t=1-\tau$ to match our
noise-to-data convention. The training time law $\nu$ is defined by
\begin{equation}
  u\sim\mathcal N(0,1),\qquad s=\operatorname{sigmoid}(u),\qquad
  \tau=\frac{a s}{1+(a-1)s},\qquad t=1-\tau,
  \label{eq:dit-time-law}
\end{equation}
with shift $a=8$ for our $256\times1024$ latents. Write
$d_\tau=\max(\tau,0.05)$ and $\rho(t)=d_\tau^{-2}$.
For a subset input $z_{t,S}$ and full-layer target $\bar z$, the implemented loss is
\begin{equation}
  \left\lVert
    \frac{z_{t,S}-g_\theta}{d_\tau}
    -\frac{z_{t,S}-\bar z}{d_\tau}
  \right\rVert^2
  =\rho(t)\lVert g_\theta-\bar z\rVert^2,
  \label{eq:dit-weighted-x}
\end{equation}
where $g_\theta=g_\theta(z_{t,S},t,c)$ denotes the clean-latent prediction at this input and time.

\begin{proof}[Proof of the DiT identity]
For fixed $t\in(0,1]$, write
\begin{equation}
  z_{t,S}=t(\bar z+\delta_S)+(1-t)\epsilon
  =\bar z_t+t\delta_S,
  \qquad
  \bar z_t=t\bar z+(1-t)\epsilon.
\end{equation}
Condition on $(x,\epsilon)$. Because the subset is sampled independently of
$\epsilon$, the conditional first and second moments of $\delta_S$ are still given by
\Cref{eq:delta-moments}. Therefore
\begin{align}
  \mathbb E_S\lVert Wz_{t,S}-\bar z\rVert^2
  &=\lVert W\bar z_t-\bar z\rVert^2
    +2t(W\bar z_t-\bar z)^\top W\mathbb E_S[\delta_S\mid x] \notag\\
  &\quad+t^2\mathbb E_S[\delta_S^\top W^\top W\delta_S\mid x] \notag\\
  &=\lVert W\bar z_t-\bar z\rVert^2
    +t^2c_K(p)\operatorname{tr}\!\left(WV(x)W^\top\right).
\end{align}
Averaging over $(x,\epsilon)$ and multiplying by the deterministic weight $\rho(t)$
proves the DiT part of \Cref{prop:stage-disagreement}. The factor $t^2$ appears because
subset variation is multiplied by $t$ before entering the linear predictor;
$\rho(t)$ preserves the velocity-space weighting of the clean-target error.
\end{proof}

\par\noindent\begin{minipage}{\linewidth}
\begin{mainproposition}{Closed-form linear consensus predictor}{consensus-shrinkage}
For $t\in(0,1]$, assume
$\Sigma_{\bar z}+c_K(p)\Gamma+\sigma_t^2 I$ is invertible
(which holds automatically for $0<t<1$). The optimal homogeneous linear predictor of $\bar z$ from $z_{t,S}$ under the weighted
loss in \Cref{eq:dit-penalty-main} is
\begin{equation}
  W_{p,t}^\star
  =
  \frac1t\,
  \Sigma_{\bar z}
  \left(
    \Sigma_{\bar z}
    +c_K(p)\Gamma
    +\sigma_t^2 I
  \right)^{-1},
  \qquad
  \sigma_t^2=\frac{(1-t)^2}{t^2}.
  \label{eq:dit-solution}
\end{equation}
If $\Sigma_{\bar z}$ and $\Gamma$ share an eigenvector with eigenvalues
$(\lambda,\gamma)$, the corresponding gain, apart from the common factor $1/t$, is
\begin{equation}
  \frac{\lambda}{\lambda+c_K(p)\gamma+\sigma_t^2}.
  \label{eq:consensus-gain}
\end{equation}
Along a shared eigendirection with nonzero signal ($\lambda>0$), increasing the drop rate suppresses the predictor gain whenever the layers disagree ($\gamma>0$), while a consensus direction ($\gamma=0$) is unaffected by the subset-induced penalty.
\end{mainproposition}

\begin{insight}
When signal and disagreement share directions in the linear model, increasing
the drop rate reduces reliance on features that vary across layers. Features
stable across layers receive no additional suppression from subset sampling.
\end{insight}
\end{minipage}\par

\begin{proof}
At fixed $t$, the positive scalar $\rho(t)$ cancels from the normal equation for
predicting $\bar z$ from $z_{t,S}$:
\begin{equation}
  W\,\mathbb E[z_{t,S}z_{t,S}^\top]
  =\mathbb E[\bar z z_{t,S}^\top].
  \label{eq:dit-normal-equation}
\end{equation}
Mean preservation and \Cref{eq:delta-moments} imply
\begin{equation}
  \mathbb{E}[\bar z z_S^\top]=\Sigma_{\bar z},
  \qquad
  \mathbb{E}[z_Sz_S^\top]=\Sigma_{\bar z}+c_K(p)\Gamma.
\end{equation}
Because $\epsilon$ is independent, zero mean, and has identity covariance, all cross
moments containing one factor of $\epsilon$ vanish. Hence
\begin{align}
  \mathbb{E}[\bar z z_{t,S}^\top]
    &=t\Sigma_{\bar z},\\
  \mathbb{E}[z_{t,S}z_{t,S}^\top]
    &=t^2\bigl(\Sigma_{\bar z}+c_K(p)\Gamma\bigr)+(1-t)^2I.
\end{align}
For $0<t<1$, the positive noise term makes this second-moment matrix positive
definite, since $\Sigma_{\bar z}$ and $c_K(p)\Gamma$ are positive semidefinite.
Substituting these moments into \Cref{eq:dit-normal-equation}, factoring out $t^2$,
and using invertibility yields \Cref{eq:dit-solution}. On a shared eigenvector, the
matrix factor acts as scalar multiplication by
$\lambda/(\lambda+c_K(p)\gamma+\sigma_t^2)$, which proves
\Cref{eq:consensus-gain}.
\end{proof}

Both DiT output heads are trained against the full-layer target. For linear heads
$W_{\mathrm{full},t}$ and $W_{\mathrm{base},t}$, with base-head coefficient $\beta=1$
in our experiments, averaging their losses over $t\sim\nu$ gives the added penalty
\begin{equation}
  c_K(p_{\mathrm{dit}})\mathbb E_{t\sim\nu}\!\left[
    \rho(t)t^2\left\{
      \Omega(W_{\mathrm{full},t})+\beta\Omega(W_{\mathrm{base},t})
    \right\}\right],
  \label{eq:dit-two-head-penalty}
\end{equation}
where $\Omega(W)=\operatorname{tr}(W\Gamma W^\top)$.
The time law and weight therefore determine how strongly each time contributes to
training, while the fixed-time closed form remains unchanged. The same base
disagreement matrix $\Gamma$ enters both stages: the decoder pays an invariance
penalty against a pixel target, and the DiT pays a time-weighted penalty while
predicting the full-layer consensus from a noisy subset input.

\Needspace{8\baselineskip}
\subsection{Guidance and the joint sensitivity of two heads}
\label{app:guidance-coupling}

Generator-side subset regularization introduces a coupling between the two latent
predictions used by RAEv2's internal guidance~\citep{singh2026raev2}, which combines
the full DiT output with an intermediate REPA head. The fixed-generator decoder-swap
trajectories in \Cref{fig:decoder-swap,tab:ditxl_all} use $p_{\mathrm{dit}}{=}0$
and are outside the scope of this regularization analysis. Write the guided prediction as
\[
\hat{z}_{\mathrm{guided}}
=\hat{z}_{\mathrm{full}}
+w\bigl(\hat{z}_{\mathrm{full}}-\hat{z}_{\mathrm{repa}}\bigr).
\]
Changes in the two heads' relative responses to layer subsets are scaled by the guidance
weight $w$, even when each prediction becomes individually more stable.

The linear surrogate in \Cref{eq:dit-penalty-main} makes this coupling explicit.
Let $W_{\mathrm{full}}$ and $W_{\mathrm{repa}}$ denote the full and REPA predictors, and
write $\Omega(W)=\operatorname{tr}(W\Gamma W^\top)$. Their guided combination
$W_{\mathrm{guided}}=(1+w)W_{\mathrm{full}}-wW_{\mathrm{repa}}$ incurs disagreement penalty
\[
\begin{aligned}
\Omega(W_{\mathrm{guided}})
&=(1+w)^2\Omega(W_{\mathrm{full}})+w^2\Omega(W_{\mathrm{repa}})\\
&\quad-2w(1+w)\operatorname{tr}
  \bigl(W_{\mathrm{full}}\Gamma W_{\mathrm{repa}}^{\top}\bigr),
\end{aligned}
\]
scaled by $\rho(t)t^2c_K(p_{\mathrm{dit}})$.
The cross term measures alignment along layer-disagreement directions: aligned responses partially cancel, whereas opposing responses reinforce one another. Reducing $W_{\mathrm{full}}$'s sensitivity alone therefore need not reduce the guided combination's sensitivity, matching the guidance-dependent rate preferences in \Cref{tab:scale-xl,tab:scale-xl-guided}.

\subsection{Coverage of nonempty subsets}
\label{app:coverage}

\par\noindent\begin{minipage}{\linewidth}
\begin{maincorollary}{Probability-weighted control of every subset}{coverage}
Let $\ell_S(f)\ge0$ denote the expected loss of any predictor $f$ conditional on the
kept set being $S$. For every $0<p<1$ and every nonempty $S_0$,
\begin{equation}
  \ell_{S_0}(f)
  \le
  \frac{\mathcal L_p(f)}{\pi_p(S_0)},
  \qquad
  \mathcal L_p(f)=\sum_{S\ne\emptyset}\pi_p(S)\ell_S(f).
  \label{eq:coverage}
\end{equation}
The statement applies to any nonnegative reconstruction loss and to the weighted squared DiT
training loss conditioned on $S_0$ (including its expectations over $t$ and
$\epsilon$).
\end{maincorollary}

\begin{insight}
Training over every nonempty subset links average performance to performance
on individual fusions. The control is strongest for frequently sampled subsets
and becomes looser for rare ones.
\end{insight}
\end{minipage}\par

\begin{proof}
Every term in the sum defining $\mathcal L_p(f)$ is nonnegative, so
$\mathcal L_p(f)\ge\pi_p(S_0)\ell_{S_0}(f)$.
\end{proof}

We evaluate reconstruction across layer subsets in \Cref{fig:sweep-all}.

\par\noindent\begin{minipage}{\linewidth}
\begin{mainproposition}{A fixed fusion need not transfer to an adjacent subset}{separation}
For any loss threshold $M>0$, there is a scalar $K=2$ squared-error problem on which
a decoder has zero loss on the all-layer fusion but loss greater than $M$ on a
Hamming-adjacent singleton fusion.
\end{mainproposition}

\begin{insight}
Two layers can cancel each other's errors in their average, making full-fusion
reconstruction look perfect. Removing one layer exposes the hidden error,
which shows why success on one fusion alone does not establish robustness to others.
\end{insight}
\end{minipage}\par

\begin{proof}
Fix $\eta>0$. Let $y\sim\mathcal N(0,1)$ and
$g\sim\mathcal N(0,\sigma^2)$ be independent, and set
$f_1=g$ and $f_2=\eta y-g$. The normalized all-layer fusion is
$\bar z=\eta y/2$, so $D_{\mathrm{full}}=2/\eta$ has zero loss. On
$S=\{1\}$, its loss is
$4\sigma^2/\eta^2+1$, which exceeds $M$ for a suitable $\sigma$.
\end{proof}

\Needspace{10\baselineskip}
\subsection{Separation from deterministic global fusion}
\label{app:obstruction}
\label{sec:obstruction}
\label{proof:rank-one}

For a general input-independent selection rule, let $w\in\mathbb{R}^K$ be its random
layer-weight vector and $z_w=\sum_k w_kf_k$. Define
$M_1=\mathbb{E}[w]$ and $M_2=\mathbb{E}[ww^\top]$.

\par\noindent\begin{minipage}{\linewidth}
\begin{mainlemma}{The linear objective depends only on two weight moments}{moments}
For a homogeneous linear consumer and squared error,
\begin{align}
  \mathcal L(D;M_1,M_2)
  &=
  \operatorname{tr}\!\left(D\Sigma(M_2)D^\top\right)
  -2\operatorname{tr}\!\left(B(M_1)D^\top\right)
  +\mathbb{E}_x\|y\|^2, \\
  \Sigma(M_2)
  &:=
  \sum_{j,k}[M_2]_{jk}\,\mathbb{E}_x[f_jf_k^\top],
  \qquad
  B(M_1):=
  \sum_k[M_1]_k\,\mathbb{E}_x[yf_k^\top].
  \label{eq:moment-objective}
\end{align}
Thus the selection rule enters the linear objective only through $(M_1,M_2)$.
\end{mainlemma}

\begin{insight}
For linear prediction with squared error on fixed features, only average layer
weights and their joint variation determine the objective. This lets us compare
fusion rules through these summaries instead of enumerating every possible subset.
\end{insight}
\end{minipage}\par

\begin{proof}
Because the selection rule is input independent, $w$ and $x$ are independent. Expanding
the square gives
\begin{align}
  \mathcal L(D)
  &=\mathbb E\!\left[
      \left\lVert D\sum_k w_kf_k-y\right\rVert^2
    \right] \notag\\
  &=\sum_{j,k}\mathbb E[w_jw_k]
      \operatorname{tr}\!\left(D\mathbb E_x[f_jf_k^\top]D^\top\right) \notag\\
  &\quad-2\sum_k\mathbb E[w_k]
      \operatorname{tr}\!\left(\mathbb E_x[yf_k^\top]D^\top\right)
      +\mathbb E_x\lVert y\rVert^2.
\end{align}
Substituting $[M_2]_{jk}=\mathbb E[w_jw_k]$ and
$[M_1]_k=\mathbb E[w_k]$, then collecting the two sums into
$\Sigma(M_2)$ and $B(M_1)$, yields \Cref{eq:moment-objective}.
\end{proof}

\paragraph{Random-subset moments versus fixed fusion.}
Let $w_S=\mathbf1_S/|S|$ and $P=I-K^{-1}\mathbf1\mathbf1^\top$.

\par\noindent\begin{minipage}{\linewidth}
\begin{mainproposition}{No deterministic global fusion matches random-subset weight moments}{rank-one}
For $K\ge2$ and $0<p<1$,
\begin{equation}
  \mathbb E[w_S]=\frac1K\mathbf1,
  \quad
  \mathbb E[w_Sw_S^\top]
  =\frac1{K^2}\mathbf1\mathbf1^\top+\frac{c_K(p)}K P.
  \label{eq:weight-moments-main}
\end{equation}
The second moment in \Cref{eq:weight-moments-main} has rank $K$, while every
deterministic input-independent fusion has second moment $ww^\top$ of rank at most
$1$. Hence no fixed global fusion matches both moments of normalized subset sampling.
\end{mainproposition}

\begin{insight}
A fixed global fusion can match the average layer weights, but those weights
never vary across training examples. Random subsets vary the relative layer
contributions, supplying a training signal that fixed weights cannot reproduce
in general.
\end{insight}
\end{minipage}\par

\begin{proof}[Proof of \Cref{prop:rank-one}]
For normalized random subsets, put $P=I-\frac1K\mathbf1\mathbf1^\top$. The weight
vector $w_S=\mathbf1_S/|S|$ is exchangeable and satisfies
$\mathbf1^\top w_S=1$, so $\mathbb{E}[w_S]=K^{-1}\mathbf1$. Exchangeability makes
all diagonal covariance entries equal and all off-diagonal entries equal, while
$\mathbf1^\top w_S=1$ implies that the covariance annihilates $\mathbf1$. It must
therefore have the form $\operatorname{Cov}(w_S)=\alpha P$ for some scalar $\alpha$.
To identify $\alpha$, note that
\[
  \operatorname{tr}\!\left(\operatorname{Cov}(w_S)\right)
  =
  \mathbb{E}\!\left[\frac1R\right]-\frac1K
  =
  \frac{K-1}{K}c_K(p),
\]
which gives $\alpha=c_K(p)/K$. Hence
\begin{equation}
  \mathbb{E}[w_S]=\frac1K\mathbf1,
  \qquad
  \operatorname{Cov}(w_S)=\frac{c_K(p)}K P,
  \qquad
  M_2=
  \frac1{K^2}\mathbf1\mathbf1^\top+\frac{c_K(p)}K P.
  \label{eq:normalized-weight-moments}
\end{equation}

The vector $\mathbf1$ is an eigenvector of $M_2$ with eigenvalue $1/K$, while every
vector orthogonal to $\mathbf1$ has eigenvalue $c_K(p)/K>0$. Thus $M_2$ has rank $K$.
The deterministic claim follows from the rank of an outer product.
\end{proof}

For a given feature distribution, \Cref{lem:moments} shows that two selection rules
induce distinct objectives as functions of $D$ exactly when their induced
$\Sigma(M_2)$ or $B(M_1)$ differs. Thus \Cref{prop:rank-one} establishes separation
of the weight moments; feature moments determine whether this separation persists
in the objective. It covers fixed subsets and globally learned layer weights. It does not
cover an input-dependent router, whose weights are correlated with $x$ and therefore
fall outside the independent selection-rule model used by \Cref{lem:moments}.

\Needspace{12\baselineskip}
\subsection{Per-layer rates continuously relax subset selection}
\label{app:relaxation}

Let $r=(r_1,\ldots,r_K)\in[0,1]^K\setminus\{\mathbf0\}$ contain per-layer keep
probabilities and define
\begin{equation}
  Z(r)=1-\prod_{k=1}^K(1-r_k),
  \qquad
  \pi_r(S)=
  \frac{
    \prod_{k\in S}r_k
    \prod_{k\notin S}(1-r_k)
  }{Z(r)}
  \quad(S\ne\emptyset).
\end{equation}
For fixed subset losses $\ell(S)$, the exact conditioned Bernoulli extension is
\begin{equation}
  F(r)
  =
  \sum_{\emptyset\ne S\subseteq[K]}\pi_r(S)\ell(S)
  =
  \frac{
    \sum_{\emptyset\ne S\subseteq[K]}
    \ell(S)\prod_{k\in S}r_k\prod_{k\notin S}(1-r_k)
  }{
    1-\prod_k(1-r_k)
  }.
  \label{eq:conditioned-relaxation}
\end{equation}
The numerator is multi-affine; conditioning makes $F$ a rational rather than a
multilinear function.

\paragraph{Recovery of deterministic subsets.}
For every nonempty $S_0\subseteq[K]$,
\begin{equation}
  F(\mathbf1_{S_0})=\ell(S_0).
\end{equation}
At this vertex, the conditional distribution places unit mass on $S_0$; hence the
nonzero vertices of the per-layer rate hypercube recover every nonempty deterministic
subset exactly.

\paragraph{Interior smoothness.}
The function $F$ in \Cref{eq:conditioned-relaxation} is real-analytic on $(0,1)^K$
because its numerator and denominator are polynomials and the denominator is positive
there. On the
exchangeable line $r_1=\cdots=r_K=1-p$, the moments in
\Cref{lem:subset-mean} and the optimal linear predictors in
\Cref{eq:decoder-solution,eq:dit-solution} are analytic for $0<p<1$ whenever their
second-moment matrices remain invertible: $c_K(p)$ is a ratio of polynomials with a
positive denominator, and matrix inversion is analytic on nonsingular matrices.

The full $K$-dimensional rate vector is the relaxation whose nonzero vertices represent
all nonempty subsets. \ours\ uses the exchangeable diagonal
$r_1=\cdots=r_K=1-p$ at each stage. Consequently
$(p_{\mathrm{dec}},p_{\mathrm{dit}})$ is a structured two-dimensional slice of the
$2K$-dimensional product relaxation, not an exact parameterization of every subset
pair.

\subsection{From exact identities to nonlinear-model intuition}
\label{app:scope}

The subset distribution, mean-preservation and covariance identities in
\Cref{lem:subset-mean}, and the full-support bound in \Cref{cor:coverage} do not assume
a linear consumer. Likewise, the vertex and analyticity statements in
\S\ref{app:relaxation} are exact for fixed subset losses. By contrast,
\Cref{prop:stage-disagreement,prop:consensus-shrinkage} assume homogeneous linear predictors
and squared loss; the eigendirection interpretation additionally assumes shared
eigenvectors. For a nonlinear consumer, latent mean preservation does not imply
prediction mean preservation or an exact disagreement penalty. We therefore use the
linear results as mechanism-level intuition and evaluate the actual ViT decoder,
mixed reconstruction loss, nonlinear DiT, and generation metrics empirically. In
particular, the analysis neither predicts the sign of the two-rate contrast nor
establishes super-additivity.

\section{FuseLift: Mathematical Analysis and Proofs}
\label{app:fuselift}

This section derives the reconstruction identity, affine noise and loss
decompositions, and nonlinear extensions behind \S\ref{sec:fuselift}.
The complete construction is specified in \S\ref{app:fuselift-method}.
The common prediction-error framework and local remainder bound are in
\S\ref{app:interface}; the empirical motivation is in
\S\ref{sec:fusereg-insight}.
\Cref{hyp:lift-preference} concerns finite-budget modeling and directional
alignment; the propositions below establish the underlying geometric identities.

\subsection{Prediction-error alignment and target depth}
\label{app:lift-preference}

\par\noindent\begin{minipage}{\linewidth}
\begin{mainhypothesis}{Prediction-error alignment and target depth}{lift-preference}
At a fixed DiT capacity and training budget, generation improves when the
denoising objective prioritizes accuracy along directions to which the pixel
decoder is most sensitive. A shallow decoder target promotes this alignment through its
induced error metric and corruption geometry, whereas deeper decoder activations trade
linear semantic accessibility for local pixel synthesis and become harder for
the same DiT to model.
\end{mainhypothesis}

\begin{insight}
A shallow decoder target may help a generator at a fixed training budget focus
on directions that matter for image reconstruction. The hypothesis is that
moving deeper eventually makes useful semantic structure harder for the same
generator to model.
\end{insight}
\end{minipage}\par

\subsection{Reconstruction invariance}
\label{app:lift-reconstruction-proof}

Use the complete frozen split $D_\phi=R_j\circ T_j$ and target normalization $\mathcal N_j$ defined in
\Cref{eq:lift-split,eq:lift-normalized-target}, with $\mathcal R_j=R_j\circ\mathcal N_j^{-1}$.

\par\noindent\begin{minipage}{\linewidth}
\begin{mainproposition}{Reconstruction is independent of target placement}{lift-reconstruction}
For any frozen decomposition in \Cref{eq:lift-split} and invertible target normalization in \Cref{eq:lift-normalized-target},
\begin{equation}
\begin{aligned}
  \mathcal R_j(u_j(x))
  &=R_j(T_j(\bar z(x)))\\
  &=D_\phi(\bar z(x))\quad\forall x.
\end{aligned}
  \label{eq:h1-latent}
\end{equation}
Consequently, if $P_z=\bar z_\#P_x$ and
$P_{u_j}=(\mathcal N_j\circ T_j)_\#P_z$, then
\begin{equation}
  (\mathcal R_j)_\#P_{u_j}=(D_\phi)_\#P_z.
  \label{eq:lift-reconstruction-distribution}
\end{equation}
The prefix $T_j$ need not be invertible.
\end{mainproposition}

\begin{insight}
Stopping a frozen decoder at an intermediate activation and then resuming its
remaining layers produces the same clean reconstruction. Moving the diffusion
target changes what the generator learns while keeping this reference
reconstruction fixed.
\end{insight}
\end{minipage}\par
\begin{proof}
Substituting $u_j(x)=\mathcal N_j(T_j(\bar z(x)))$ cancels $\mathcal N_j^{-1}\circ\mathcal N_j$, leaving $R_j\circ T_j=D_\phi$. Equality of the pushforward reconstruction distributions follows immediately.
\end{proof}

Thus the original and relocated targets yield identical real-data reconstructions, including all paired metrics and rFID. Their learned sampling distributions nevertheless differ: encoder-space generation renders its learned latent distribution through $D_\phi$, whereas FuseLift renders its learned distribution $Q_{u_j}$ through $\mathcal R_j$.

\subsection{An affine relocation already changes diffusion}
\label{sec:lift-affine}

Consider an extraction point whose frozen prefix is affine on flattened features in $\mathbb R^n$ ($n=Nd$), such as the input projection at $h_0$ together with fixed affine preprocessing:
\begin{equation}
  h_0=W_0\bar z+b_0.
  \label{eq:lift-affine-prefix}
\end{equation}
A tokenwise projection corresponds to repeated block structure in $W_0$, and fixed positional embeddings are absorbed into $b_0$. Input-dependent normalization prior to the split is covered by the nonlinear analysis in \S\ref{sec:h1-interpretation}.

Writing explicit affine standardizations $\mathcal N_z(v)=S_z(v-\mu_z)$ and $\mathcal N_0(h)=S_0(h-\mu_0)$, let $\zeta=\mathcal N_z(\bar z)\in\mathbb R^n$ denote the standardized encoder target and $u_0=\mathcal N_0(h_0)\in\mathbb R^m$ the standardized lifted target. Then
\begin{equation}
\begin{aligned}
  u_0&=A\zeta+a,\qquad A=S_0W_0S_z^{-1},\\
  a&=S_0(W_0\mu_z+b_0-\mu_0).
\end{aligned}
\label{eq:lift-effective-affine}
\end{equation}
The effective linear map $A$ thus incorporates the standardization of both spaces.

\par\noindent\begin{minipage}{\linewidth}
\begin{mainproposition}{The induced noise covariance and error metric}{lift-affine-geometry}
Suppose $A\in\mathbb R^{m\times n}$ has full column rank. Define
$M=A^\top A\in\mathbb R^{n\times n}$, the left pseudoinverse
$A^\dagger=M^{-1}A^\top$, and the orthogonal projector
$\Pi=AA^\dagger$. For the flow-matching interpolant at $t\in(0,1)$,
\begin{equation}
  u_{0,t}=t(A\zeta+a)+(1-t)\epsilon,
  \quad \epsilon\sim\mathcal N(0,I_m),
  \label{eq:lift-affine-noising}
\end{equation}
the pulled-back coordinates on the data-bearing subspace satisfy
\begin{equation}
\begin{aligned}
  A^\dagger(u_{0,t}-t a)
    &=t\zeta+(1-t)\xi,\\
  \xi&=A^\dagger\epsilon\sim\mathcal N(0,M^{-1}).
\end{aligned}
\label{eq:lift-effective-noise}
\end{equation}
For any clean-target prediction $\hat u_0$, define
$\hat\zeta=A^\dagger(\hat u_0-a)$ and $\delta=\hat\zeta-\zeta$. Its squared target error decomposes exactly as
\begin{equation}
\begin{aligned}
  \|\hat u_0-u_0\|_2^2
  &=\delta^\top M\delta
    +\|(I_m-\Pi)(\hat u_0-a)\|_2^2.
\end{aligned}
\label{eq:lift-loss-decomposition}
\end{equation}
\end{mainproposition}

\begin{insight}
An information-preserving affine projection makes prediction errors cost more
in directions it amplifies and reduces effective noise there. It also penalizes
predictions that leave the transformed latent space.
\end{insight}
\end{minipage}\par
\begin{proof}
Full column rank gives $A^\dagger A=I_n$ and
$A^\dagger(A^\dagger)^\top=M^{-1}$, proving \Cref{eq:lift-effective-noise}. Since $u_0-a=A\zeta$, decomposing
$\hat u_0-u_0=A(\hat\zeta-\zeta)+(I_m-\Pi)(\hat u_0-a)$ into orthogonal range and nullspace components yields \Cref{eq:lift-loss-decomposition}.
\end{proof}

For predictions lying in the affine range of $A$, the second term in \Cref{eq:lift-loss-decomposition} vanishes, leaving $\|A\delta\|_2^2=\delta^\top M\delta$. High-gain directions of $A$ receive larger squared-error weight and smaller pulled-back noise variance simultaneously.

\paragraph{Rectangular and rank-deficient projections.}
When $m>n$ with full column rank, $(I_m-\Pi)(u_{0,t}-t a)=(1-t)(I_m-\Pi)\epsilon$ carries pure noise orthogonal to the data subspace, and the second term of \Cref{eq:lift-loss-decomposition} penalizes residual off-subspace predictions. When $A$ is rank-deficient, the same identities hold on $\ker(A)^\perp$ with the Moore--Penrose pseudoinverse $A^\dagger$, while components in $\ker(A)$ do not affect $u_0$; exact real-data reconstruction still holds by \Cref{prop:lift-reconstruction} because $D_\phi$ itself factors through $T_0$.

\paragraph{Expressivity versus finite-budget optimization.}
Although sufficiently wide input/output projections could absorb an invertible coordinate change in principle, isotropic noise injection and Euclidean squared loss in $u_0$-coordinates alter both the conditioning noise covariance and the gradient weighting (\Cref{eq:lift-effective-noise,eq:lift-loss-decomposition}), changing finite-budget optimization unless $A$ has orthonormal columns up to scale.

\subsection{Directional signal-to-noise ratios}
\label{sec:lift-snr}

Write the thin singular value decomposition $A=U\operatorname{diag}(s_1,\dots,s_n)V^\top$ ($s_i>0$), with right singular vectors $v_i$. Projecting \Cref{eq:lift-affine-noising} onto the left singular basis $\widetilde u_t=U^\top(u_{0,t}-t a)$ gives
\begin{equation}
  \widetilde u_{t,i}
  =t s_i\langle v_i,\zeta\rangle
    +(1-t)\widetilde\epsilon_i,
  \label{eq:lift-singular-channel}
\end{equation}
with $\widetilde\epsilon_i\stackrel{\mathrm{iid}}{\sim}\mathcal N(0,1)$. Writing $C_z=\operatorname{Cov}(\zeta)$, the marginal signal-to-noise ratio along singular direction $i$ at time $t\in(0,1)$ is
\begin{equation}
  \operatorname{SNR}_i(t)
  =\frac{t^2}{(1-t)^2}\,s_i^2\,v_i^\top C_zv_i,
  \label{eq:lift-directional-snr}
\end{equation}
and the data-subspace squared-error metric is
\begin{equation}
  \|A\delta\|_2^2
  =\sum_{i=1}^n s_i^2\langle v_i,\delta\rangle^2.
  \label{eq:lift-singular-metric}
\end{equation}

\paragraph{Why channel standardization does not cancel directional gains.}
Per-channel standardization fixes marginal channel variances without removing cross-channel correlations. Consider a scalar latent coordinate $s$ replicated across $r$ standardized channels, each corrupted independently as $u_{t,q}=t s+(1-t)\epsilon_q$ ($q=1,\dots,r$). Their average satisfies
\begin{equation}
  \frac1r\sum_{q=1}^r u_{t,q}
  =t s+\frac{1-t}{\sqrt{r}}\eta,
  \quad \eta\sim\mathcal N(0,1),
  \label{eq:lift-replication}
\end{equation}
multiplying both the directional SNR and the squared cost of a coherent error $\delta s$ by $r$. Singular values $s_i$ generalize this mechanism across correlated feature directions. Likewise, a scalar noise-schedule shift~\citep{zheng2025rae} rescales all channels in \Cref{eq:lift-directional-snr} by one time-dependent factor and cannot reproduce anisotropic weights $s_i^2$.

\subsection{Spectral concentration and decoder alignment}
\label{sec:lift-spectrum}

Define the participation ratio of a nonzero covariance matrix $C$ as
\begin{equation}
  \operatorname{PR}(C)
  =\frac{(\operatorname{tr}C)^2}{\operatorname{tr}(C^2)}.
  \label{eq:lift-pr}
\end{equation}
For $C_{u_0}=AC_zA^\top$, in the isotropic reference case $C_z=I_n$,
\begin{equation}
  \operatorname{PR}(C_{u_0})
  =\frac{(\sum_i s_i^2)^2}{\sum_i s_i^4}.
  \label{eq:lift-pr-spectrum}
\end{equation}
Under the protocol in \S\ref{app:lift-depth-protocol}, the encoder target $\bar z$ has participation ratio $59.8$, compared with $49.9$ for RAEv2 $h_0$ and $29.1$ for \ours\ $h_0$ (\Cref{tab:decoder_layer_sweep}), confirming stronger spectral concentration after the trained input projection.

\paragraph{Relating the prefix metric to full pixel sensitivity.}
Writing $\mathcal D=\mathcal R_0\circ F_0$ with $F_0(\zeta)=A\zeta+a$, the chain rule gives
\begin{equation}
  G_D=J_{\mathcal D}^\top J_{\mathcal D}
  =A^\top J_{\mathcal R_0}^\top J_{\mathcal R_0}A.
  \label{eq:lift-decoder-metric}
\end{equation}
FuseLift weights denoising errors by $M=A^\top A$, serving as a prefix-derived proxy for $G_D=A^\top G_{\mathcal R_0}A$; alignment along singular vectors $v_i$ can be evaluated by comparing $s_i^2$ against $\mathbb E_\zeta\|J_{\mathcal D}(\zeta)v_i\|_2^2$.

\subsection{Nonlinear relocation to deeper decoder blocks}
\label{sec:h1-interpretation}

For a general split depth $j\ge 1$ (such as $h_1$, or $h_2,h_4,h_8$ in the depth sweep), define the normalized prefix map
\begin{equation}
  F_j=\mathcal N_j\circ T_j\circ\mathcal N_z^{-1}.
  \label{eq:lift-nonlinear-map}
\end{equation}
For a small perturbation $\delta$ around $\zeta$, differentiability gives
\begin{equation}
\begin{aligned}
  \|F_j(\zeta+\delta)-F_j(\zeta)\|_2^2
  &=\delta^\top M_j(\zeta)\delta
    +o(\|\delta\|_2^2),\\
  M_j(\zeta)&=J_{F_j}(\zeta)^\top J_{F_j}(\zeta),
\end{aligned}
\label{eq:jacobian-metric}
\end{equation}
which generalizes $M=A^\top A$ locally while incorporating both normalizations.

\paragraph{Noisy trajectories and suffix decoding.}
For nonlinear $F_j$, corrupting the lifted target differs from pushing forward a corrupted encoder latent:
\begin{equation}
  F_j\bigl(t\zeta+(1-t)\epsilon_z\bigr)
  \ne t F_j(\zeta)+(1-t)\epsilon_u
  \quad\text{in general}.
\label{eq:lift-path-change}
\end{equation}
At inference, encoder-space samples follow $\hat\zeta\mapsto\mathcal D(\hat\zeta)$, whereas FuseLift samples bypass $T_j$ and enter the suffix directly via $\hat u_j\mapsto\mathcal R_j(\hat u_j)$.

\subsection{Connection to layer-disagreement regularization}
\label{sec:lift-fusereg-connection}

In standardized encoder coordinates, let $\zeta_S=\mathcal N_z(z_S)$, $\zeta=\mathcal N_z(\bar z)$, and $\eta_S=\zeta_S-\zeta$, with $\mathbb E_S[\eta_S\mid x]=0$ and $\operatorname{Cov}(\eta_S\mid x)=C_S(x)$ from \Cref{eq:interface-subset-covariance}. Under any frozen affine map $F(\zeta)=A\zeta+a$,
\begin{equation}
\begin{aligned}
  \mathbb E_S[F(\zeta_S)-F(\zeta)\mid x]&=0,\\
  \operatorname{Cov}(F(\zeta_S)\mid x)
    &=AC_S(x)A^\top.
\end{aligned}
\label{eq:lift-transformed-disagreement}
\end{equation}
Factoring $\mathcal D=\mathcal R_j\circ F_j$ gives the local disagreement propagation identity
\begin{equation}
  \operatorname{tr}(G_D C_S)
  =\operatorname{tr}\!\left(
    J_{\mathcal R_j}J_{F_j}C_S
    J_{F_j}^\top J_{\mathcal R_j}^\top
  \right),
\label{eq:lift-disagreement-chain}
\end{equation}
with all Jacobians evaluated at clean real-data activations. In our combined experiments, the pixel decoder is first trained with \ours, frozen, and split at $h_j$ to train FuseLift with $p_{\mathrm{dit}}=0$, combining decoder-side disagreement regularization with prefix-relocated diffusion.

\FloatBarrier
\section{Extended Discussion}
\label{sec:discussion}

\paragraph{Why the two stages respond differently.}
The shared disagreement covariance $\Gamma$ in \Cref{prop:stage-disagreement} enters two distinct prediction tasks: the decoder reconstructs pixels across subset views, whereas the DiT predicts a deterministic full-layer target $\bar z$ along a noisy trajectory with time-weighted penalty $\rho(t)t^2$ (\Cref{eq:dit-penalty-main}). Empirically, applying \ours\ to both stages yields the lowest gFID across all three encoders on 40-epoch DiT-Base (\Cref{tab:fusereg-summary}), whereas 40-epoch DiT-XL attains its best unguided and guided gFID with decoder regularization and $p_{\mathrm{dit}}{=}0$ (\Cref{tab:scale-xl,tab:scale-xl-guided}). Generator-side subset regularization is thus most beneficial at smaller model capacity under a fixed budget, while decoder-side regularization consistently improves unguided generation across scales and motivates aligning diffusion coordinates with decoder sensitivity via FuseLift (\S\ref{sec:fuselift}).

\paragraph{\ours\ under internal guidance.}
Fixed-generator evaluations ($p_{\mathrm{dit}}{=}0$) show that decoder-swap gains shrink as guided sampling converges (\Cref{fig:decoder-swap,tab:ditxl_all}), likely because internal guidance keeps sampled latents closer to the clean manifold where standard decoders remain accurate. By contrast, when $p_{\mathrm{dit}}{>}0$, internal guidance couples the full and REPA heads along layer-disagreement directions (\S\ref{app:guidance-coupling}), accounting for the guidance-dependent $p_{\mathrm{dit}}$ preferences in \Cref{tab:scale-xl,tab:scale-xl-guided}.

\end{document}